\documentclass[12pt]{article}

\usepackage{url}

\usepackage{authblk}
\usepackage{natbib}
\usepackage{amsmath,amssymb}
\usepackage[margin=1in]{geometry}   % page margins
\usepackage{mathtools}
\usepackage{booktabs}
\usepackage{multirow}
\usepackage{longtable}
\usepackage{rotating}
\usepackage{makecell}
\usepackage{bm}
\usepackage{xcolor}
\usepackage{subcaption}
\usepackage{tikz}
\usetikzlibrary{arrows.meta, positioning, calc, fit, backgrounds, decorations.pathreplacing}
\usepackage{hyperref}
\usepackage{breakcites}
\usepackage{epstopdf}
\newcommand{\E}{\mathbb{E}}
\newcommand{\Var}{\mathrm{Var}}

\newcommand{\R}{\mathbb{R}}
\newcommand{\N}{\mathcal{N}}
\newcommand{\PDP}{\bar{\mu}}
\newcommand{\dPDP}{\Delta\bar{\mu}}

\begin{document}
\raggedbottom
% Title of paper
\title{Neural ODE enhanced linear mixed effect models for 
estimating complex association patterns of time-varying covariates with the marker trajectory}

\author{
  Zhe Aurore Li\textsuperscript{1,2},
  Quentin Clairon\textsuperscript{1,2},
  Cécilia Samieri\textsuperscript{1},
  Rodolphe Thiébaut\textsuperscript{1,2,3}, 
  Mélanie Prague\textsuperscript{1,2,3,*}, 
  Cécile Proust-Lima\textsuperscript{1,*, \textdagger} \\
  * Equal contribution \\
  \textdagger Correspondence: \texttt{cecile.proust-lima@u-bordeaux.fr}
}

\affil{
  $^1$Univ. Bordeaux, Inserm, BPH, UMR 1219, 33000 Bordeaux, France
  \\
  $^2$INRIA SISTM, 33000 Bordeaux, France\\
  $^3$Vaccine Research Institute, 94010 Creteil, France\\
}
% List of authors

% Running headers
\markboth%
{Z.A. Li and others}
{Neural ODE-LMM for longitudinal data}

\maketitle

\begin{abstract}
{Longitudinal cohort studies produce repeated data that enable the assessment of 
time-varying association patterns between exposures and health outcomes.
Classical linear mixed-effects models (LMMs) can accommodate a large variety of 
association patterns while accounting 
for the irregularly spaced, partially observed measurement. But they
require the analyst to pre-specify the functional form linking
the exposure history to the outcome.
We propose the Neural ODE-LMM, which embeds a Neural Ordinary Differential
Equation (Neural ODE) within the linear mixed-effects framework: a learned vector field encodes
covariate trajectories into a continuous-time latent state that drives both
the fixed- and random-effect design, while preserving the standard LMM
observation model. This retains classical likelihood-based inference while learning complex, 
potentially cumulative, covariate effects flexibly.
All parameters are estimated by maximising a penalised marginal likelihood.
To quantify covariate effects, we introduce contrasts of counterfactual
predictions that compare the expected outcome under alternative covariate
trajectories with variance estimated via the delta method.
In simulations, the model recovers both instantaneous and cumulative-burden
effects without prior
specification of the functional form.
Applied to the Trois-Cit\'{e}s (3C) cohort, a population-based
study of 7{,}324 participants, the method reveals
trajectory-dependent associations of BMI and fasting glucose
with cognitive decline.}
{Neural ordinary differential equations; linear mixed-effects models;
longitudinal data; counterfactual predictions; cognitive aging.}
\end{abstract}

\section*{Fundings}
This work was supported by Inserm through 
\href{https://exposomeinserm.fr/the-exposome-booster-program/}
{the Exposome Booster Program.}.  
This work was also carried out within project ANR EDyLES (ANR-25-CE36-0150) 
and the University of Bordeaux’s France 2030 program/RRI PHDS.

\section*{Acknowledgments}
We are deeply thankful to Christophe Tzourio, Thibault Mura Todesco 
and Catherine Helmer for the access to 3C data
used in Section \ref{sec:3c_data}. 
The experiments presented in this
paper were carried out using the PlaFRIM, supported by Inria, CNRS, Université de Bordeaux, Bordeaux INP and Conseil Régional d’Aquitaine.
Parts of this manuscript were refined with the assistance of Claude (Anthropic) 
for language editing and clarity improvements. 
All AI-generated suggestions were reviewed and validated by the authors. 
The scientific conception, methodological design, analyses and interpretations 
are entirely the work of the authors.

{\it Conflict of Interest}: None declared.
% ══════════════════════════════════════════════════════════════
\section{Introduction}
\label{sec:intro}
% ══════════════════════════════════════════════════════════════

Longitudinal cohort studies are a cornerstone of modern epidemiology, producing
repeated measurements on the same individuals at irregular intervals with
partial missingness across visits.
Leveraging longitudinal cohort data to understand how time-varying exposures shape 
long-term health outcomes requires both principled statistical inference  
and flexible representation of covariate effects, 
a combination that remains a central methodological challenge.
It is particularly acute in life-course epidemiology 
where an individual's health status likely reflects 
not the current exposure profile 
but the cumulative, dynamic and intricate evolution of risk factors over years 
\citep{wang2016quantifying}.
This is the case in cerebral aging, for instance with the association between cardiometabolic risk factors (adiposity, blood pressure, blood glucose
and lipid levels) and cognitive trajectory \citep{wagner2018}.

Linear mixed-effects models (LMMs) \citep{laird1982random,verbeke2000linear}
have long been the workhorse for the repeated data irregularly measured in longitudinal studies.
Decomposing variability into population-level fixed effects and subject-specific
random effects yields interpretable association at population scale despite inter-subject variability, 
and valid inference under missing-at-random mechanism.
The framework nonetheless requires the analyst to pre-determine the structure of association between covariates
and outcome, usually limiting them to simple functional forms (e.g., concomitant and independent associations), 
with the risk of providing misleading conclusions when the mechanisms are much more complicated.

Several lines of work have addressed distinct limitations of the LMM.
Generalized additive mixed models \citep{wood2017generalized} relax the
linear functional form through smooth splines but require pre-specified
interactions and still operate on instantaneous covariate--outcome
associations. Functional regression and its mixed-effects extensions
\citep{ramsay2005functional,greven2017general,volkmann2023multivariate}
treat covariate trajectories as functional objects and estimate their
effect nonparametrically, but the model class must still be chosen a
priori and the approach relies on basis expansions to reconstruct
trajectories rather than modelling their temporal dynamics directly.
More recently, semi-parametric deep learning extensions of the LMM have
replaced or augmented the fixed-effect component with a neural network
while preserving the random-effect structure, addressing identifiability
and inference for the parametric coefficients
\citep{rugamer2023pho,dorigatti2023frequentist} or replacing the
fixed- or random-effect predictor entirely with neural network
as LMMNN of
\citet{simchoni2023lmmnn} and GNMM of
\citet{mandel2023gnmm}.
These methods accommodate interactions and nonlinearity but not accumulation.

A separate line of work addresses continuous-time dynamics directly.
Neural ODEs \citep{chen2018neural} parameterize a
latent state by a learned vector field, enabling principled modelling of
irregularly sampled sequences.
Neural Controlled Differential Equations (Neural CDEs) \citep{kidger2020neural} drive the
latent state with an exogenous control path, making the dynamics sensitive to
the rate of change of covariates. Unlike mechanistic ODE models,
in which the vector field encodes a known biological or physical
mechanism, here Neural ODEs or Neural CDEs is a flexible nonparametric function
learned entirely from data.
Several works have incorporated random effects into the Neural ODE
framework to capture inter-individual variability
\citep{rubanova2019latent,nazarovs2021menode,braem2025node}.
These approaches, however, have been designed for prediction rather than for understanding the
association between time-varying covariates and an outcome. None embedded
the Neural ODE within a classical LMM observation model with an explicit
marginal likelihood, nor addressed the interpretability requirements of
epidemiological applications.

One of the major limitation of non-parameteric and machine learning-based approaches is the very limited interpretability.
A key challenge when replacing parametric fixed effects with a neural network is indeed
the loss of directly interpretable coefficients.
The machine learning literature has developed model-agnostic tools to address this,
notably with variable importance measures and partial dependence plots (PDPs) \citep{friedman2001greedy},
which isolate the marginal effect of a covariate by averaging predictions over the
empirical distribution of all other covariates \citep[see][for a comprehensive review]{molnar2022interpretable}.

In this paper we propose the Neural ODE-LMM, which embeds a Neural ODE
within the linear mixed-effects observation model to directly learn the
association structure between covariate trajectories and a marker trajectory.
This hybrid technique encodes the covariate trajectories
into a continuous-time latent state that drives both the fixed- and
random-effect design matrices, thus preserving the LMM statistical
properties. The complex patterns of association retrieved from the data
are then estimated along with confidence intervals using contrasts of
counterfactual predictions in the spirit of PDPs \citep{friedman2001greedy}.

The remainder of this paper is organised as follows.
Section~\ref{sec:method} reviews the classical LMM, introduces its neural extensions 
like LMMNN and GNMM and develops the Neural ODE-LMM as
a continuous-time extension.
Section~\ref{sec:pdp} presents the counterfactual prediction contrast and its
variance estimation within the penalised-likelihood framework.
Section~\ref{sec:simulations} reports a simulation study that aimed at demonstrating the flexibility 
of the approach and validating the proposed estimator.
Section~\ref{sec:application} applies the methodology to the
Trois-Cit\'{e}s (3C) cohort \citep{3CStudyGroup2003}, a
population-based aging study to assess the association between
cardiometabolic health and cognitive trajectory.
Section~\ref{sec:discussion} discusses the results, limitations and future
directions.

\section{Method}
\label{sec:method}

\subsection{Linear mixed-effects models}
\label{sec:lmm_review}

We begin with the standard LMM to establish notation.
We consider a sample of $N$ subjects ($i = 1,\ldots,N$) with $K_s$
time-invariant covariates $X_i^s \in \R^{K_s}$, as well as repeated
data collected at $n_i$ visits at subject-specific times
$t_{ij}$ ($j = 1,\ldots,n_i$): an outcome $Y_i(t_{ij}) \in \R$ and
$K$ time-varying covariates $X_i(t_{ij}) \in \R^K$.
Denoting by $Z_i(t_{ij}) \in \R^q$ the random-effect design vector,
generally composed of a subset of $X_i(t_{ij})$,
the LMM can be written as
$\label{eq:classical_lmm}
  Y_i(t_{ij}) = X_i^{s\top} \beta_s
  + X_i(t_{ij})^\top \beta_x
  + Z_i(t_{ij})^\top b_i + \varepsilon_{ij},
$
where $\beta_s \in \R^{K_s}$ and $\beta_x \in \R^K$ are
fixed-effect coefficients, $b_i \sim \N(0, D)$ are subject-specific
random effects with $q \times q$ covariance~$D$ and random noise
$\varepsilon_{ij} \sim \N(0, \sigma^2)$, generally assumed i.i.d..
Stacking the $n_i$ observations gives the outcome vector
$Y_i = \bigl(Y_i(t_{i1}),\ldots,Y_i(t_{in_i})\bigr)^\top$
and per-subject design matrices
$X_i \in \R^{n_i \times p}$, $Z_i \in \R^{n_i \times q}$.
By denoting $\beta = (\beta_s, \beta_x) \in \R^p$ which collects all
fixed-effect coefficients,
the estimation of $(\beta, D, \sigma^2)$ relies on the marginal distribution
of the outcome, obtained by integrating over~$b_i$:
$
  Y_i \sim \N\bigl(X_i \beta,\; V_i\bigr) 
  $ where
  $V_i = Z_i D Z_i^\top + \sigma^2 I_{n_i}$.
We derive the marginal negative
log-likelihood (NLL):
\begin{equation}\label{eq:classical_nll}
  \mathrm{NLL}(\beta, D, \sigma^2) = \frac{1}{2}\sum_{i=1}^{N}\Bigl[
    (Y_i - X_i\beta)^\top V_i^{-1}(Y_i - X_i\beta)
    + \log\det V_i + n_i \log 2\pi
  \Bigr].
\end{equation}
The maximum-likelihood estimator $(\hat\beta, \hat{D}, \hat\sigma^2)$
is obtained by minimising the NLL over $(\beta, D, \sigma^2)$.
Subject-specific predictions rely on the random effect predictor, 
often chosen as the conditional expectation of the random effect $\hat{b}_i = \mathbb{E}(b_i \mid Y_i)
  = \hat{D}\, Z_i(t)^\top \hat{V_i}^{-1}
    \bigl(Y_i - X_i^\top \hat{\beta}\bigr)$:
\begin{equation}\label{eq:blup_classical}
  \hat{Y}_i(t) = X_i(t)^\top \hat{\beta}
  + Z_i(t)^\top \hat{b}_i.
\end{equation}

\subsection{Neural mixed-effects models}
\label{sec:neural_lmm}

In the classical LMM, the fixed-effect structure
$X_i(t)^\top\beta$ requires the analyst to specify the form of
association between covariates and outcome in advance. This
parametric design is restrictive: interactions and nonlinear effects
must be declared explicitly, and time-varying covariates usually enter
through their current values only, implicitly assuming an
instantaneous association with no memory of past exposure. A natural solution is to replace the parametric fixed-effect and
random-effect structures with flexible nonlinear functions learned
from the data, while retaining the additive decomposition into
population-level and subject-specific components. 
This idea has been pursued in two recent
frameworks, both of which write the observation model as
\begin{equation}\label{eq:nn_lmm_general}
  Y_i(t_{ij}) = \underbrace{\rho_\psi\bigl(X_i(t_{ij})\bigr)^\top
  \beta}_{\text{neural fixed effects}}
  + \underbrace{g_{\xi}\bigl(Z_i(t_{ij})\bigr)^\top b_i}
  _{\text{neural random effects}} + \varepsilon_{ij},
\end{equation}
where $\rho_\psi$ and $g_\xi$ are neural networks: LMMNN
\citep{simchoni2023lmmnn} replaces both the fixed- and random-effect
design matrices with deep networks and estimates all parameters
jointly by minimising the marginal NLL~\eqref{eq:classical_nll} using
mini-batch stochastic gradient descent (SGD). The GNMM \citep{mandel2023gnmm} 
similarly embeds a feed-forward
network, using a Laplace-approximated
quasi-likelihood for estimation. They share a fundamental
limitation: the neural network operates on the \emph{instantaneous}
covariate vector $X_i(t_{ij})$ at each visit, with no mechanism to
encode how covariates evolved between visits. Capturing
trajectory-dependent effects (cumulative burden, rate of change) requires
a model that integrates the covariate trajectory in continuous time.

\subsection{The Neural ODE-LMM}
\label{sec:model}

\begin{figure}[!p]
\centering
\begin{tikzpicture}[
  >=Stealth,
  node distance=1.2cm and 1.4cm,
  block/.style={draw, rounded corners, minimum height=0.9cm, minimum width=1.6cm, align=center, font=\small},
  data/.style={draw, minimum height=0.7cm, minimum width=1.2cm, align=center, font=\small},
  lmm/.style={draw, rounded corners, minimum height=0.9cm, minimum width=1.4cm, align=center, font=\small, fill=gray!12},
  arrow/.style={->, thick},
  darrow/.style={->, thick, dashed},
]

% ── Row 1 (main flow) ──
\node[data] (xbase) {$X_{i}(0), X^s_i, t_{i0}$};
\node[block, right=1.4cm of xbase] (enc) {Encoder\\$\mathrm{Enc}_\eta$};
\node[block, right=1.4cm of enc] (ode) {Neural ODE\\$f_\phi$};
\node[right=1.0cm of ode, font=\small] (zt) {$\Lambda_i(t_{ij})$};
\node[lmm, right=1.6cm of zt] (rho) {$\rho_\psi^\top \beta$};
\node[lmm, below=0.6cm of rho] (gxi) {$g_\xi^\top b_i$};
\node[right=1.2cm of rho, yshift=-0.45cm, font=\small] (yij) {$\hat{Y}_i(t)$};

% ── Row 0 (above): time-varying covariates ──
\node[data, above=1.2cm of xbase] (xt) {$\bar{X}_i(t),\, M_i(t),\, X^s_i, \, t$};

% ── Main-flow arrows ──
\draw[arrow] (xbase) -- (enc);
\draw[arrow] (enc) -- node[above, font=\footnotesize] {$\Lambda_i(0)$} (ode);
\draw[arrow] (ode) -- (zt);
\draw[arrow] (zt) -- (rho);
\draw[arrow] (zt) -- (gxi);
\draw[arrow] (rho) -- (yij);
\draw[arrow] (gxi) -- (yij);

% ── Pathway 1: x(t) → Neural ODE (solid) ──
\draw[arrow] (xt.east) -| ([xshift=-0.3cm]ode.north);

% ── Pathway 2: x(t) → decoder (dashed, instantaneous pathway) ──
\draw[darrow, rounded corners=6pt]
  (xt.east) -- (xt.east -| rho.north) -- (rho.north);

% ── Box around decoder with LMM label below ──
\node[draw, rounded corners, inner sep=6pt,
      fit=(rho)(gxi), label={[font=\footnotesize]below:LMM observation model}] (lmmbox) {};

% ── Latent trajectory inset ──
\begin{scope}[shift={(ode.south)}, yshift=-3.5cm, xshift=-2.0cm]
  \draw[->, thin, gray!70] (-0.2,0) -- (4.2,0)
      node[right, font=\footnotesize, black] {$t$};
  \draw[->, thin, gray!70] (0,-0.2) -- (0,1.4)
      node[above, font=\footnotesize, black] {$\Lambda_i(t)$};
  \draw[blue!70!black, thick]
      plot[smooth, tension=0.55] coordinates {
        (0.0,0.25) (0.45,0.50) (1.0,0.80) (1.55,0.65)
        (2.2,0.95) (2.8,0.75)  (3.4,1.05) (4.0,0.95)
      };
  \foreach \tx/\ty/\lab in {
      0.0/0.25/$t_{i1}$,
      1.0/0.80/$t_{i2}$,
      2.2/0.95/$t_{i3}$,
      3.4/1.05/$t_{i4}$%
    }{
      \fill[blue!70!black] (\tx,\ty) circle (2pt);
      \draw[gray, densely dotted, thin] (\tx,0) -- (\tx,\ty);
      \node[below=2pt, font=\footnotesize, gray!80!black] at (\tx,0) {\lab};
    }
  \node[font=\footnotesize, blue!70!black, anchor=south west] at (2.6,1.1)
      {$\Lambda_i(t_{ij})$};
\end{scope}

\end{tikzpicture}
\caption{Architecture of the Neural ODE-LMM.
Baseline covariates $X_i(0)$ are mapped to an initial latent state $\Lambda_i(0)$ by the encoder.
The Neural ODE integrates the latent state forward in continuous time, driven by interpolated time-varying covariates.
At each visit, two learned networks $\rho_\psi$ and $g_\xi$ produce fixed- and random-effect design vectors from $\Lambda_i(t_{ij})$, preserving the standard LMM observation model.
Dashed arrow: the instantaneous pathway gives $\rho_\psi$ direct access to current covariate values; a group lasso penalty encourages parsimonious use of this pathway for time-varying covariates.
Bottom inset: the continuous latent trajectory $\Lambda_i(t)$, with filled dots marking the states $\Lambda_i(t_{ij})$ at (irregularly spaced) observation times.}
\label{fig:architecture}
% {\footnotesize\baselineskip=12pt\raggedright\noindent 
% \textbf{Alt text}: Architecture
% diagram of the Neural ODE linear mixed model, read left to right. Baseline
% covariates enter an encoder that produces an initial latent state. A Neural
% ODE block integrates this state forward in continuous time, driven from above
% by the time-varying covariates, and outputs the latent state at each visit.
% This latent state feeds two learned networks that produce the fixed-effect and
% random-effect design vectors, which combine in the linear mixed model
% observation block to give the predicted outcome. A dashed arrow carries the
% time-varying covariates directly to the fixed-effect network, representing the
% instantaneous pathway. A bottom inset shows the continuous latent trajectory
% as a curve, with filled dots marking the states at four irregularly spaced
% observation times.\par}
\end{figure}
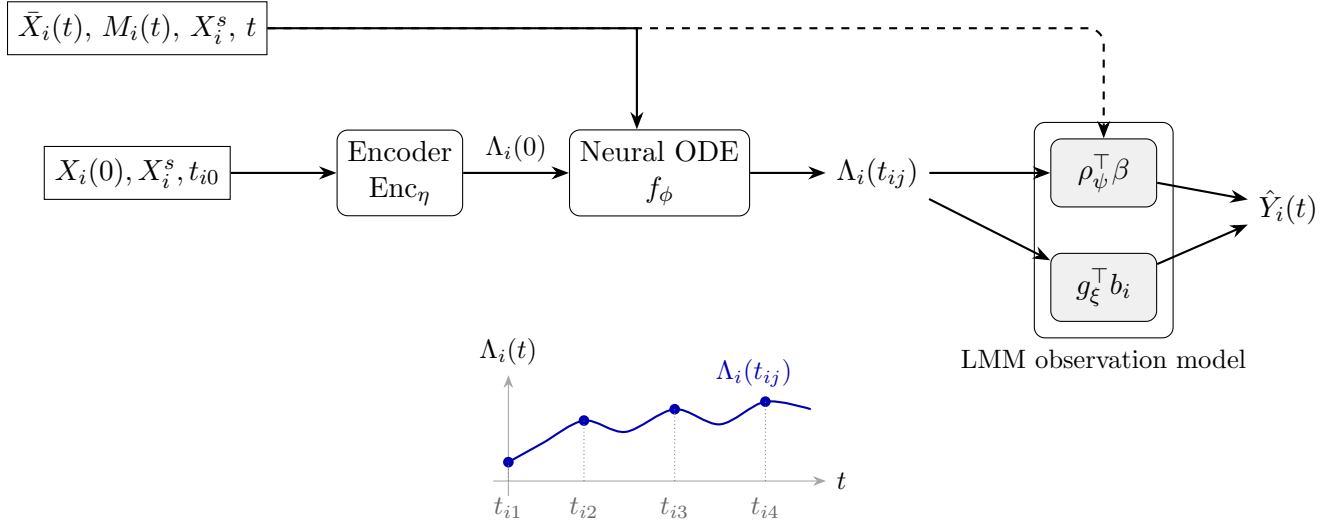

We generalise the neural mixed-effects
model~\eqref{eq:nn_lmm_general} by introducing a $d$-dimensional continuous-time latent state
$\Lambda_i(t) \in \R^d$ that encodes
the covariate
trajectory up to time~$t$ (Figure~\ref{fig:architecture}). Since
time-varying covariates $X_i(t_{ij})$ may be partially missing, we
fill unobserved covariate values by within-subject linear interpolation,
yielding a continuous path $\bar{X}_i(t) \in \R^K$ together with a
binary mask $M_i(t) \in \{0,1\}^K$ indicating whether each covariate
was observed at time~$t$.

The latent state evolves according to a Neural ODE driven by the
time-varying covariates:
\begin{equation}\label{eq:ode}
  \frac{d}{dt}\Lambda_i(t) = f_\phi\bigl(\Lambda_i(t),\;
  \bar{X}_i(t),\; M_i(t),\; t\bigr),
\end{equation}
where $f_\phi$ is a neural network with nonlinear activation function. For predictive purposes, we need a starting point, or initial condition
$\Lambda_i(0) = \mathrm{Enc}_\eta(X_i(0), X^s_i)$, where
$\mathrm{Enc}_\eta$ is a feedforward network. Integrating the
vector field forward gives
\begin{equation}\label{eq:ode_integral}
  \Lambda_i(t) = \Lambda_i(0) + \int_0^{t}
  f_\phi\bigl(\Lambda_i(\tau),\; \bar{X}_i(\tau),\;
  M_i(\tau),\; \tau\bigr)\, d\tau,
\end{equation}
so that $\Lambda_i(t)$ encodes the full covariate history up to
time~$t$.

The observation model preserves the LMM structure:
\begin{equation}\label{eq:lmm}
  Y_{i}(t_{ij}) = \rho_\psi\bigl(\Lambda_i(t_{ij}),\;
  \bar{X}_i(t_{ij}),\; M_i(t_{ij}),\;
  X_i^s\bigr)^\top \beta \;+\;
  g_\xi\bigl(\Lambda_i(t_{ij})\bigr)^\top\,
  b_i \;+\; \varepsilon_{ij}.
\end{equation}
The fixed-effect network $\rho_\psi : \R^{d+2K+K_s} \to \R^p$
receives both the latent state $\Lambda_i(t_{ij})$ which encodes
the accumulated covariate history and the current covariate values
$(\bar{X}_i(t_{ij}), M_i(t_{ij}), X_i^s)$ which provide a direct
instantaneous pathway that bypasses the ODE integration, hereafter 
referred to as the \emph{skip connection}. This dual-input design lets the data
determine whether each covariate operates through its accumulated
history or its current value. In practice, however, the
instantaneous pathway offers a shorter gradient path than the ODE
encoder, so that gradient-based optimisation preferentially routes
signal through the skip connection; without regularisation, the ODE
may fail to learn cumulative representations even when they are
present in the data. To address this, the direct covariate inputs
to $\rho_\psi$ are organised in groups: for each time-varying
covariate~$k$, the interpolated value $\bar{X}_{i,k}(t_{ij})$ and
its mask $M_{i,k}(t_{ij})$ form one group; each static covariate
forms a separate group. A group lasso penalty
\citep{yuan2006model} on the first-layer weights of $\rho_\psi$
shrinks entire groups toward zero, encouraging the ODE pathway
to absorb the covariate effect.
Let $W \in \R^{h_1 \times (d + 2K + K_s)}$ denote the
first-layer weight matrix of $\rho_\psi$, where $h_1$ is the number
of hidden units in its first layer, and let
$W_g \in \R^{h_1 \times c_g}$ denote the submatrix corresponding to
skip group~$g$ ($g = 1, \ldots, G = K + K_s$; $c_g = 2$ for a dynamic covariate; $c_g = 1$ for a
static covariate). The group lasso penalty is
$\lambda_{\mathrm{GL}}\sum_{g=1}^{G}\|W_g\|_F$
where $\|W_g\|_F = \bigl(\sum_{a=1}^{h_1}\sum_{b=1}^{c_g}
W_{g,ab}^2\bigr)^{1/2}$.
When $\|W_g\|_F = 0$, all weights connecting group~$g$ to the hidden
layer vanish simultaneously and covariate~$g$ can influence the
prediction only through the ODE latent state; when
$\|W_g\|_F > 0$, the covariate retains a direct instantaneous
pathway.

\subsection{Estimation}
\label{sec:estimation}

The full parameter set is
$\theta = \{\theta_{\mathrm{nn}},\, \beta,\, D,\, \sigma^2\}$,
where $\theta_{\mathrm{nn}} = \{\eta, \phi, \psi, \xi\}$ collects
the neural-network weights. Stacking the decoder outputs over the $n_i$ visits
yields per-subject design matrices
$X_{i,\theta_{\mathrm{nn}}} \in \R^{n_i \times p}$ and
$Z_{i,\theta_{\mathrm{nn}}} \in \R^{n_i \times q}$, with rows
$\rho_\psi(\Lambda_i(t_{ij}), \bar{X}_i(t_{ij}), M_i(t_{ij}),
X_i^s)^\top$ and $g_\xi(\Lambda_i(t_{ij}))^\top$ respectively.
The marginal distribution for subject~$i$ is
$
  Y_i \sim \N\bigl(X_{i,\theta_{\mathrm{nn}}} \beta,\; V_i\bigr),
$ where
  $V_i = Z_{i,\theta_{\mathrm{nn}}}\, D\,
  Z_{i,\theta_{\mathrm{nn}}}^\top + \sigma^2 I_{n_i}.
$ All parameters are estimated jointly by minimising the penalised
marginal NLL:
\begin{equation}\label{eq:nll}
\begin{split}
  pL(\theta) ={} &
  \underbrace{\frac{1}{N} \sum_{i=1}^{N} \frac{1}{2}\Bigl[
    \log\det V_i
    + (Y_i - X_{i,\theta_{\mathrm{nn}}}\beta)^\top
      V_i^{-1}(Y_i - X_{i,\theta_{\mathrm{nn}}}\beta)
    + n_i \log(2\pi)\Bigr]}_{\text{marginal NLL}} \\
  & \qquad + \underbrace{\lambda_{\mathrm{GL}}\sum_{g=1}^{G}
    \|W_g\|_F}_{\text{group lasso}}
  + \underbrace{\frac{\lambda_{\mathrm{wd}}}{2}\|\theta_{\mathrm{nn}}^{-W}\|^2}_{\text{weight decay}},
\end{split}
\end{equation}
where $\theta_{\mathrm{nn}}^{-W} = \theta_{\mathrm{nn}} \setminus \{W_g\}_{g=1}^G$ 
excludes the first decoder layer weights which are already penalised by the group lasso term.
The weight decay acts on $\theta_{\mathrm{nn}}^{-W}$ to avoid overfitting; 
$\beta$,
$D$, and $\sigma^2$ are unpenalised. Both penalties depend only on
model parameters and not on the data, giving the estimation the structure 
of a penalised likelihood
\citep{commenges2015penalized}.

The data are split into a training set and a validation set.
All parameters are estimated simultaneously on the training set
by mini-batch SGD using the Adam
optimiser \citep{kingma2015adam}. At each epoch, the penalised
marginal NLL~\eqref{eq:nll} is evaluated on the validation set;
training stops when the validation NLL does not improve for
$P$ consecutive epochs, and the checkpoint with the lowest
validation NLL is retained.

\paragraph{Variance of the estimator.}
We estimate the variance of $\hat\theta$ using
the inverse of a shrinkage-regularised empirical Fisher
matrix:
$\hat{\Sigma} = F_{\mathrm{reg}}^{-1}$.
The $F_{\mathrm{reg}}$ is defined as:
\begin{equation}\label{eq:ledoit_wolf}
  F_{\mathrm{reg}} = (1 - \alpha^*)\, F
  + \alpha^*\,
  \frac{\mathrm{tr}(F)}{\dim(\theta)}\, I,
\end{equation}
where $\alpha^* \in [0,1]$ is the true shrinkage intensity that minimises the expected Frobenius
loss \citep{ledoit2004well} and $F = \sum_{i=1}^{N} U_i\, U_i^\top$
is the empirical Fisher matrix.
The per-subject negative log-likelihood can be written as
\[
  \mathrm{nll}_i(\theta) = \tfrac{1}{2}\bigl[\log\det V_i
  + (Y_i - X_{i,\theta_{\mathrm{nn}}}\beta)^\top V_i^{-1}
  (Y_i - X_{i,\theta_{\mathrm{nn}}}\beta) + n_i\log 2\pi\bigr],
\]
and the per-subject penalised score is
$
  U_i = \nabla_\theta\, \mathrm{nll}_i(\hat\theta)
  + c_{\mathrm{pen}}$,
and $
  c_{\mathrm{pen}} = \lambda_{\mathrm{GL}}\,\nabla_\theta
  \textstyle\sum_{g=1}^{G} \|W_g\|_F\big|_{\hat\theta}
  \;+\; \lambda_{\mathrm{wd}}\,
  \hat\theta^{-W}_{\mathrm{nn}}.
$
The stationarity condition $\bar{U} \approx 0$ is
verified via the ratio
$\|\bar{U}\|/\overline{\|U_i\|}$, where
$\bar{U} = \frac{1}{N}\sum_i U_i$ and
$\overline{\|U_i\|} = \frac{1}{N}\sum_i \|U_i\|$.

\subsection{Prediction}
\label{sec:prediction}

Subject-specific predictions use the BLUP as in the classical
LMM~\eqref{eq:blup_classical}. Two modes are distinguished, both mirroring classical 
random effect predictor for LLM exposed in section \ref{sec:lmm_review}.

\paragraph{Fit mode.}
All $n_i$ observations of subject~$i$ are used:
\begin{equation}\label{eq:fit_blup}
  \hat{b}_i
  = \hat{D}\, Z_{i,\hat\theta_{\mathrm{nn}}}^\top
  V_i^{-1}(Y_i - X_{i,\hat\theta_{\mathrm{nn}}}\hat{\beta}),
\end{equation}
\begin{equation}\label{eq:fit_pred}
  \hat{Y}_i^{\,\mathrm{fit}}(t) =
  \rho_\psi\bigl(\Lambda_i(t),\; \bar{X}_i(t),\;
  M_i(t),\; X_i^s\bigr)^\top \hat{\beta}
  \;+\; g_\xi\bigl(\Lambda_i(t)\bigr)^\top \hat{b}_i.
\end{equation}

\paragraph{Forecast mode.}
Only observations up to a landmark time~$t$ are used:
\begin{equation}\label{eq:forecast_blup}
  \hat{b}_i(t)
  = \hat{D}\, Z_i^{(\leq t)\top}
  V_i^{(\leq t)\,-1}
  \bigl(Y_i^{(\leq t)} - X_i^{(\leq t)}\hat{\beta}\bigr),
\end{equation}
\begin{equation}\label{eq:forecast_pred}
  \hat{Y}_i^{\,\mathrm{fcst}}(t^*) =
  \rho_\psi\bigl(\Lambda_i^{(\leq t)}(t^*),\;
  \bar{X}_i^{(\leq t)}(t^*),\;
  M_i^{(\leq t)}(t^*),\; X_i^s\bigr)^\top \hat{\beta}
  \;+\; g_\xi\bigl(\Lambda_i^{(\leq t)}(t^*)\bigr)^\top
  \hat{b}_i(t),
\end{equation}
where $Y_i^{(\leq t)}$, $X_i^{(\leq t)}$, $Z_i^{(\leq t)}$, and
$V_i^{(\leq t)}$ denote the outcome vector, design matrices and
marginal covariance restricted to visits at or before~$t$.

\subsection{Hyperparameter selection}
\label{sec:hyperparams}

The model involves several architectural and regularisation choices,
including the latent dimension~$d$, the fixed-effect and
random-effect output dimensions~$p$ and~$q$, the number of hidden
units in each network, the ODE solver order, and the regularisation
weights. Exhaustive search over the full configuration space is
impractical; we therefore fix the network widths, the
random-effect dimension~$q$, and the ODE solver to a fourth-order
Runge--Kutta scheme, and select the latent dimension~$d$, the
fixed-effect output dimension~$p$, the penalty
weight~$\lambda_{\mathrm{GL}}$, and the covariate pathway
configuration by ranking candidates according to their
marginal NLL\@ on the validation set. 

% ══════════════════════════════════════════════════════════════
\section{Interpretability and inference}
\label{sec:pdp}
% ══════════════════════════════════════════════════════════════

Most applications in epidemiology aim to assess the association between a set of
covariates (exposures) and an outcome (marker of the disease), thus
requiring interpretable association measures to be extracted from the
fitted model, along with uncertainty assessment.
We therefore adopt a counterfactual strategy: we compare the model's
predictions under alternative covariate trajectories, holding
everything else fixed, and summarise the difference through a partial
dependence contrast.

\subsection{Counterfactual prediction contrasts}

For a given covariate~$k$ and subject~$i$, replacing the observed path $\bar{X}_{i,k}(\cdot)$ with a
counterfactual path~$v(\cdot)$ while keeping all other covariates at their
observed values yields the counterfactual conditional expectation:
\begin{equation}\label{eq:counterfactual_i}
  \hat\mu^{(k \leftarrow v)}_{i}(t)
    = \E\bigl(Y_i^*(t) \mid \bar{X}_i^{(k)}(t) = v(t),\;
      \bar{X}_i^{(-k)}(t) = \bar{x}_i^{(-k)}(t)\bigr)
    = \rho_{\hat\psi}\bigl(\Lambda_{i,v}(t),\,
      \bar{X}_{i,v}(t),\, M_i(t),\, X_i^s\bigr)^\top \hat\beta,
\end{equation}
where $\Lambda_{i,v}(t)$ denotes the latent state obtained by integrating the
Neural ODE with covariate~$k$ set to~$v(\cdot)$ and $\bar{X}_{i,v}(t)$ the
corresponding covariate vector with covariate~$k$ replaced.
The population-level counterfactual mean under intervention~$v$ is then obtained by
marginalising over the joint distribution of the remaining
covariates, following the same principle as partial dependence
\citep{friedman2001greedy}:
\begin{equation}\label{eq:pdp_pop}
  \PDP(v,\, t) = \E_{X_{-k},\, X^s}\bigl[\hat\mu^{(k \leftarrow v)}(t)\bigr].
\end{equation}
In practice, this expectation is estimated by averaging over the $N$ observed subjects:
\begin{equation}\label{eq:pdp}
  \widehat{\PDP}(v,\, t) = \frac{1}{N}\sum_{i=1}^{N}
    \hat\mu^{(k \leftarrow v)}_{i}(t).
\end{equation}
The association between the covariate and the outcome is then quantified by the counterfactual contrast between 
two intervention paths $v_1(\cdot)$ and~$v_2(\cdot)$ at any time $t$:
\begin{equation}\label{eq:dpdp}
  \dPDP(v_1, v_2;\, t)
    = \widehat{\PDP}(v_1,\, t) - \widehat{\PDP}(v_2,\, t).
\end{equation}
In practice, the counterfactual contrast is evaluated at a grid of times
$t^*_1 < \cdots < t^*_L$ spanning the follow-up window.
The paths $v_1(\cdot)$ and $v_2(\cdot)$ can be constant
(e.g.\ the 25th and 75th percentiles of the observed distribution)
or time-varying (e.g.\ a trajectory declining from Q75 to Q25
over follow-up); the latter allows the $\dPDP$ to detect
trajectory-dependent effects. When the variables marginalized over 
satisfy the back-door criterion relative to $v$, 
the $\dPDP$ coincides 
with a causal contrast under Pearl's do-calculus \citep{zhao2021causal}.

\subsection{Variance estimation via delta method}
\label{sec:inference}

The estimated $\widehat{\dPDP}(t)$ is a smooth function
of~$\hat\theta$. Treating $\hat\theta$ as an M-estimator of the 
penalised likelihood~\eqref{eq:nll} with asymptotic variance 
$F_{\mathrm{reg}}^{-1}$, the delta method yields the approximate variance 
of the $\dPDP$ at any time~$t$:
\begin{equation}\label{eq:delta_method}
  \Var\bigl(\widehat{\dPDP}(t)\bigr) \approx
  \nabla_\theta \widehat{\dPDP}(t)\big|_{\hat\theta}^{\;\top}\;
  F_{\mathrm{reg}}^{-1}\;
  \nabla_\theta \widehat{\dPDP}(t)\big|_{\hat\theta},
\end{equation}
where the gradient is evaluated at the estimated parameter
value~$\hat\theta$ and
$F_{\mathrm{reg}}$ is the regularised empirical Fisher information defined
in~\eqref{eq:ledoit_wolf}.
Pointwise 95\% confidence intervals are
$\widehat{\dPDP}(t) \pm 1.96\,
\sqrt{\Var\bigl(\widehat{\dPDP}(t)\bigr)}$.

\section{Simulation study}
\label{sec:simulations}
 
We report this simulation study following 
the ADEMP framework of \citet{morris2019using}, 
structuring the design around explicit Aims, 
the Data-generating process (DGP), the target Estimands, 
the Methods under comparison, and the pre-specified Performance measures.
 
\subsection{Aims}
\label{sec:sim_aims}
 
The simulation study assesses the ability of the Neural ODE-LMM to recover the $\dPDP$ 
in two contrasting regimes: (i)~an instantaneous linear association 
between the exposure and the outcome, modulated by a time-invariant covariate, 
which a correctly specified LMM can recover (S1); 
and (ii)~a cumulative effect of the exposure, 
which an LMM could accommodate only if the analyst pre-specified 
the correct accumulation functional, whereas the Neural ODE-LMM learns it from the data (S2).
 
\subsection{Data-generating process}
\label{sec:dgm}
 
The simulation design was anchored to the population-based 3C cohort
described in Section~\ref{sec:3c_data}, with a focus on the association
between body mass index (BMI) and the cognitive trajectory.
We considered the $N = 5{,}859$ participants of the analytical 3C sample
and retained their observed time-independent covariates: centered
baseline age ($\mathrm{AGE}_i$, centered at 74.86~years), sex and 3 educational levels.
Each subject's follow-up window $(t_{\min,i}, t_{\max,i})$ was
preserved from the 3C data, but visit times within the window were
set on a regular grid with spacing proportional to the window length,
yielding between 3 and 15 visits per subject.
 
For each of $R = 100$ replications, the time-varying covariate BMI was
simulated from a linear mixed model (random intercept and slope on
time) whose parameters were estimated from the BMI values of 3C data, producing
realistic within-subject variability.
Repeated outcome values $Y_{ij}$ were then generated according to:
$\label{eq:dgp}
  Y_{ij} = X_{ij}^\top \beta^\ast
  + f_s\bigl(\mathrm{BMI}_{ij},\,
    \mathrm{AGE}_{i},\, t_{ij}\bigr)
  + Z_{ij}^\top b_i + \varepsilon_{ij},
$
where $X_{ij}$ contains a basis of natural splines for time
(3~d.f., internal knots at the empirical quartiles of follow-up time),
sex, and educational level;
$Z_{ij}$ contains the basis of natural splines for time used in the
random-effect structure (2~d.f., internal knot at the median follow-up
time); $b_i \sim \N(0, D_0)$ with $D_0 \in \R^{3 \times 3}$;
$\varepsilon_{ij} \sim \N(0, {\sigma_0}^2)$; and $f_s$ is a
parameterised function specifying the BMI--outcome association,
detailed below.
Parameter values and spline knots were all calibrated from a preliminary 
LMM fit on the 3C data (reported in Supplementary Table \ref{tab:sim_params}).

\subsection{Scenarios}
\label{sec:scenarios}
 
\textbf{Scenario S1} (instantaneous BMI $\times$ age association).
The BMI--outcome association is instantaneous and age-modulated:
$
  f_s(\mathrm{BMI}_{ij}, \mathrm{AGE}_i, t_{ij}) = \beta_{\mathrm{BMI}}\, \mathrm{BMI}_{ij} + \beta_{\mathrm{int}}\, \mathrm{BMI}_{ij} \cdot \mathrm{AGE}_{i},
$
with $\beta_{\mathrm{BMI}} = -0.30$ and $\beta_{\mathrm{int}} = -0.05$.
A correctly specified LMM can recover this effect exactly; 
the scenario serves as a calibration benchmark to verify that the additional flexibility of the
Neural ODE-LMM does not compromise inferential validity when the true
effect is simple.
 
\noindent \textbf{Scenario S2} (cumulative BMI burden).
The outcome level depends on the time-integrated BMI trajectory:
$
  f_s(\mathrm{BMI}_{ij}, \mathrm{AGE}_i, t_{ij}) = -0.05\int_0^{t_{ij}} \mathrm{BMI}_i(\tau)\,d\tau.
$
There is no instantaneous BMI effect ($\beta_{\mathrm{BMI}}=0$)
and no age interaction ($\beta_{\mathrm{int}}=0$).
No standard LMM can recover this effect without an explicitly 
cumulative definition. 
More details are reported in Supplementary Section \ref{sec:supp_sim_params}.
 
\subsection{Estimands}
\label{sec:estimands}
 
The primary estimand is the $\dPDP(v_{\mathrm{lo}}, v_{\mathrm{hi}};\, t)$ as defined in~\eqref{eq:dpdp}, 
evaluated at constant intervention values defined according to the 25th and 75th percentiles of the observed BMI distribution.
We report the $\dPDP$ and its variance~\eqref{eq:delta_method} for a grid of times spanning the follow-up window.
 
\subsection{Methods}
\label{sec:sim_methods}
 
For all Neural ODE-LMM configurations, the sample was split into 80\%
for training and 20\% for computing the early-stopping criterion.
 
\paragraph{Scenario S1.}
We compare the Neural ODE-LMM against a correctly specified true LMM
that matches the DGP.
The covariate pathway configuration was selected by validation NLL
among the candidates described in Section~\ref{sec:hyperparams};
the best-performing configuration routed BMI and all static covariates
through the skip pathway only, excluding BMI from the ODE vector field,
with no group lasso penalty ($\lambda_{\mathrm{GL}} = 0$).
The full pathway comparison is reported in Supplementary
Table~\ref{tab:supp_model_sel}.
 
\paragraph{Scenario S2.}
We didn't consider a LMM for this DGP. We only assessed the neural ODE-LMM, since
$f_s = -0.05\int_0^{t} \mathrm{BMI}(\tau)\,d\tau$ requires an explicit
cumulative definition.
The covariate pathway
configuration was again selected by NLL on the validation set;
the best-performing configuration routed BMI through the ODE vector
field, and a group lasso penalty ($\lambda_{\mathrm{GL}} > 0$) on the
skip connection weights encouraged the ODE to learn the cumulative
representation
(Supplementary Table~\ref{tab:supp_model_sel}).

To demonstrate that the model captures trajectory-dependent effects, we
constructed two counterfactual BMI trajectories that reach the same
terminal value but differ in their history: a ``late spike'' profile
that starts at the 25th percentile of BMI and rises linearly to the
75th percentile, and a ``late decline'' profile that starts at the
75th percentile and falls to the 25th percentile.
Under the cumulative-burden DGP, the late decline profile accumulates
higher BMI over most of the follow-up and should therefore produce
worse cognitive outcomes.
 
\subsection{Performance measures}
\label{sec:sim_perf}
 
Over the $R = 100$ replications, we report four quantities.
Let $\widehat{\dPDP}^{(r)}$ denote the estimate from replicate~$r$ and $\dPDP_0$ the true value.
\begin{itemize}\setlength\itemsep{2pt}
  \item \emph{Bias}: $\overline{\dPDP} - \dPDP_0$, where $\overline{\dPDP} = R^{-1}\sum_{r=1}^R \widehat{\dPDP}^{(r)}$.
  \item \emph{Empirical variance}: $\mathrm{Var}_{\mathrm{emp}} = (R-1)^{-1}\sum_r \bigl(\widehat{\dPDP}^{(r)} - \overline{\dPDP}\bigr)^2$.
  \item \emph{Estimated variance}: $\overline{\mathrm{Var}}_{\mathrm{est}} = R^{-1}\sum_r \widehat{\mathrm{Var}}^{(r)}$, from the delta method~\eqref{eq:delta_method}.
  \item \emph{Coverage}: proportion of replicates whose 95\% confidence interval
    $\widehat{\dPDP}^{(r)} \pm 1.96\,\sqrt{\widehat{\mathrm{Var}}^{(r)}}$
    contains the true $\dPDP_0$.
\end{itemize} 
\subsection{Results}
\label{sec:sim_results}
 
\paragraph{Scenario S1: instantaneous BMI $\times$ age association}
 
Table~\ref{tab:sim_results} reports the quantitative results.
The Neural ODE-LMM succeeds in retrieving the $\dPDP$ with small relative bias (approximately 3-6\%) and 
correct coverage rates of the 95\% confidence interval at every time point, 
even though the bias remains larger than with the oracle LMM as expected.
The purpose of S1 is not to demonstrate improved inference, 
but to verify that the model produces near-nominal coverage 
and bounded bias when the linear structure lies 
inside the hypothesis class the neural decoder can represent.
 
\paragraph{Scenario S2: cumulative BMI burden}
 
\begin{figure}[!pt]
  \centering
  \includegraphics[width=\textwidth]{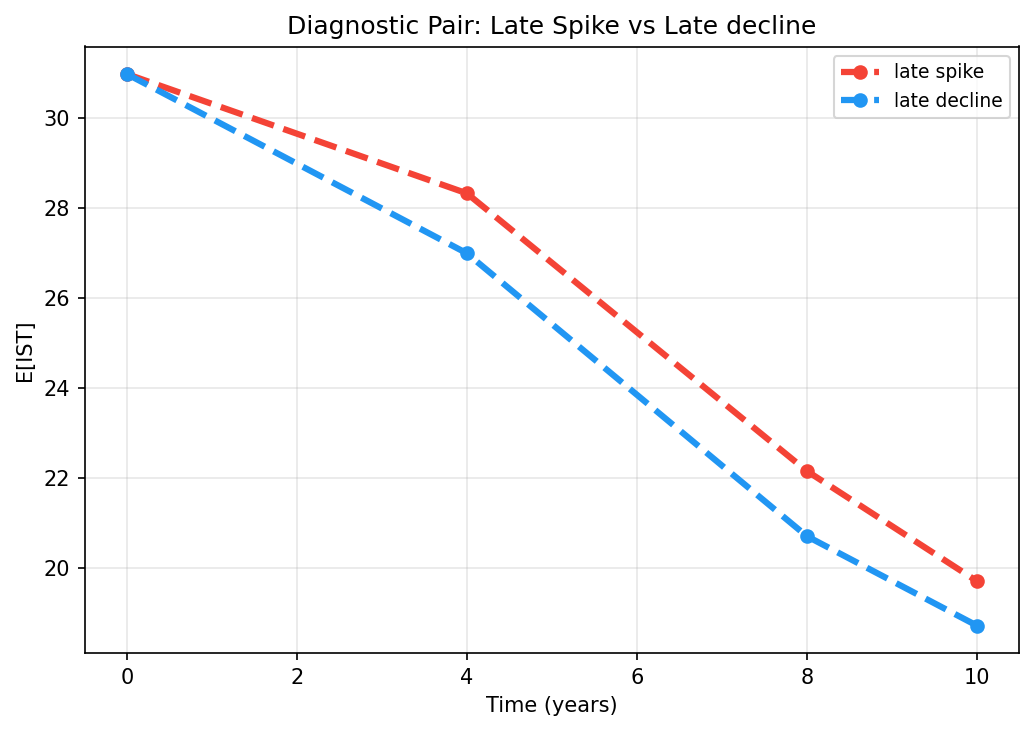}
  \caption{Scenario S2 (cumulative BMI burden): counterfactual trajectory-profile 
  diagnostic on a representative replicate. 
  The two counterfactual BMI profiles share the same terminal value 
  but differ in their history: ``late spike'' starts at Q25 and rises to Q75, 
  while ``late decline'' starts at Q75 and falls to Q25. 
  The Neural ODE-LMM correctly assigns worse predicted outcome to the ``late decline'' 
  profile, which accumulated higher BMI over most of the 
  follow-up.}
  \label{fig:s2_pdp_reg}
% {\footnotesize\baselineskip=12pt\raggedright\noindent 
% \textbf{Alt Text}: Graph
% comparing two counterfactual body mass index histories that share the same
% final value but differ in path: late spike, which starts low and rises, and
% late decline, which starts high and falls. Time in years, from zero to ten, is
% on the horizontal axis and predicted cognitive score on the vertical axis.
% Both dashed curves fall from about thirty-one toward twenty over the
% follow-up. The late decline curve stays below the late spike curve throughout,
% showing the model assigns a worse predicted outcome to the profile that
% carried higher body mass index for longer.\par}
\end{figure}

This scenario reveals the central advantage of the ODE encoder.
Table~\ref{tab:sim_results} presents the quantitative summary.
The Neural ODE-LMM recovers the cumulative effect across all
evaluation times: the true $\Delta\mu_0$ grows linearly
from $-0.50$ at $t=2$ to $-2.50$ at $t=10$, reflecting the
progressive accumulation of the BMI burden, and the estimated
$\Delta\bar{\mu}$ tracks this growth with small relative bias
(approximately 3--5\%).
The delta-method variance estimator closely matches the empirical
variance, with $\widehat{\text{Var}}_{\text{est}}$ nearly identical
to $\text{Var}_{\text{emp}}$ at later time points confirming that the uncertainty
quantification remains well calibrated even as the cumulative
effect grows.
Coverage of the 95\% confidence interval is near-nominal ranging from 0.92 to 0.98.
Figure~\ref{fig:s2_pdp_reg}
confirms that the model correctly assigns worse outcomes to the ``late decline'' 
profile than to the ``late spike'' profile,
despite both profiles reaching the same terminal BMI.

More details about the model selection in the simulation can be found 
in Supplementary Section \ref{sec:supp_model_selection}.
The validity of regularized empirical Fisher matrix is checked empirically 
as shown in Supplementary Section \ref{sec:supp_stationarity}.

\section{Application to cardiometabolic health and cognitive decline}
\label{sec:application}

\subsection{The 3C cohort}
\label{sec:3c_data}

The Trois-Cit\'{e}s (3C) study is a population-based cohort
initiated in 1999, enrolling $9{,}294$ participants aged
$\geq 65$~years from three French cities (Bordeaux, Dijon,
Montpellier) and followed for up to 12~years with clinical visits
every 2--3~years \citep{3CStudyGroup2003}. The 3C Study was approved 
by the ethics committee of Kremlin-Bicêtre University Hospital, 
Paris, and all participants gave written informed consent. Cognitive function was assessed at each visit by
the Isaacs Set Test truncated at 15~seconds (IST), a timed verbal
fluency task. We considered $K = 5$ time-varying cardiometabolic
covariates: BMI and blood pressure (systolic and diastolic),
measured at every visit, and fasting glucose and HDL cholesterol,
measured only at baseline, the 4-year, and the 10-year visits; all
five covariates could additionally be missing at any given visit.
Time-independent covariates ($K_s = 3$) were baseline age $\mathrm{AGE}_i$ (centered
at the sample mean), sex, and 3 educational
level. Missing time-varying covariates were handled in our method by
within-subject linear interpolation, with
last-observation-carried-forward at the boundaries.

\begin{figure}[!t]
\centering
\includegraphics[width=1\textwidth]{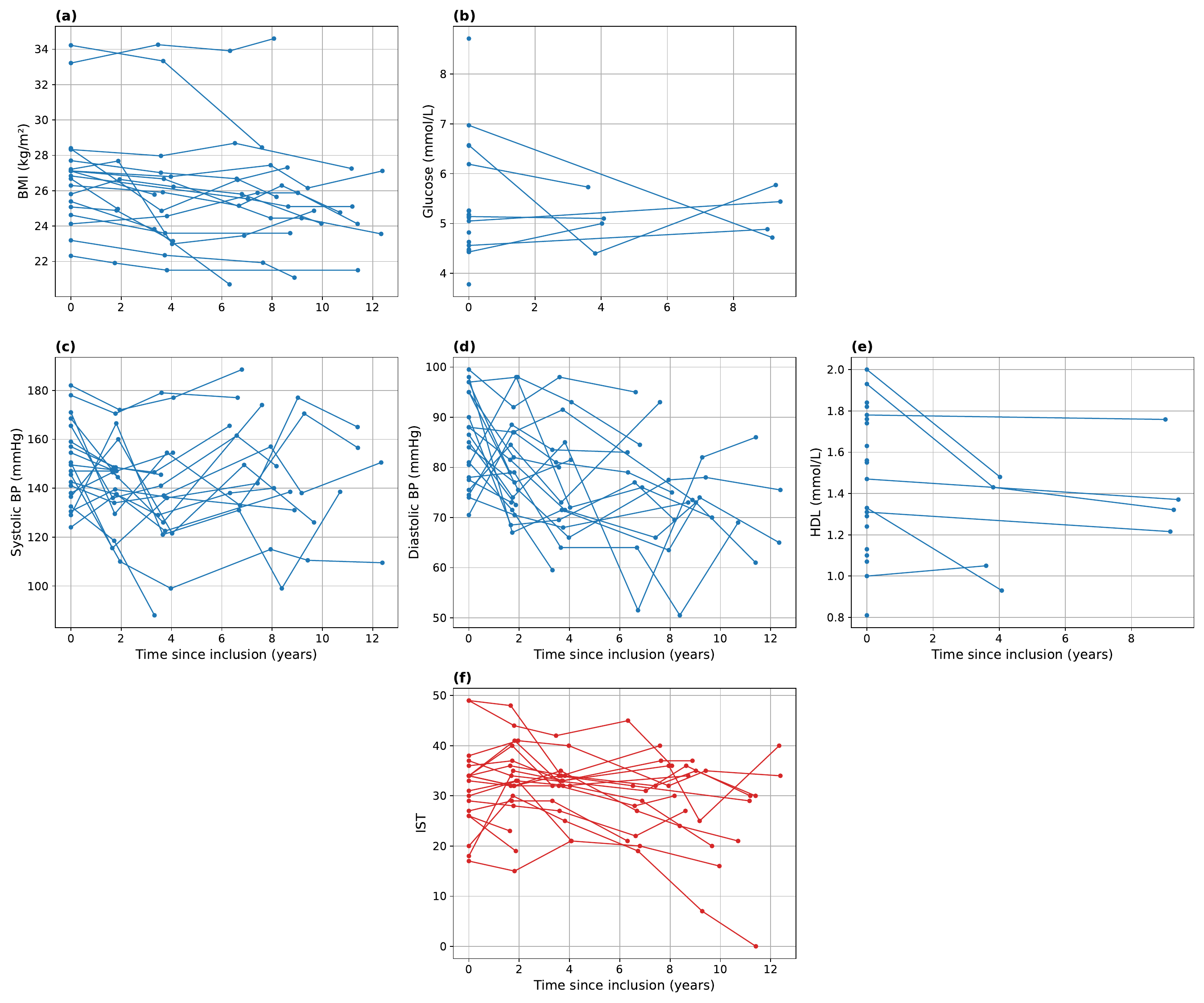}
\caption{Observed individual trajectories for a random subset of 50 participants in the 3C cohort. Panels (a)--(e) (blue): time-varying exposures; panel (f) (red): cognitive outcome. BMI = body mass index (kg/m$^2$); PAS = systolic blood pressure (mmHg); PAD = diastolic blood pressure (mmHg); 
HDL = high-density lipoprotein cholesterol (mmol/L); 
IST = Isaacs Set Test score. Time is measured in years since 
cohort inclusion.}
\label{fig:real_data}
% {\footnotesize\baselineskip=12pt\raggedright\noindent 
% \textbf{Alt Text}: Graphs
% showing individual participant trajectories over time since inclusion in years
% for a random subset of fifty people in the 3C cohort. Panels a to e,
% in blue, show time-varying exposures: panel a, body mass index; panel b,
% glucose; panel c, systolic blood pressure; panel d, diastolic blood pressure;
% panel e, high-density lipoprotein cholesterol. Panel f, in red, shows the
% cognitive outcome, the Isaacs Set Test score. Each panel plots many
% overlapping lines, one per participant, densely clustered at inclusion and
% spreading out over follow-up. \par}
\end{figure}

After excluding participants with missing baseline confounders
($n = 18$), missing baseline cardiometabolic measurements
($n = 1{,}151$), no follow-up visit ($n = 636$), or prevalent
dementia ($n = 165$), the analytical sample comprised
$N = 7{,}324$ participants with 2--7~visits. Twenty percent
($n = 1{,}465$) was held out for external validation; all models
were trained on the remaining $5{,}859$ participants (with internal split of $4{,}685$ for training and $1{,}174$ for validation). Figure~\ref{fig:real_data} 
displays the observed trajectories of time-varying covariates 
(BMI, fasting glucose, HDL cholesterol, and blood pressure) for a random subset of participants.
Full cohort details and covariate availability across visits are reported in Supplementary Table
\ref{tab:supp_covariates}.

\subsection{LMM specification}
\label{sec:hlme}

As a benchmark we fitted a series of classical linear mixed-effects
models using the \texttt{lcmm} package in R
\citep{proustlima2017lcmm}. All LMMs shared a common random-effect
structure (random intercept and random slopes on a natural cubic
spline of time with 2~d.f.) and approximated the fixed-effect
trajectory with a natural cubic spline of time (3~d.f.). We considered five nested
specifications for the fixed-effect mean, progressively adding the
time-varying covariates and their interactions with time
through products of the spline basis with $\mathrm{AGE}_i$ and/or
BMI (Supplementary Section  \ref{sec:supp_hlme}). The best model,
including a time~$\times$~$\mathrm{AGE}_i$ interaction, was selected by BIC
(Supplementary Table~\ref{tab:supp_bic}).

\subsection{Neural ODE-LMM specification}
\label{sec:ode_3c}

The ODE vector field takes as input the $K = 5$ time-varying
covariates and their binary observation masks. Static covariates entered through both the encoder network
and the skip connection. A group lasso penalty on the first-layer
decoder weights (Section~\ref{sec:model}) regulated the dynamic
covariate skip pathway, encouraging time-varying covariates to be
routed through the ODE when the cumulative representation improved
the fit. The final architecture and regularisation configuration
(latent dimension, regularisation mode, and decoder capacity) were
selected by grid search on the held-out validation set
(Supplementary Section \ref{sec:supp_cv}; Supplementary
Tables~\ref{tab:supp_cv}).
Architecture details and hyperparameters are given in
Supplementary Table~\ref{tab:supp_architecture}.
The model was trained with Adam optimizer (learning rate $10^{-3}$, weight decay $10^{-5}$),
a batch size of 128 subjects, and early stopping with a patience of 300 epochs
without the improvement of NLL on the validation data.

\subsection{Results}
\label{sec:results}

\subsubsection{Goodness of fit}

\begin{figure}[!t]
\centering
\includegraphics[width=1\textwidth]{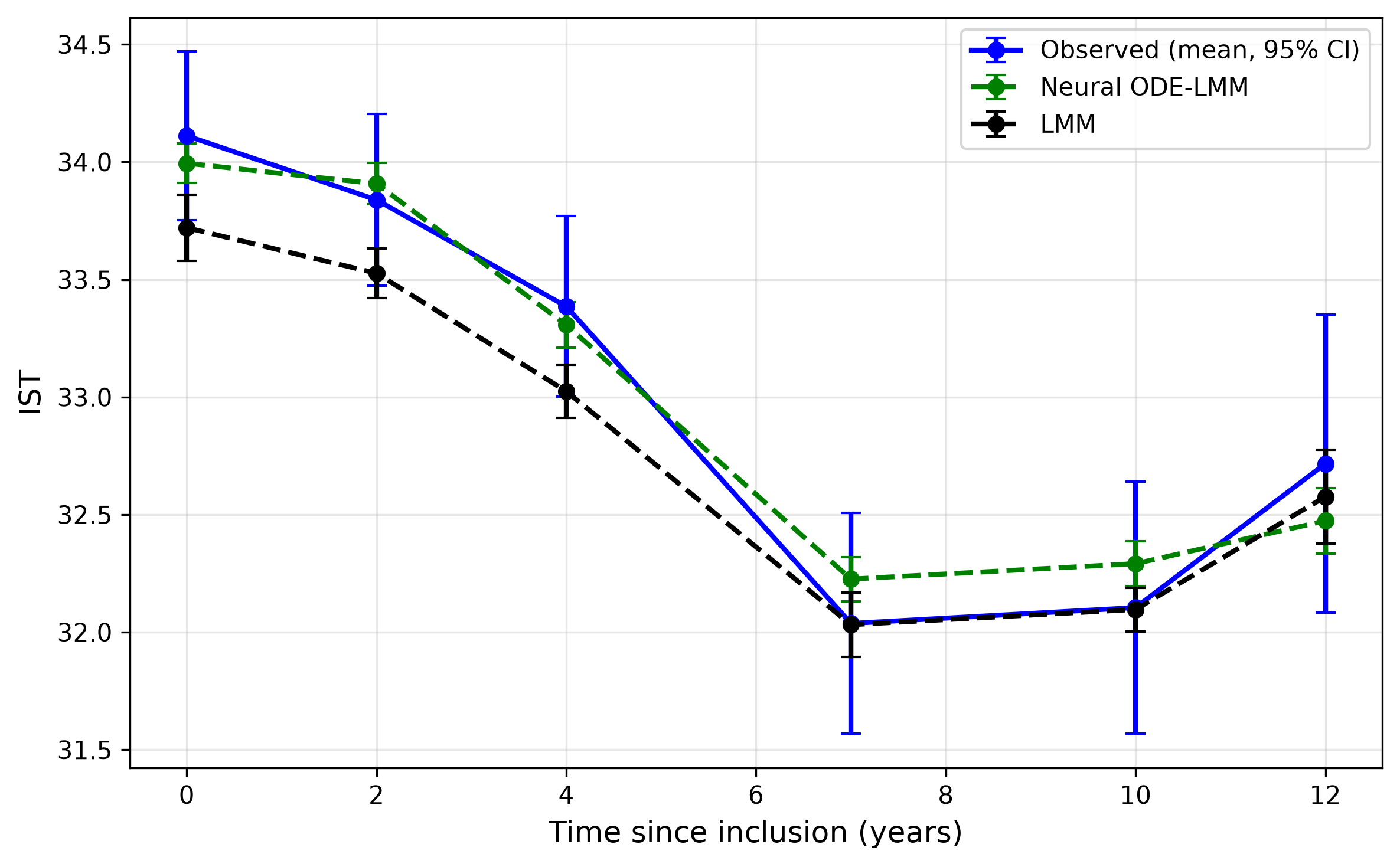}
\caption{Population-averaged predictions in fit mode on the 3C cohort. 
Blue: observed IST mean $\pm$ 95\% CI. 
Green dashed: Neural ODE-LMM predictions with 95\% CI.
Black dashed: classical LMM predictions with 95\% CI.}
\label{fig:pop_avg}
% {\footnotesize\baselineskip=12pt\raggedright\noindent 
% \textbf{Alt Text}: Line plot with
% time since inclusion in years, from zero to twelve, on the horizontal axis and
% cognitive score on the vertical axis. Three curves with ninety-five percent
% confidence interval bars are shown: observed mean, Neural ODE linear mixed
% model, and classical linear mixed model. All three decline from about
% thirty-four at inclusion to a minimum near thirty-two around year seven, then
% rise slightly by year twelve. The Neural ODE curve tracks the observed mean
% closely throughout, while the classical linear mixed model sits below both in
% the early years and fits the data closely between years seven and
% ten.\par}
\end{figure}

Table~\ref{tab:prediction_summary} compares prediction performance of the two models on the $5{,}859$ training participants.
The Neural ODE-LMM achieves a slightly higher marginal log-likelihood on both training and test data, indicating a better fit to the observations, and comparable MSE at the subject level.
Figure~\ref{fig:pop_avg} displays the population-averaged BLUP
predictions in fit mode on the test set, with pointwise 95\%
confidence intervals.
The observed mean IST (blue) declines from approximately 34.1 at
baseline to 32.0 at year~7, with a slight increase at years 10--12
likely reflecting survivorship bias.
The Neural ODE-LMM (green) tracks the observed mean closely
throughout, while the classical LMM (black) underestimates early time points.
Confidence intervals for both models are narrower than those of
the observed mean.
Individual predictions for randomly selected subjects are shown in 
Supplementary Figures~\ref{fig:supp_individual_fit}--\ref{fig:supp_individual_forecast}.

\subsubsection{Interpretation}
\label{sec:bmi_results}
\begin{figure}[!t]
\centering
\begin{subfigure}[t]{0.48\textwidth}
  \includegraphics[width=\textwidth]{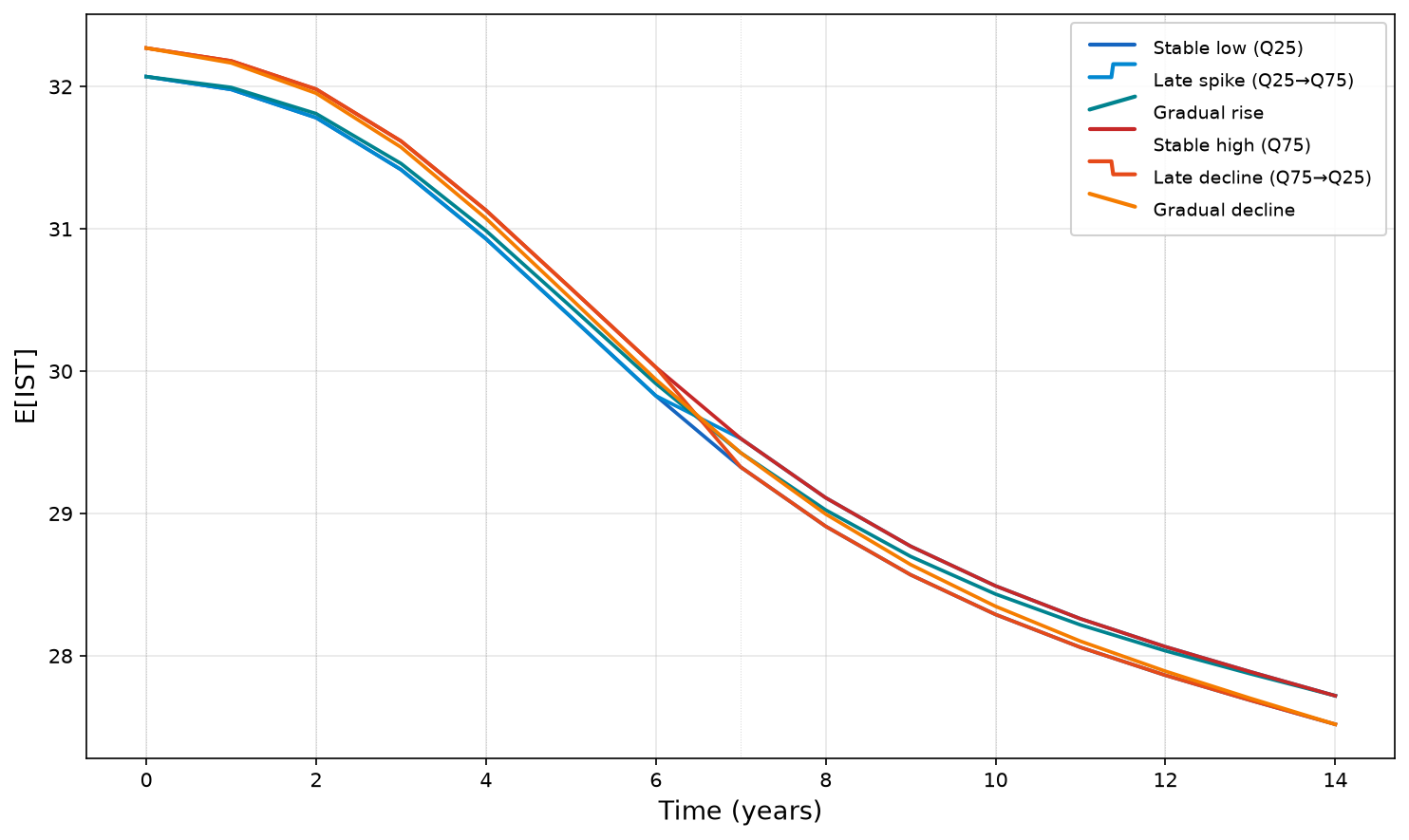}
  \caption{LMM --- BMI}
  \label{fig:profile_pdp_bmi_lmm}
\end{subfigure}
\hfill
\begin{subfigure}[t]{0.48\textwidth}
  \includegraphics[width=\textwidth]{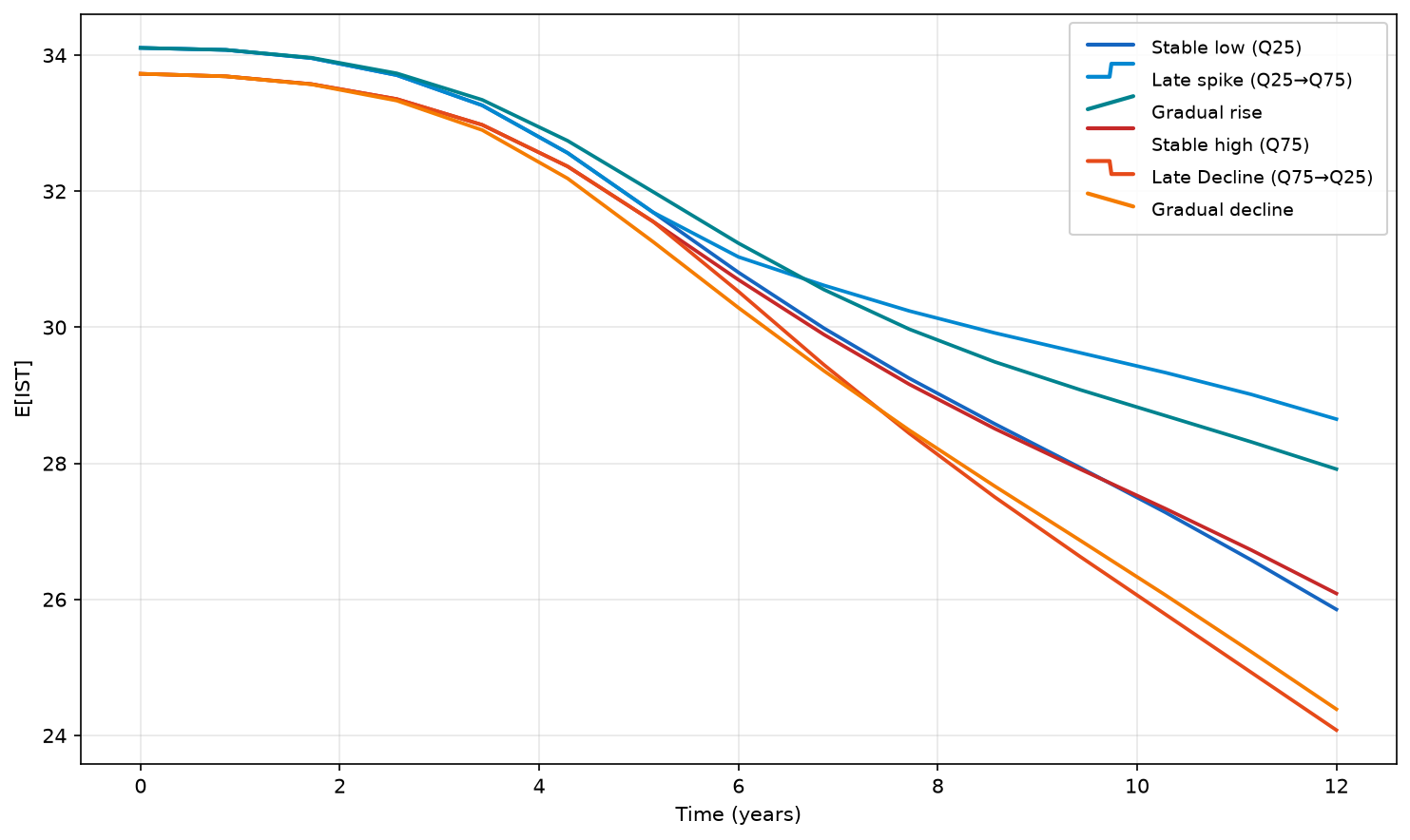}
  \caption{Neural ODE-LMM --- BMI}
  \label{fig:profile_pdp_bmi_ode}
\end{subfigure}

\begin{subfigure}[t]{0.48\textwidth}
  \includegraphics[width=\textwidth]{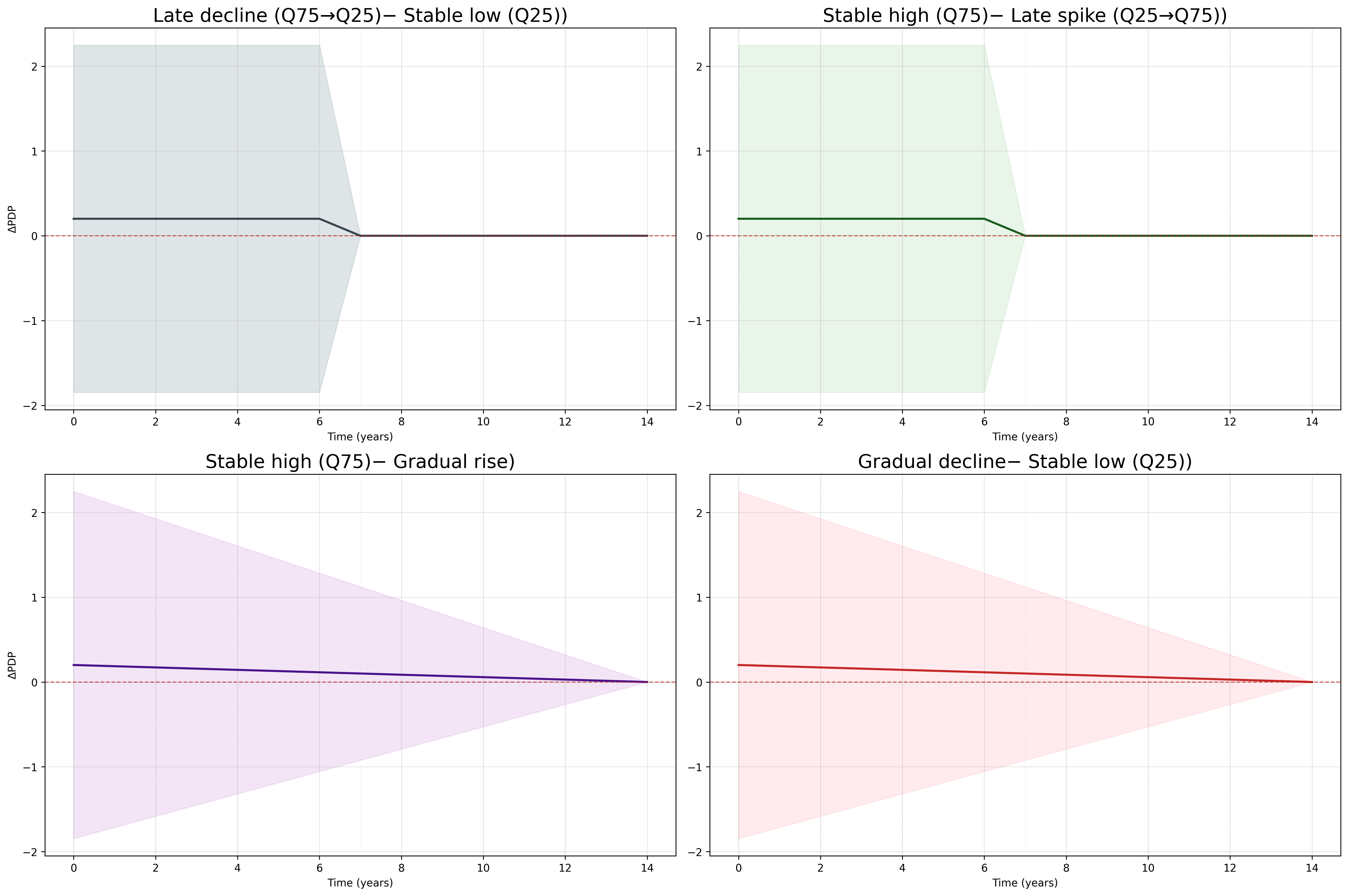}
  \caption{LMM --- BMI $\dPDP$}
  \label{fig:profile_deltapdp_lmm}
\end{subfigure}
\hfill
\begin{subfigure}[t]{0.48\textwidth}
  \includegraphics[width=\textwidth]{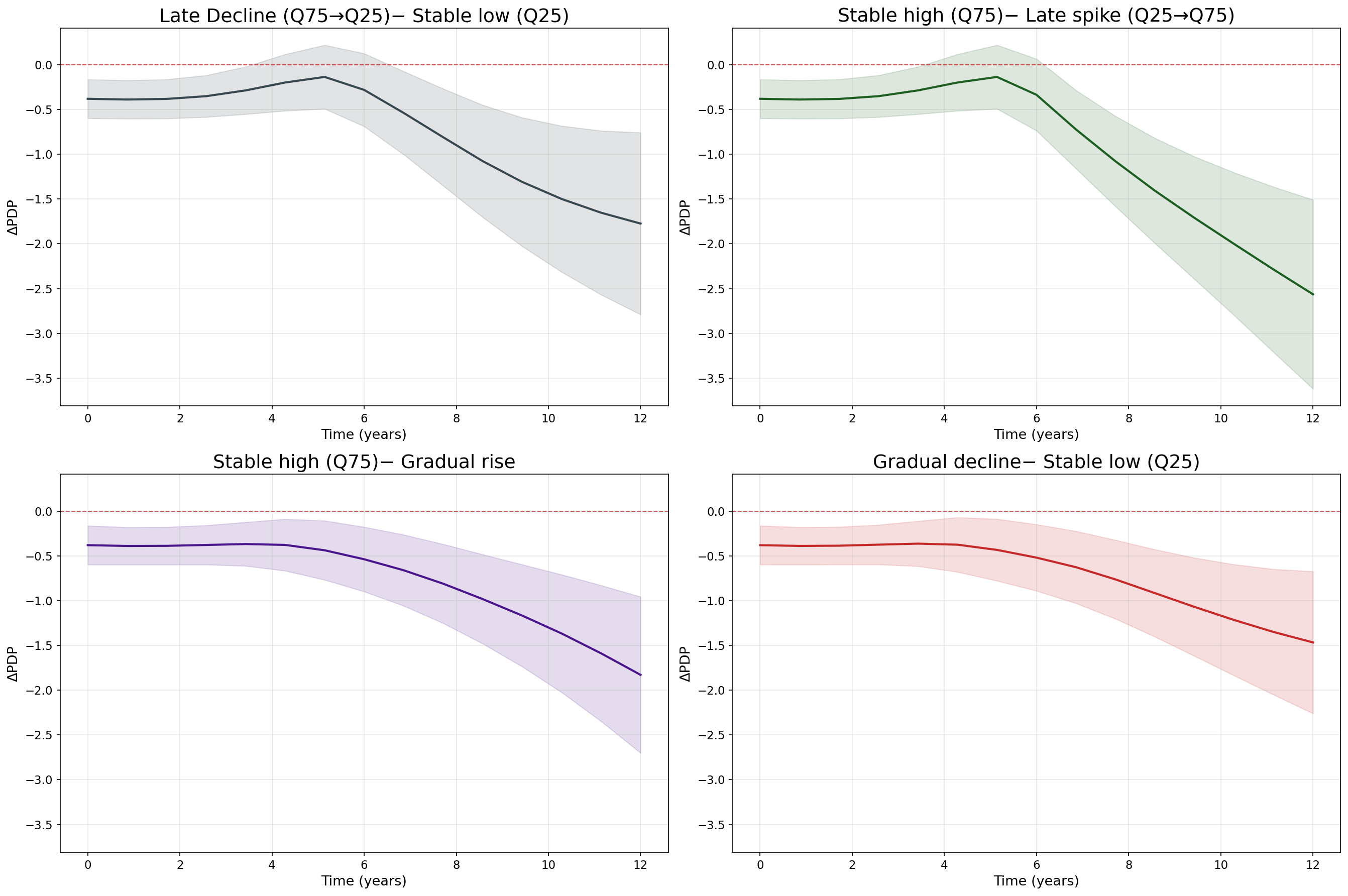}
  \caption{Neural ODE-LMM --- BMI $\dPDP$}
  \label{fig:profile_deltapdp_bmi_ode}
\end{subfigure}

\caption{Counterfactual trajectory-profile predictions (top) and pairwise $\dPDP$ with
pointwise delta-method 95\% confidence intervals (bottom) for BMI.
Left: LMM, where $\dPDP$ contrasts between crossing profiles flip sign
at the midpoint, confirming that the model responds only to the
instantaneous BMI value.
Right: Neural ODE-LMM, where all four contrasts become increasingly
negative over time; profiles with declining or sustained-high BMI
are associated with steeper cognitive decline than profiles with
rising or sustained-low BMI. The confidence intervals exclude
zero at later follow-up times, indicating that the ODE encoder has
learned trajectory-dependent associations.}

% {\footnotesize\baselineskip=12pt\raggedright\noindent
% \textbf{Alt Text}: Four panels in two rows. Top row shows
% counterfactual predicted cognitive trajectories over follow-up time for
% six body mass index profiles. Bottom row shows pairwise differences with
% pointwise confidence intervals. Left column, linear mixed model: the
% differences cross and change sign midway through follow-up. Right column,
% neural ordinary differential equation model: all differences grow steadily
% more negative and separate from zero at later times. \par}
\label{fig:profile_pdp}
\end{figure}

To examine whether cardiometabolic risk factors exert trajectory-dependent 
effects on cognitive decline, we computed counterfactual predictions under trajectory profiles for BMI, fasting glucose, and HDL cholesterol.
For each covariate, the six counterfactual profiles were defined according to the 25th and 75th percentiles of the covariate distribution (stable low, stable high, late spike Q25$\to$Q75, late decline Q75$\to$Q25, gradual rise, and gradual decline).
We concentrate in the main document on BMI, presenting the trajectory-profile counterfactual predictions (Figure~\ref{fig:profile_pdp}, top) and the pairwise $\dPDP$ with pointwise 95\% delta-method confidence intervals (Figure~\ref{fig:profile_pdp}, bottom).
Analogous figures and interpretations for glucose and HDL are given in Supplementary 
\ref{fig:supp_gluc_pdp}---\ref{fig:supp_hdl_delta}.

The Neural ODE-LMM reveals a trajectory-dependent association between BMI trajectory shape and cognition (Figure~\ref{fig:profile_pdp}, \subref{fig:profile_pdp_bmi_ode}).
Subjects with gradually rising BMI trajectories (Q25$\to$Q75 over follow-up) maintain the highest predicted cognitive scores,
while those with declining BMI (late decline Q75$\to$Q25, or gradual decline) experience the steepest cognitive decline.
The late-decline profile produces worse outcomes than the late-spike profile at later visits,
despite both profiles crossing the same BMI value at the midpoint:
a declining BMI is associated with a worse cognitive outcome, while, although high initial BMI is associated with worse cognition early on, a rising BMI trajectory appears relatively protective over time.
The LMM predicts trajectories that swap at the midpoint (Figure~\ref{fig:profile_pdp}, \subref{fig:profile_pdp_bmi_lmm}): the ``late decline'' profile (Q75$\to$Q25) drops to match the ``stable low'' profile after the crossover, while ``late spike'' (Q25$\to$Q75) rises to match ``stable high.''
The $\dPDP$ between late decline and late spike flips sign at the midpoint, 
because the LMM responds only to the current covariate
value, whichever profile has the higher BMI at time $t$ receives the
worse predicted outcome, regardless of its earlier history.
By contrast, the Neural ODE-LMM (Figure~\ref{fig:profile_pdp}, \subref{fig:profile_pdp_bmi_ode}) 
produces a persistent negative $\dPDP$ after the crossover, 
confirming that the ODE encoder has learned trajectory-dependent 
associations that the LMM cannot represent. 
More analyses can be found in Supplementary Section \ref{sec:supp_zeroskip}---\ref{sec:supp_latent}.

\section{Discussion}
\label{sec:discussion}

We proposed the Neural ODE-LMM, which embeds a Neural ODE within the
linear mixed-effects framework, thereby preserving its good statistical
properties.
A learned vector field encodes covariate trajectories into a
continuous-time latent state that drives the fixed- and random-effect
designs.
To extract interpretable association measures from the fitted model, we
introduced the counterfactual prediction contrast $\dPDP$, 
which compares population-averaged predictions under alternative
covariate paths in the spirit of PDP \citep{friedman2001greedy}, 
and whose variance is estimated via the delta
method applied to the inverse empirical Fisher information stabilised
by Ledoit--Wolf shrinkage.
% the first-order stationarity condition
% $\sum_i U_i \approx 0$ serves as a practical diagnostic that the
% penalised score is well-behaved at the optimum.

In simulations, the Neural ODE-LMM sucessfully retrieved with negligible bias and 
well calibrated confidence intervals both a classical instantaneous 
covariate association (S1) and a cumulative effect (S2), 
illustrating its ability to learn the pattern of an association.
Applied to the 3C cohort, the $\dPDP$ detected a
trajectory-dependent association between BMI trajectory shape and cognitive
decline, while revealing qualitatively distinct temporal mechanisms for
fasting glucose and HDL cholesterol.
The cumulative BMI effect recovered by the Neural ODE-LMM is
consistent with evidence that, in elderly populations, BMI decline is a
marker of preclinical neurodegeneration rather than a sign of improved
health.
In the same 3C cohort, \citet{wagner2018} found that BMI declined
significantly more in prodromal dementia cases than in controls
($p < 0.001$ for group-by-time interaction), with trajectories diverging
approximately 7~years before diagnosis. Using a weighted cumulative index of exposure 
technique, the same authors \citep{Wagner2021} also estimated a time-dependent 
association of BMI with subsequent 
cognition in the Nurses' Health Study: 
higher level of BMI measured long before the cognitive assessment 
was associated with lower cognition while higher BMI 
measured at the approach of the assessment was associated with higher cognition. 
Our trajectory-profile analysis recapitulates this time-dependent and cumulative 
association pattern without
pre-specifying case-control status or a parametric form for the
trajectory effect and provides formal inference via the delta-method
confidence intervals on the $\dPDP$.

This work has limitations. First, the model treats covariates as measured 
without error, conditioning on values obtained by within-subject 
linear interpolation between irregularly spaced visits; 
this error, unlike in classical regression, propagates through 
the ODE integration path and may accumulate over long inter-visit intervals. 
Jointly modelling the outcome and covariate processes 
as smooth latent trajectories with proper uncertainty would address this, 
but requires further development. 
Second, the delta-method variance via empirical Fisher information was 
demonstrated for parsimonious architectures 
(fewer than 2{,}000 parameters); 
the Fisher matrix becomes computationally 
prohibitive and the outer-product estimator may deteriorate as dimension grows. 
Scaling frequentist uncertainty quantification to larger networks 
through low-rank approximations of the
Fisher matrix or alternative strategies
remains open. More broadly, the delta-method variance relies on M-estimator 
asymptotics whose regularity conditions, 
notably identifiability and local convexity, 
are not formally guaranteed for overparameterised networks; 
nevertheless, the simulations show near-nominal coverage (91--98\% across scenarios), 
supporting the approximation in this setting.

\section{Data and code availability}
\label{sec:software}
Anonymized Three-City Study data may be shared upon reasonably justified
request to the 3C scientific committee
(\href{mailto:e3C.CoordinatingCenter@u-bordeaux.fr}
{e3C.CoordinatingCenter@u-bordeaux.fr}).
Scripts replicating the application and simulations, with documentation, are
available at \url{https://github.com/AuroraFr/NodeLmm}. Models were trained
using PyTorch 2.6.0 and Python 3.10+ on a dual-socket Intel Cascade Lake node
(2 $\times$ 18 cores, 192 GB RAM) without GPU acceleration; the sequential
nature of ODE integration offered no benefit from GPU offloading at this scale.
Training a single Neural ODE-LMM on the 3C cohort took approximately 30 minutes,
and each simulation replicate 30--40 minutes; the model can be trained on a
standard laptop. Classical LMM comparison models were fitted in R using the
\texttt{lcmm} package \citep{proustlima2017lcmm}.

\bibliographystyle{biorefs}
\bibliography{neuralode_refs}

% ══════════════════════════════════════════════════════════════
% TABLES AND FIGURES
% ══════════════════════════════════════════════════════════════
\begin{table}[!p]
\centering
\caption{$\dPDP$ performance across $R = 100$ replications for Scenario~S1
(instantaneous BMI $\times$ age interaction) and Scenario~S2 (cumulative BMI burden),
evaluated at a grid of times~$t$ for the contrast between $v_{\mathrm{lo}} = 23$
and $v_{\mathrm{hi}} = 28$~kg/m$^2$. Bias, empirical variance ($\mathrm{Var}_{\mathrm{emp}}$),
mean estimated variances from the delta method
($\overline{\mathrm{Var}}_{\mathrm{est}}$) and coverage of the nominal 95\%
confidence interval based on estimated variance.}
\label{tab:sim_results}
\small
\begin{tabular}{@{}llrrrrr@{}}
\hline\\[-9pt]
Scenario & $t$ & $\dPDP_0$ & Bias & $\mathrm{Var}_{\mathrm{emp}}$ & $\overline{\mathrm{Var}}_{\mathrm{est}}$ & Cov. \\
\hline\\[-9.75pt]
\multicolumn{7}{@{}l}{\emph{S1 — true LMM}} \\[2pt]
 & 0  & $-1.725$ & $-0.038$ & 0.009 & 0.007 & 0.94 \\
 & 4  & $-1.725$ & $-0.038$ & 0.009 & 0.007 & 0.94 \\
 & 8  & $-1.725$ & $-0.038$ & 0.009 & 0.007 & 0.94 \\
 & 10 & $-1.725$ & $-0.038$ & 0.009 & 0.007 & 0.94 \\[3pt]
\multicolumn{7}{@{}l}{\emph{S1 — Neural ODE-LMM}} \\[2pt]
 & 0  & $-1.725$ & $-0.053$ & 0.015 & 0.011 & 0.92 \\
 & 4  & $-1.725$ & $-0.078$ & 0.017 & 0.013 & 0.91 \\
 & 8  & $-1.725$ & $-0.085$ & 0.022 & 0.018 & 0.91 \\
 & 10 & $-1.725$ & $-0.101$ & 0.018 & 0.020 & 0.92 \\[3pt]
\multicolumn{7}{@{}l}{\emph{S2 — Neural ODE-LMM}} \\[2pt]
 & 2  & $-0.500$ & $-0.027$ & 0.008 & 0.012 & 0.98 \\
 & 4  & $-1.000$ & $-0.028$ & 0.012 & 0.015 & 0.96 \\
 & 8  & $-2.000$ & $-0.061$ & 0.024 & 0.024 & 0.93 \\
 & 10 & $-2.500$ & $-0.096$ & 0.033 & 0.033 & 0.92 \\
\hline
\end{tabular}
\end{table}

\begin{table}[!pt]
\centering
\caption{Prediction performance on the 3C cohort. 
MSE is computed on the IST. 
The LMM column reports results for the best model by BIC.}
\label{tab:prediction_summary}
\small
\begin{tabular}{@{}l rr rr@{}}
\hline\\[-9pt]
& \multicolumn{2}{c}{LMM} & \multicolumn{2}{c}{Neural ODE-LMM} \\
\cline{2-3} \cline{4-5}\\[-9pt]
Metric & Train & Test & Train & Test \\
\hline\\[-9.75pt]
Log-likelihood   & $-63\,565$ & $-20\,200$ & $-63\,286$ & $-19\,993$ \\[3pt]
MSE(pop)         & 44.81 & 45.67 & 43.92 & 44.99 \\
MSE(fit mode)        & 9.11 & 9.32 & 9.27 & 9.62 \\
MSE(forecast mode)        & 26.20 & 28.81 & 25.67 & 26.57 \\
\hline
\end{tabular}
\end{table}

% ── Figures ───────────────────────────────────────────────────
%% ── Architecture figure ──────────────────────────────────────
%% ── End figure ───────────────────────────────────────────────


\clearpage 
\appendix
\section*{Supplementary Materials}
\setcounter{table}{0}
\setcounter{figure}{0}
\renewcommand{\thetable}{S\arabic{table}}
\renewcommand{\thefigure}{S\arabic{figure}}
\renewcommand{\thesection}{S\arabic{section}}

% ══════════════════════════════════════════════════════════════
\section{Simulation data-generating parameters}
\label{sec:supp_sim_params}
% ══════════════════════════════════════════════════════════════

All simulated datasets are calibrated to the 3C cohort so that the
marginal distributions, visit schedule, and variance components
resemble those of the real data.
\paragraph{Reference model.}
A linear mixed-effects model is first fitted to the 3C analytic sample
using the \texttt{hlme()} function of the \texttt{lcmm} package
in~R.
The outcome is the IST score; the fixed-effect specification includes
a natural-cubic-spline basis of follow-up time (3~d.f.), baseline age
(centred), sex, education level, and BMI as a time-varying covariate.
Random effects are placed on an intercept and a natural-spline basis
of time (2~d.f.), yielding a $3 \times 3$ unstructured covariance
matrix~$D$.
The estimated fixed-effect vector~$\hat\beta$, random-effect
covariance~$D_0$, and residual variance~$\sigma_0^2$ serve as
the reference parameters from which scenario-specific modifications
are derived.

\paragraph{Covariate generation.}
Each time-varying covariate is generated from its own auxiliary
linear mixed-effects model fitted to the 3C data.
Specifically, an auxiliary LMM is fitted for BMI with a linear
function of time as fixed effect and a random intercept and slope,
together with baseline covariates (centered age, sex, education level).
The fitted fixed effects, random-effect covariance, and residual
variance define the marginal model from which new covariate
trajectories are drawn.
For each simulated subject, a random effect is sampled from the
estimated covariance, the fixed-effect trajectory is computed from the
subject's baseline covariates, and i.i.d.\ Gaussian noise is added to
produce realistic measurement variability.

\paragraph{Visit schedule.}
Each simulated subject is assigned an adaptive time grid of
3--15 irregularly spaced visits spanning the same follow-up window as
the real data.
The number of visits per subject is aligned with the subject's observed
follow-up duration in the 3C cohort, so that subjects with shorter
follow-up receive fewer visits.
Natural-cubic-spline bases for time are evaluated at these simulated
visit times using the same internal and boundary knots as in the
reference model, ensuring that the spline bases in simulated data are
exactly comparable to those in the 3C analysis.

\paragraph{Outcome generation (common structure).}
Given a simulated dataset with covariates and visit times, the
outcome for subject~$i$ at visit~$j$ is generated as
\begin{equation}\label{eq:sim_dgp}
  Y_{ij} = X_{ij}^\top \beta^\ast + f_s(X_i^{(t \le t_{ij})})
  + Z_{ij}^\top b_i + \varepsilon_{ij},
\end{equation}
where $\beta^\ast$ is a scenario-specific modification of the
reference~$\hat\beta$,
$f_s(\cdot)$ is a scenario-specific function of the covariate history
(identically zero in the instantaneous scenario),
$b_i \sim \mathcal{N}(0, D_0)$,
and $\varepsilon_{ij} \sim \mathcal{N}(0, \sigma_0^2)$.
The random effects and residual variance are held at the reference
estimates across all scenarios, so that differences in performance
are driven entirely by the fixed-effect structure.

\paragraph{Scenario~S1 (instantaneous effect).}
The covariate effect is purely instantaneous and linear:
$f_s \equiv 0$ and $\beta^\ast$ includes a non-zero coefficient for
BMI and a BMI\,$\times$\,centred-age interaction.
The fixed-effect mean is thus
\[
  \mu_{ij} = \beta_0^\ast
  + \sum_{k=1}^{3}\beta_k^\ast\,\mathrm{ns}_k(t_{ij})
  + \beta_{\mathrm{BMI}}^\ast\,\mathrm{BMI}_{ij}
  + \beta_{\mathrm{age}}^\ast\,\mathrm{AGE}_{c,i}
  + \beta_{\mathrm{sex}}^\ast\,\mathrm{sex}_i
  + \beta_{\mathrm{educ}}^\ast\,\mathrm{educ}_i
  + \beta_{\mathrm{int}}^\ast\,\mathrm{BMI}_{ij} \cdot \mathrm{AGE}_{c,i}.
\]
The true $\dPDP$ for a contrast
$\Delta v = v_{\mathrm{hi}} - v_{\mathrm{lo}}$ is
$\Delta v \cdot (\beta_{\mathrm{BMI}}^\ast +
\bar{a}\,\beta_{\mathrm{int}}^\ast)$ at every evaluation time, where
$\bar{a}$ is the population mean of centred age.
The LMM is fitted with the oracle specification
(BMI\,$\times$\,age interaction) as well as a general specification
that replaces the BMI\,$\times$\,age term with
spline\,$\times$\,age interactions, testing robustness to mild
misspecification.

\begin{table}[t]
  \centering
  \caption{Parameter values for the data-generating model in
    Scenario~S1. All values were calibrated from a preliminary LMM
    fit to the 3C data, except those marked with~$\dagger$, which
    were set to predefined values.}
  \label{tab:sim_params}
  \small
  \begin{tabular}{@{}llr@{}}
    \toprule
    \textbf{Component} & \textbf{Parameter} & \textbf{Value} \\
    \midrule
    \multicolumn{3}{@{}l}{\textit{Fixed effects $(\beta^\ast)$}} \\[2pt]
    Intercept
      & $\beta^\ast_0$                              &  $30.71$  \\
    Natural spline for time (3 d.f.)
      & $\beta^\ast_1$                              &  $-2.82$  \\
      & $\beta^\ast_2$                              &  $0.02$   \\
      & $\beta^\ast_3$                              &  $-3.13$  \\
    BMI$^\dagger$
      & $\beta^\ast_{\mathrm{BMI}}$                 &  $-0.30$  \\
    Centred age
      & $\beta^\ast_{\mathrm{age}}$                 &  $-0.51$  \\
    Sex (female)
      & $\beta^\ast_{\mathrm{sex}}$                 &  $1.85$   \\
    Education (medium)
      & $\beta^\ast_{\mathrm{educ},2}$              &  $2.67$   \\
    Education (high)
      & $\beta^\ast_{\mathrm{educ},3}$              &  $3.32$   \\
    BMI $\times$ centred age$^\dagger$
      & $\beta^\ast_{\mathrm{BMI \times age}}$      &  $-0.05$  \\
    \midrule
    \multicolumn{3}{@{}l}{\textit{Random effects $(b_i \sim \mathcal{N}(0, D_0))$}} \\[2pt]
    \multirow{3}{*}{$D_0$}
      & $D_{0,11}$  &  $28.95$  \\
      & $D_{0,22}$  &  $44.70$  \\
      & $D_{0,33}$  &  $27.59$  \\
      & $D_{0,12}$  &  $-5.95$  \\
      & $D_{0,13}$  &  $-1.31$  \\
      & $D_{0,23}$  &  $26.01$  \\
    \midrule
    \multicolumn{3}{@{}l}{\textit{Residual}} \\[2pt]
    Standard deviation
      & $\sigma_0$  & $3.63$    \\
    \bottomrule
  \end{tabular}
\end{table}

\paragraph{Scenario~S2 (cumulative effect).}
The cumulative effect enters through
$
  f_s(X_i^{(t \le t_{ij})})
  = \alpha \int_0^{t_{ij}} \mathrm{BMI}_i(\tau)\,d\tau,
$
the time-integrated BMI exposure up to time~$t_{ij}$,
scaled by a coefficient~$\alpha < 0$ so that greater accumulated BMI
exposure is associated with worse cognitive scores.
The integral is evaluated numerically from the simulated visit-level
BMI values using the trapezoidal rule.
Because the cumulative exposure depends on the full trajectory history
and not on the current BMI value alone, it cannot be recovered by
any model that conditions only on the contemporaneous covariate.
Under constant-BMI interventions at level~$v$, the cumulative
component reduces to $\alpha\,v\,t$, so the oracle $\dPDP$ between two
constant profiles $v_{\mathrm{hi}}$ and $v_{\mathrm{lo}}$ is
$\Delta v \cdot (\beta_{\mathrm{BMI}}^\ast
  + \bar{a}\,\beta_{\mathrm{int}}^\ast + \alpha\,t)$,
which grows in magnitude with follow-up time---in contrast to
Scenario~S1, where the $\dPDP$ is time-invariant.

\paragraph{Replication and evaluation.}
For each scenario, $D = 100$ independent datasets are generated.
On each replicate, the HLME is fitted, and $\dPDP$ estimates are
computed for the contrast BMI\,$= 23$ versus BMI\,$= 28$ at
evaluation times $t \in \{0, 4, 8, 10\}$ years.
Delta-method standard errors are obtained from the fixed-effect
variance--covariance matrix returned by \texttt{lcmm}.
The simulation summary reports bias, empirical variance, mean
estimated variance and 95\% coverage rate of the delta-method
confidence intervals, aggregated over replicates and evaluation times.
% ══════════════════════════════════════════════════════════════
\section{Trois-Cit\'{e}s cohort: data description}
\label{sec:supp_data}
% ══════════════════════════════════════════════════════════════
Of the initial 9,294 participants, we excluded 18 with missing educational data, 1,151 with missing baseline cardiometabolic measurements, 636 without a follow-up visit, and 165 with prevalent dementia.
The final analytic sample comprised $n = 7{,}324$ participants.
Table~\ref{tab:supp_covariates} summarises the availability of time-varying covariates across follow-up visits.
\begin{sidewaystable}
  \centering
\caption{Time-varying cardiometabolic factors in the 3C cohort study
across visits. An ``---'' indicates the variable was not measured at
that visit. Glucose and HDL were measured at T4 in Dijon and Bordeaux
only, using centre-specific variable names.}
\label{tab:supp_covariates}
\small
\begin{tabular}{@{}lcccccc@{}}
\toprule
& \textbf{T0 (S0)} & \textbf{T2 (S1)} & \textbf{T4 (S2)}
& \textbf{T7 (S4)} & \textbf{T10 (S5)} & \textbf{T12 (S6)} \\
\midrule
BMI & \makecell{WEIGHT01,\\HEIGHT}
    & WEIGHT1 & WEIGHT2
    & WEIGHT4 & WEIGHT5 & WEIGHT6 \\[4pt]
SBP & \makecell{SBP0\_1\\SBP0\_2}
    & \makecell{SBP1\_1\\SBP1\_2}
    & \makecell{SBP2\_1\\SBP2\_2}
    & \makecell{SBP4\_1\\SBP4\_2}
    & \makecell{SBP5\_1\\SBP5\_2}
    & \makecell{SBP6\_1\\SBP6\_2} \\[4pt]
DBP & \makecell{DBP0\_1\\DBP0\_2}
    & \makecell{DBP1\_1\\DBP1\_2}
    & \makecell{DBP2\_1\\DBP2\_2}
    & \makecell{DBP4\_1\\DBP4\_2}
    & \makecell{DBP5\_1\\DBP5\_2}
    & \makecell{DBP6\_1\\DBP6\_2} \\[4pt]
Glucose & GLUC1 & ---
        & \makecell{GLUC2M (Dijon)\\GLUC\_S2 (Bordeaux)}
        & --- & GLUC5 & --- \\[4pt]
HDL & HDL1 & ---
    & \makecell{HDL2M (Dijon)\\HDL\_S2 (Bordeaux)}
    & --- & HDL5 & --- \\
\bottomrule
\end{tabular}
\end{sidewaystable}

% \begin{figure}[!t]
% \centering
% \includegraphics[width=1\textwidth]{LMC_figures/3C_subject_trajectories.pdf}
% \caption{Individual trajectories of cardiometabolic covariates and the IST 
% cognitive outcome over follow-up in the 3C cohort. 
% Each line represents one participant.}
% \label{fig:supp_spaghetti}
% \end{figure}

% ══════════════════════════════════════════════════════════════
\section{Classical LMM specification for 3C cohort}
\label{sec:supp_hlme}
% ══════════════════════════════════════════════════════════════

All five baseline LMMs share the observation model
\begin{equation*}
  Y_{ij} = \mu_{ij} + Z_{ij}^\top b_i + \varepsilon_{ij},
  \qquad b_i \sim \mathcal{N}(0, D), \quad \varepsilon_{ij} \sim \mathcal{N}(0, \sigma^2),
\end{equation*}
with a common random-effect design
$Z_{ij} = \bigl(1,\; \mathrm{ns}_1(t_{ij}),\; \mathrm{ns}_2(t_{ij})\bigr)^\top$
(random intercept $+$ random slopes on a natural-spline basis of time with 2~d.f.),
$D \in \mathbb{R}^{3 \times 3}$,
and i.i.d.\ residuals.
Here $\mathrm{ns}_k(t)$ denotes the $k$-th basis function of a natural cubic spline of time (boundary and internal knots placed at the empirical quantiles of follow-up time in the 3C cohort).

The five models differ only in the fixed-effect mean $\mu_{ij}$:

\paragraph{TV1} (demographics only).
\begin{equation*}
  \mu_{ij} = \beta_0 + \sum_{k=1}^{3} \beta_k\, \mathrm{ns}_k(t_{ij})
  + \beta_{\mathrm{age}}\, \mathrm{AGE}_{i}
  + \beta_{\mathrm{sex}}\, \mathrm{sex}_i
  + \beta_{\mathrm{educ},1}\, \mathrm{educ}_{i,1}
  + \beta_{\mathrm{educ},2}\, \mathrm{educ}_{i,2}.
\end{equation*}

\paragraph{TV2} (time-varying covariates).
\begin{equation*}
  \mu_{ij} = \mu_{ij}^{\mathrm{TV1}}
  + \beta_{\mathrm{BMI}}\, \mathrm{BMI}_{ij}
  + \beta_{\mathrm{SBP}}\, \mathrm{SBP}_{ij}
  + \beta_{\mathrm{DBP}}\, \mathrm{DBP}_{ij}
  + \beta_{\mathrm{gluc}}\, \mathrm{GLUC}_{ij}
  + \beta_{\mathrm{HDL}}\, \mathrm{HDL}_{ij}.
\end{equation*}

\paragraph{TV3} (time $\times$ AGE interaction).
\begin{equation*}
  \mu_{ij} = \mu_{ij}^{\mathrm{TV2}}
  + \sum_{k=1}^{3} \gamma_k\, \mathrm{ns}_k(t_{ij}) \cdot \mathrm{AGE}_{i}.
\end{equation*}

\paragraph{TV4} (time $\times$ BMI interaction).
\begin{equation*}
  \mu_{ij} = \mu_{ij}^{\mathrm{TV2}}
  + \sum_{k=1}^{3} \delta_k\, \mathrm{ns}_k(t_{ij}) \cdot \mathrm{BMI}_{ij}.
\end{equation*}

\paragraph{TV5} (both interactions).
\begin{equation*}
  \mu_{ij} = \mu_{ij}^{\mathrm{TV2}}
  + \sum_{k=1}^{3} \gamma_k\, \mathrm{ns}_k(t_{ij}) \cdot \mathrm{AGE}_{i}
  + \sum_{k=1}^{3} \delta_k\, \mathrm{ns}_k(t_{ij}) \cdot \mathrm{BMI}_{ij}.
\end{equation*}
All models were fitted using the \texttt{hlme()} function of the
\texttt{lcmm} package in~R.
Model~TV3 was selected by BIC (Table~\ref{tab:supp_bic}).

\begin{table}[!t]
\centering
\caption{BIC for the five classical LMM specifications fitted on the 3C cohort ($n = 5{,}859$). Lower is better.}
\label{tab:supp_bic}
\small
\begin{tabular}{@{}lccccc@{}}
\hline\\[-9pt]
 & TV1 & TV2 & TV3 & TV4 & TV5 \\
\hline\\[-9.75pt]
BIC & 127{,}498 & 127{,}502 & 127{,}325 & 127{,}527 & 127{,}350 \\
\hline
\end{tabular}
\end{table}
% ══════════════════════════════════════════════════════════════
\section{Neural ODE-LMM architecture and hyperparameters}
\label{sec:supp_architecture}
% ══════════════════════════════════════════════════════════════

Table~\ref{tab:supp_architecture} reports the full architecture
and hyperparameter configuration of the Neural ODE-LMM used in the
3C cohort analysis.
The configuration was selected via the two-tier model selection
procedure described in the main text (Section~5.4).

\begin{table}[!t]
\centering
\caption{Neural ODE-LMM architecture and training hyperparameters
for the 3C cohort analysis.}
\label{tab:supp_architecture}
\small
\begin{tabular}{@{}llc@{}}
\hline\\[-9pt]
Component & Hyperparameter & Value \\
\hline\\[-9.75pt]
\multicolumn{3}{@{}l}{\emph{Encoder}} \\[2pt]
 & Input dimension ($K + K_s + 1$)  & 10 \\
 & Hidden layers                     & 1 \\
 & Hidden units per layer            & 16 \\
 & Activation                        & SiLU \\
 & Output dimension ($d = \dim z$)   & 16 \\[3pt]
\multicolumn{3}{@{}l}{\emph{ODE vector field ($f_\phi$)}} \\[2pt]
 & Input dimension ($d + 2K + 1$) &  27\\
 & Hidden layers                     & 2 \\
 & Hidden units per layer            & 16 \\
 & Activation                        & SiLU \\
 & Output nonlinearity               & $\tanh$ \\[3pt]
\multicolumn{3}{@{}l}{\emph{ODE solver}} \\[2pt]
 & Method                            & RK4 \\
\multicolumn{3}{@{}l}{\emph{Fixed-effect network ($\rho_\psi$)}} \\[2pt]
 & Input dimension ($d + \dim \bm{s}$) & 24 \\
 & Hidden layers                       & 2 \\
 & Hidden units per layer              & 16 \\
 & Output dimension ($p$)              & 2 \\
\multicolumn{3}{@{}l}{\emph{Random-effect network ($g_\xi$)}} \\[2pt]
 & Input dimension ($d$)              & 16 \\
 & Hidden layers                      & 2 \\
 & Hidden units per layer             & 8 \\
 & Output dimension ($q$)             & 3 \\[3pt]
\multicolumn{3}{@{}l}{\emph{Gated skip connections}} \\[2pt]
 & Static covariate groups ($K_s$)    & 4 \\
 & Group lasso $\lambda_{\mathrm{GL}}$ & 0.1 \\
\multicolumn{3}{@{}l}{\emph{Training}} \\[2pt]
 & Optimiser                          & Adam \\
 & Learning rate                      & 1e-3 \\
 & Batch size (subjects)              & 128 \\
 & Weight decay $\lambda_{\mathrm{wd}}$ & 1e-5 \\
 & Maximum epochs                     & 1000 \\
 & Early stopping patience            & 300 \\
\multicolumn{3}{@{}l}{\emph{Total parameters}} \\[2pt]
 & $\dim(\theta)$                     & $\approx 1{,}600$ \\
\hline
\end{tabular}
\end{table}

% ══════════════════════════════════════════════════════════════
\section{Model selection results}
\label{sec:supp_cv}
% ══════════════════════════════════════════════════════════════

The Neural ODE-LMM exposes several architectural and regularisation choices whose impact on the IST predictor cannot be anticipated from theory alone.
We carried out a held-out dataset of the 3C cohort to select the configuration used in the main analyses.

\paragraph{Protocol.}
The 3C cohort was partitioned at the subject level into three disjoint sets: training (4,687 subjects), validation (1,172 subjects), and test (1,465 subjects).
For every configuration in the grid, the model was fitted on the training set with early stopping driven by the marginal NLL evaluated on the validation set; the best-validation checkpoint was retained.
Configurations were ranked by their validation-set NLL.
The test set was touched exactly once, to report the unbiased performance of the selected configuration.
Covariate standardisation statistics were computed from the training set only.

\paragraph{Grid.}
Two successive grids were explored.
Tier~1 varied two axes: the regularisation mode (none, group lasso)
with penalty weight $\lambda_{\mathrm{GL}} \in \{0.001,\,0.01,\,0.1\}$,
and the presence or absence of dynamic covariates in the skip path.
Configurations in which the skip path was absent while a penalty was
active were omitted as vacuous.
Tier~2 fixed the Tier-1 winner's regularisation
($\lambda_{\mathrm{GL}} = 0.1$) and skip choices and
swept the decoder capacity: latent dimension $d \in \{8, 16\}$
and fixed-effect output dimension $p \in \{2, 4\}$.
Table~\ref{tab:supp_cv} reports both tiers.
All other hyperparameters (learning rate, batch size, solver order) 
were held at the defaults reported in Table~\ref{tab:supp_architecture}.

\begin{table}[!t]
\centering
\caption{Model selection results on the 3C cohort ($n_{\mathrm{val}} = 1{,}172$).
Total validation-set NLL
($\mathrm{NLL}_{\mathrm{per\text{-}subject}} \times n_{\mathrm{val}}$);
lower is better.
\emph{Tier~1} varies regularisation and skip-path configuration
(all with $d = 8$, $p = 4$, $q = 3$).
\emph{Tier~2} fixes the Tier-1 winner's regularisation
($\lambda_{\mathrm{GL}} = 0.1$, dynamic skip) and varies
latent dimension~$d$ and fixed-effect output dimension~$p$.
The $\rho$-normalisation flag had no effect on NLL and is suppressed.
The selected configuration is marked with~$\star$.}
\label{tab:supp_cv}
\small
\begin{tabular}{@{}lllccr@{}}
\hline\\[-9pt]
Tier & Config & Regularisation & $d$ & $p$ & Total val.\ NLL \\
\hline\\[-9.75pt]
\multirow{5}{*}{1}
 & 1 & None                              & 8 & 4 & 15{,}871 \\
 & 2 & None (no skip)                    & 8 & 4 & 15{,}843 \\
 & 3 & Group lasso ($\lambda_{GL}=0.001$)     & 8 & 4 & 15{,}862 \\
 & 4 & Group lasso ($\lambda_{GL}=0.01$)      & 8 & 4 & 15{,}845 \\
 & 5 & Group lasso ($\lambda_{GL}=0.1$)       & 8 & 4 & 15{,}841 \\[3pt]
\hline\\[-9.75pt]
\multirow{3}{*}{2}
 & 6  & Group lasso ($\lambda_{GL}=0.1$) &  8 & 4 & 15{,}841 \\
 & 7$^\star$ & Group lasso ($\lambda_{GL}=0.1$) & 16 & 2 & 15{,}837 \\
 & 8  & Group lasso ($\lambda_{GL}=0.1$) & 16 & 4 & 15{,}845 \\
\hline
\end{tabular}
\end{table}
\paragraph{Tier~1 selection.}
In Tier~1, Config~2 (no skip, no regularisation; NLL\,$= 15{,}843$)
and Config~5 (group lasso $\lambda_{GL} = 0.1$; NLL\,$= 15{,}841$)
achieve comparable predictive performance.
We prefer Config~5 because it retains the skip pathway and allows
the group lasso penalty to adjudicate covariate routing in a
data-driven manner: the resulting near-zero skip weight norms
(Section~\ref{sec:supp_skip_norms}) provide empirical evidence
that covariate effects are mediated through the ODE latent state,
rather than imposing this routing by architectural constraint.
That Config~2 and Config~5 reach essentially the same NLL is
itself informative. It confirms that the group lasso recovers
the no-skip solution when the data support it.

\paragraph{Tier~2 selection.}
Tier~2 configurations achieve similar validation NLL (within
8~units over $n_{\mathrm{val}} = 1{,}172$ subjects).
Config~7 ($d = 16$, $p = 2$) is selected as it attains the
lowest NLL and the larger latent dimension provides richer
trajectory encoding for the downstream counterfactual prediction analyses.

% ══════════════════════════════════════════════════════════════
\section{Model selection validation (simulation)}
\label{sec:supp_model_selection}
% ══════════════════════════════════════════════════════════════

To select the best model architecture for each simulation scenario, 
we compared three configurations on 10 replicates and ranked them by total test NLL. 
Ten replicates were sufficient to reveal a clear winner in both scenarios (Table~\ref{tab:supp_model_sel}). 
For Scenario~S1 (instantaneous effect), the
three configurations are: (A)~BMI in skip only (ODE receives $[\Lambda,
t]$), (B)~BMI in both ODE and skip, no group lasso, and (C)~BMI in
both, with group lasso ($\lambda_{\mathrm{GL}} = 0.01$). For Scenario~S2
(cumulative effect): (A)~BMI in ODE only (no skip), (B)~BMI in both, no
group lasso, and (C)~BMI in both, with group lasso ($\lambda_{\mathrm{GL}} = 0.5$).
Table~\ref{tab:supp_model_sel} reports the total test NLL with $n_{\mathrm{test}} = 1{,}172$ and the number
of replicates in which each configuration achieved the lowest NLL.

In S1, model selection consistently favours the skip-only configuration
(9/10 wins), correctly identifying the instantaneous mechanism.
In S2, the group lasso configuration dominates (8/10 wins),
confirming that the penalty correctly steers BMI toward the ODE pathway
when the effect is cumulative. In both scenarios, the configuration
without regularisation is never selected, indicating that the penalised
likelihood criterion effectively distinguishes the two covariate
pathways.
\begin{table}[!t]
\centering
\caption{Model selection for simulation: total test NLL
and number of wins across 10 replicates.}
\label{tab:supp_model_sel}
\begin{tabular}{@{}llrrr@{}}
\hline\\[-9pt]
Scenario & Config & Mean total NLL & SE & Wins \\
\hline\\[-9.75pt]
\multirow{3}{*}{S1}
  & A: skip only           & 32{,}495 & 85.7 & 9/10 \\
  & B: both, no reg        & 32{,}504 & 90.6 & 0/10 \\
  & C: both, group lasso   & 32{,}505 & 91.0 & 1/10 \\[3pt]
\multirow{3}{*}{S2}
  & A: ODE only            & 32{,}542 & 93.4 & 2/10 \\
  & B: both, no reg        & 32{,}546 & 91.8 & 0/10 \\
  & C: both, group lasso   & 32{,}540 & 93.6 & 8/10 \\
\hline
\end{tabular}
\end{table}
% ══════════════════════════════════════════════════════════════
\section{Stationarity and Fisher diagnostics}
\label{sec:supp_stationarity}
% ══════════════════════════════════════════════════════════════

The validity of the empirical Fisher~$F$ as a variance estimator
requires the first-order stationarity condition
$\sum_i U_i \approx 0$ to hold at the estimated parameters. We report
two diagnostics: the norm of the mean penalised score
$\|\bar{U}\| = \|N^{-1}\sum_i U_i\|$ and the average individual score
norm $\overline{\|U_i\|} = N^{-1}\sum_i \|U_i\|$.
\begin{itemize}
  \item \textit{Simulation.}
        Table~\ref{tab:supp_stationarity_sim} reports these diagnostics
        averaged over the 100 replications for each simulation scenario.
  \item \textit{3C cohort.}
        For the selected Neural ODE-LMM model:
        $\|\bar{U}\| = 1.45$,
        $\overline{\|U_i\|} = 41.3$, and
        stationarity ratio $\|\bar{U}\| / \overline{\|U_i\|} = 0.035$.
\end{itemize}
\begin{table}[!t]
\centering
\caption{Stationarity diagnostics for simulation results (mean $\pm$ SD across 100 replicates).
The stationarity ratio $\|\bar{U}\| / \overline{\|U_i\|}$ quantifies
how close the mean score is to zero relative to typical individual
score magnitudes; values near zero indicate good stationarity.}
\label{tab:supp_stationarity_sim}
\vspace{4pt}
\begin{tabular}{@{}lrrr@{}}
\toprule
Scenario & $\|\bar{U}\|$ & $\overline{\|U_i\|}$ & Ratio \\
\midrule
S1 (skip only)      & $0.98 \pm 0.75$ & $23.5 \pm 1.5$ & $0.031 \pm 0.032$ \\
S2 (ODE + GL)       & $1.26 \pm 0.75$ & $40.2 \pm 2.7$ & $0.036 \pm 0.022$ \\
\bottomrule
\end{tabular}
\end{table}
We additionally verify that the $\Delta\bar{\mu}$ gradient
$\mathbf{d}_\ell = \partial \Delta\bar{\mu}(v_1, v_2;\, t^*_\ell)
/ \partial \boldsymbol{\theta}$ does not lie predominantly in the
null space of the raw (unregularised) Fisher $F$. For each evaluation time~$t^*_\ell$, we project
$\bm{d}_\ell$ onto the eigenvectors of~$F$ and compute the fraction of
squared energy below a relative threshold
($\lambda_k < 10^{-10}\lambda_{\max}$).
Table~\ref{tab:supp_nullspace} reports these fractions.
\begin{table}[!t]
\centering
\caption{Null-space fraction check for simulation results $\|d_{\mathrm{null}}\|^2 / \|d\|^2$ for
the $\dPDP$ gradient at each evaluation time (mean across replicates).}
\label{tab:supp_nullspace}
\vspace{5pt}
\begin{tabular}{@{}lrrrr@{}}
\toprule
& $t=0$ & $t=4$ & $t=8$ & $t=12$ \\
\midrule
S1  & 0.000 & 0.000 & 0.000 & 0.000 \\
S2  & 0.000 & 0.000 & 0.000 & 0.000 \\
\bottomrule
\end{tabular}
\end{table}

The stationarity ratios $\|\bar{U}\| /
\overline{\|U_i\|} < 0.05$ confirm that the mean penalised
score is small relative to typical individual score magnitudes,
validating the first-order condition underlying the empirical Fisher
variance estimator.
The null-space fraction is exactly zero at all evaluation times,
indicating that the $\dPDP$ gradient has full support in the
well-estimated subspace of parameter space and the delta-method
variance is not inflated by near-singular Fisher directions.

For Ledoit–Wolf shrinkage, the oracle shrinkage intensity $\alpha^{\ast}$ was 0.0009 ± 0.0001 (S2) and 
0.0030 ± 0.0005 (S1) 
across replicates. Values being close to zero indicate that the empirical Fisher $F$ was 
already well-conditioned; the shrinkage target contributed less than 0.3\% 
to the regularised estimator $F_{reg}$, 
confirming that the delta-method variances are driven by the data rather than by the regularisation.

% ══════════════════════════════════════════════════════════════
\section{Subject-level predictions}
\label{sec:supp_individual}
% ══════════════════════════════════════════════════════════════

\begin{figure}[!t]
\centering
\includegraphics[width=\textwidth]{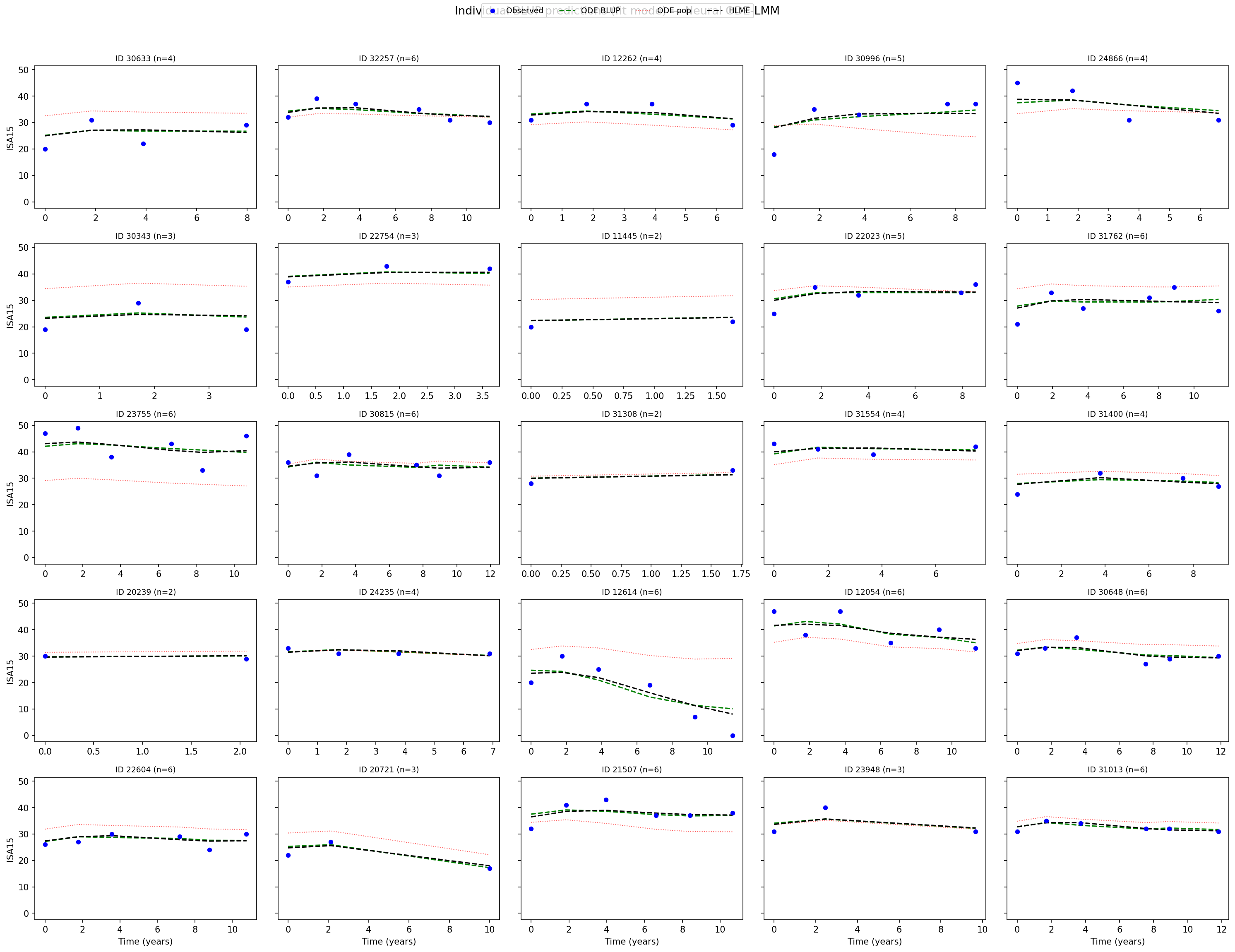}
\caption{Subject-level predictions in 3C cohort application
(training set, 25 randomly selected subjects). 
Blue dots: observed IST. 
Green dashed: Neural ODE-LMM predictions in fit mode. 
Red dotted: Neural ODE-LMM population mean.
Black dashed: LMM predictions.}
\label{fig:supp_individual_fit}
\end{figure}

\begin{figure}[!t]
\centering
\includegraphics[width=\textwidth]{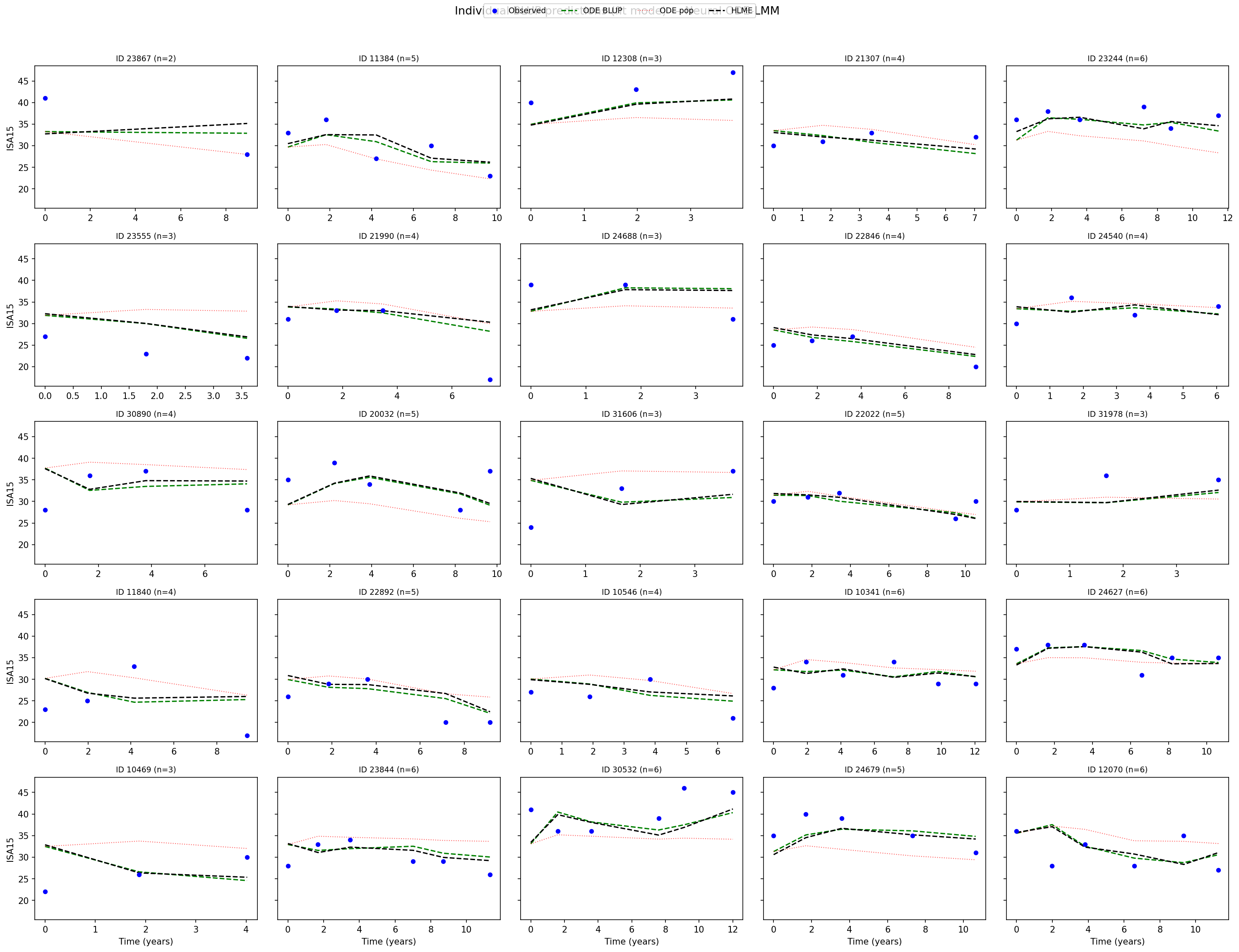}
\caption{Subject-level predictions in 3C cohort application (test set, 25 randomly selected subjects). Same colour scheme as Figure~\ref{fig:supp_individual_fit}.}
\label{fig:supp_individual_forecast}
\end{figure}

Figures~\ref{fig:supp_individual_fit} and~\ref{fig:supp_individual_forecast} display BLUP-based subject-level trajectory predictions for 25 randomly selected participants from the training and test sets, respectively.

% ══════════════════════════════════════════════════════════════
\section{Glucose counterfactual predictions}
\label{sec:supp_gluc}
% ══════════════════════════════════════════════════════════════

\begin{figure}[!t]
\centering
\includegraphics[width=0.48\textwidth]{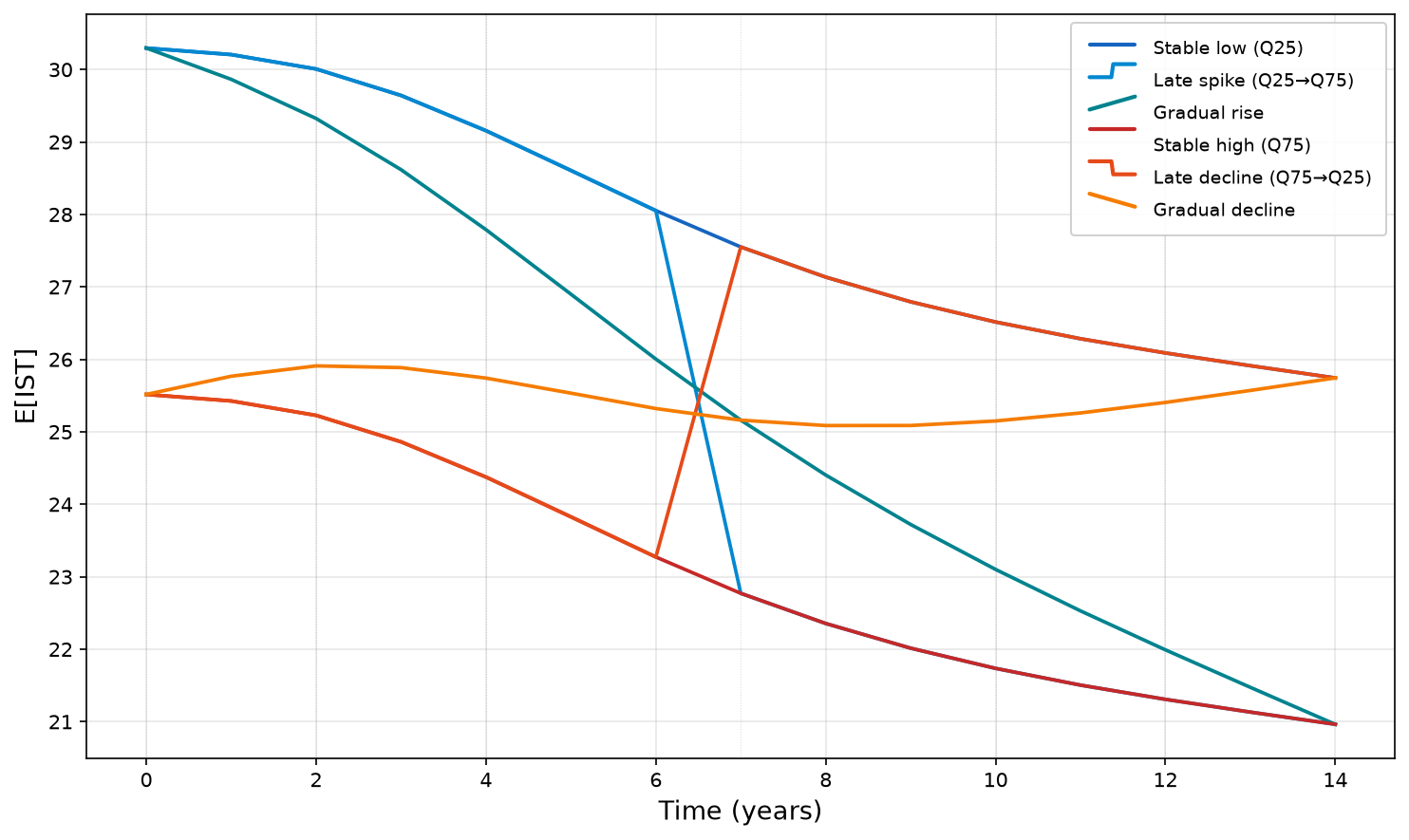}
\hfill
\includegraphics[width=0.48\textwidth]{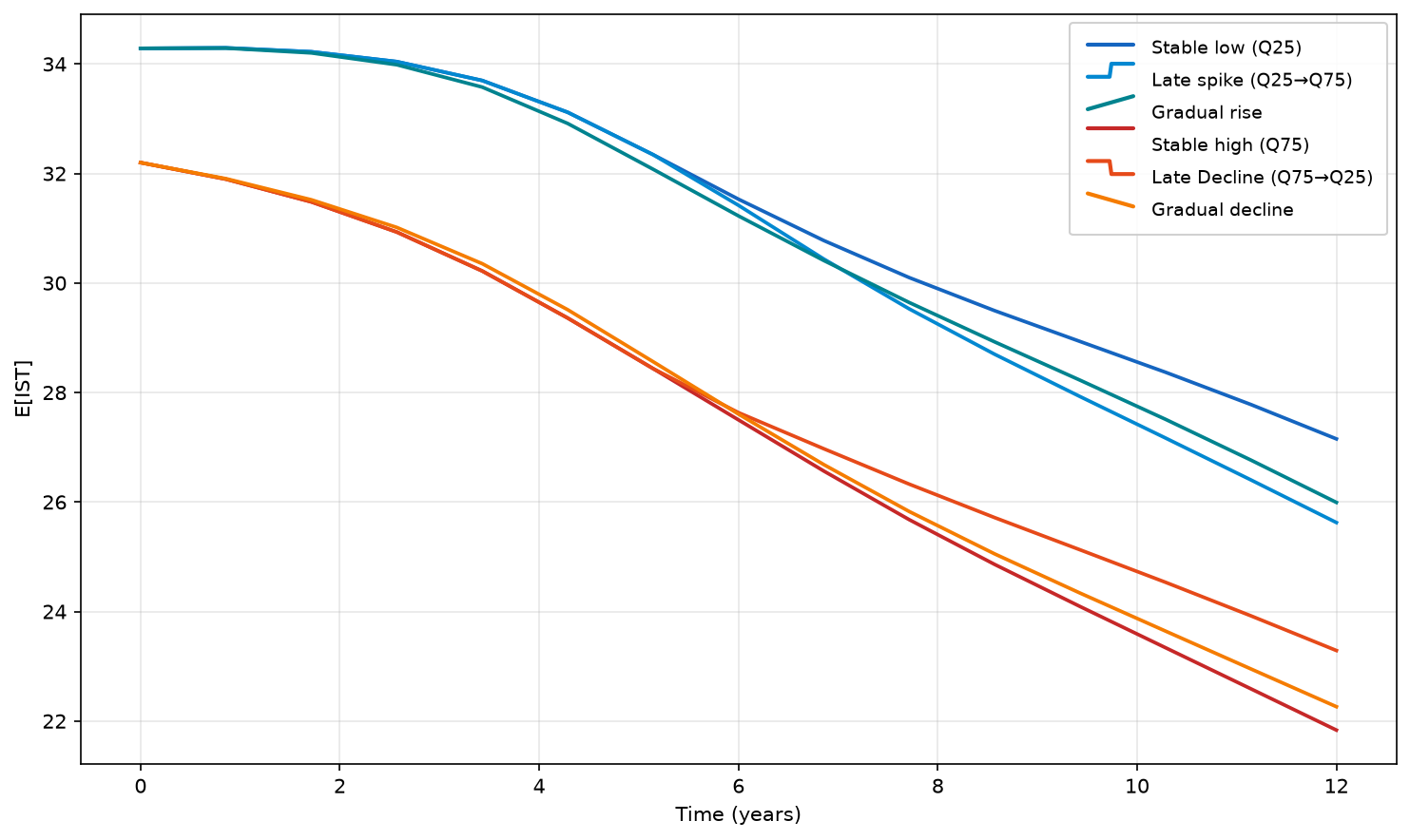}
\caption{trajectory-profile counterfactual predictions for fasting glucose in 3C cohort application.
Left: LMM. Right: Neural ODE-LMM.}
\label{fig:supp_gluc_pdp}
\end{figure}

Figure~\ref{fig:supp_gluc_pdp} reports the trajectory-profile counterfactual predictions for
fasting glucose, complementing the BMI analysis in the main text.

For fasting glucose, the trajectory-profile analysis reveals a
cumulative pattern. Profiles with sustained low glucose throughout
follow-up (stable low, late spike) maintain the highest predicted
cognitive scores, while profiles with sustained high glucose (stable
high, gradual decline) produce the worst outcomes. Critically, the late
spike profile (Q25$\to$Q75) remains substantially above the stable high
profile (Q75) at late follow-up, despite both profiles reaching the same
terminal glucose value. This indicates that the model has learned a
cumulative association: early low-glucose exposure confers lasting
protection even as glucose rises, while early high-glucose exposure
continues to penalise cognition even after glucose drops. The
approximately 5~IST-point spread between the best and worst profiles at
$t = 12$ years suggests a substantial effect of accumulated glucose
exposure on cognitive decline.

\begin{figure}[!t]
\centering
\includegraphics[width=0.48\textwidth]{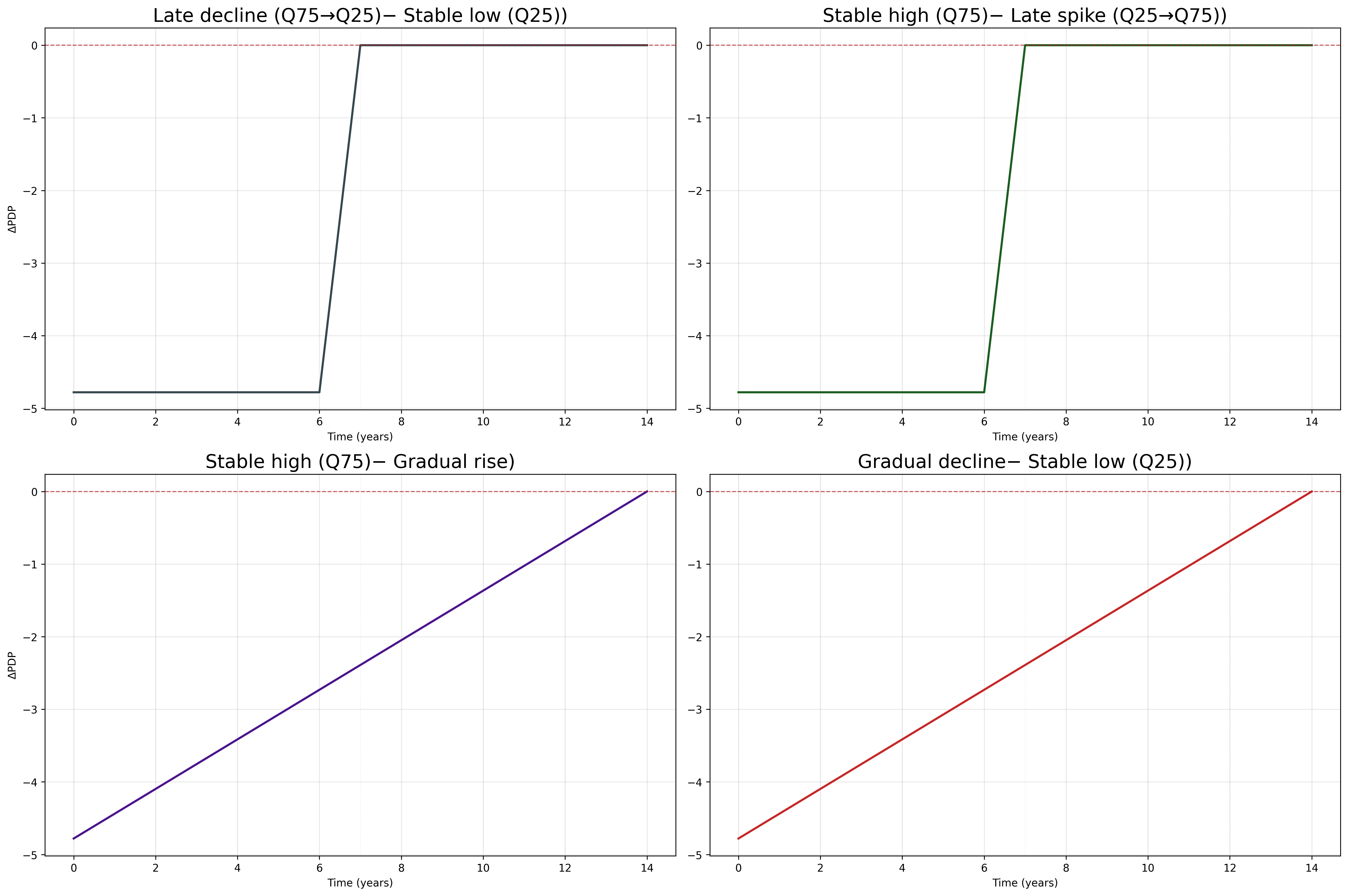}
\hfill
\includegraphics[width=0.48\textwidth]{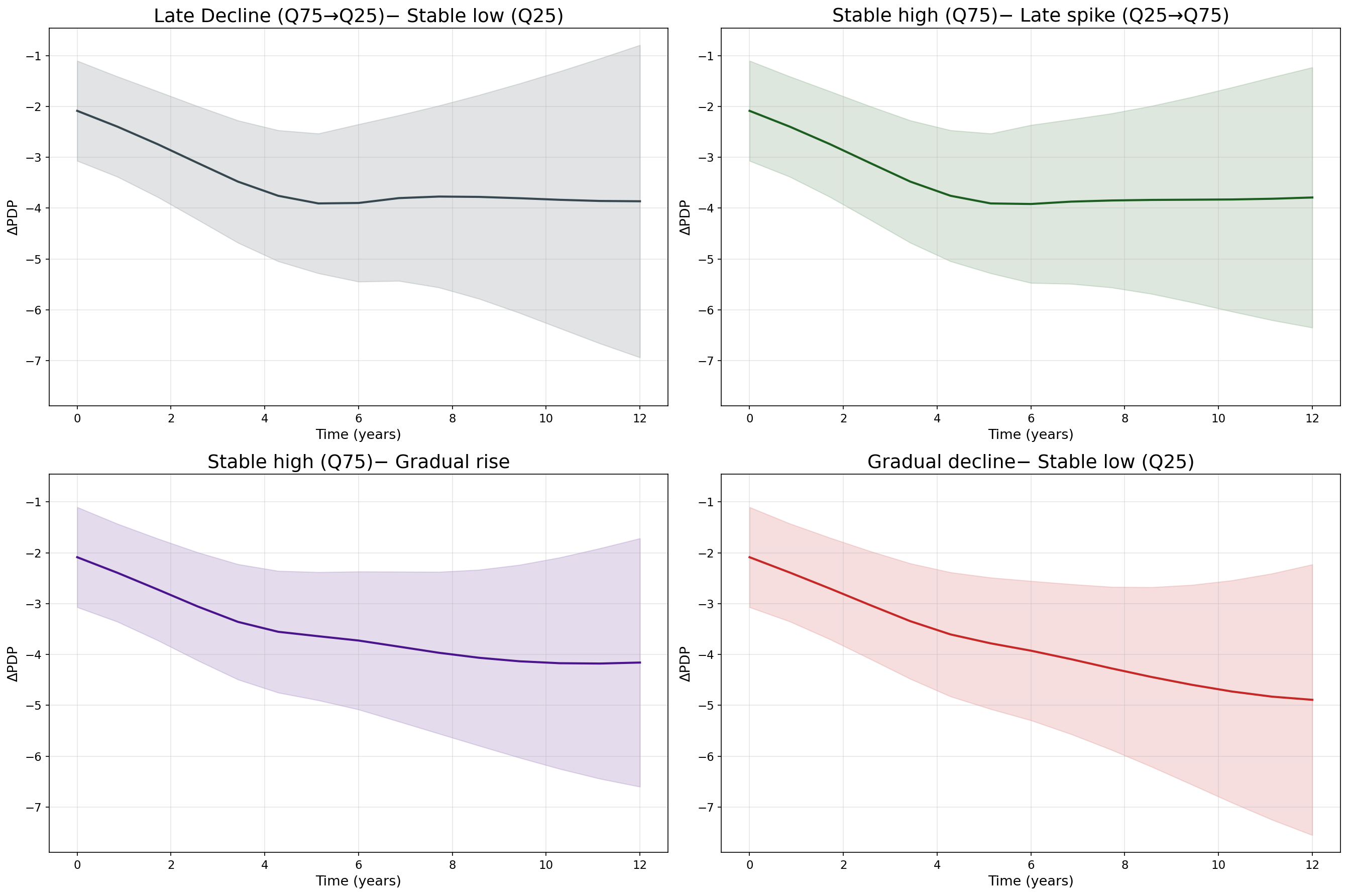}
\caption{Pairwise $\dPDP$ contrasts for fasting glucose in 3C cohort application.
Left: Classical LMM. Right: Neural ODE-LMM with 95\% delta-method CIs
(shaded bands).
Each panel shows $\dPDP(t) = \bar{\mu}(v_a, t) - \bar{\mu}(v_b, t)$
for one profile pair.}
\label{fig:supp_gluc_delta}
\end{figure}

Figure~\ref{fig:supp_gluc_delta} displays the pairwise $\dPDP$ contrasts
with 95\% delta-method confidence intervals for four representative
profile pairs.
Under the LMM, the $\dPDP$ reflects the purely instantaneous linear
effect of glucose: pairs that share the same current value (e.g.\
after the late-decline crossover) produce $\dPDP = 0$ exactly,
while pairs with a constant level difference yield a flat $\dPDP$
proportional to $\hat\beta_{\mathrm{gluc}}$.
Under the Neural ODE-LMM, the $\dPDP$ trajectories are smoother and
persist even after profiles converge to the same current value,
consistent with the cumulative encoding documented above.
All four contrasts are significantly negative throughout
follow-up (CIs exclude zero), confirming that higher glucose
trajectories are associated with worse cognition regardless of
the specific profile shape.
This cumulative glucose-cognition association is consistent with
epidemiological evidence from the Adult Changes in Thought (ACT)
study \citep{crane2013glucose}, which demonstrated a graded
relationship between average glucose levels over a five-year window
and subsequent dementia risk, even among individuals without diabetes.
In that cohort ($n = 2{,}067$, mean age 76), each increment in
average glucose was associated with increased dementia hazard,
with no apparent safe threshold. Our trajectory-profile analysis
extends this finding by showing that early glucose exposure leaves a
persistent imprint on predicted cognition that is not erased by later
normalisation.

% ══════════════════════════════════════════════════════════════
\section{HDL cholesterol trajectory-profile counterfactual predictions}
\label{sec:supp_hdl}
% ══════════════════════════════════════════════════════════════

\begin{figure}[!t]
\centering
\includegraphics[width=0.48\textwidth]{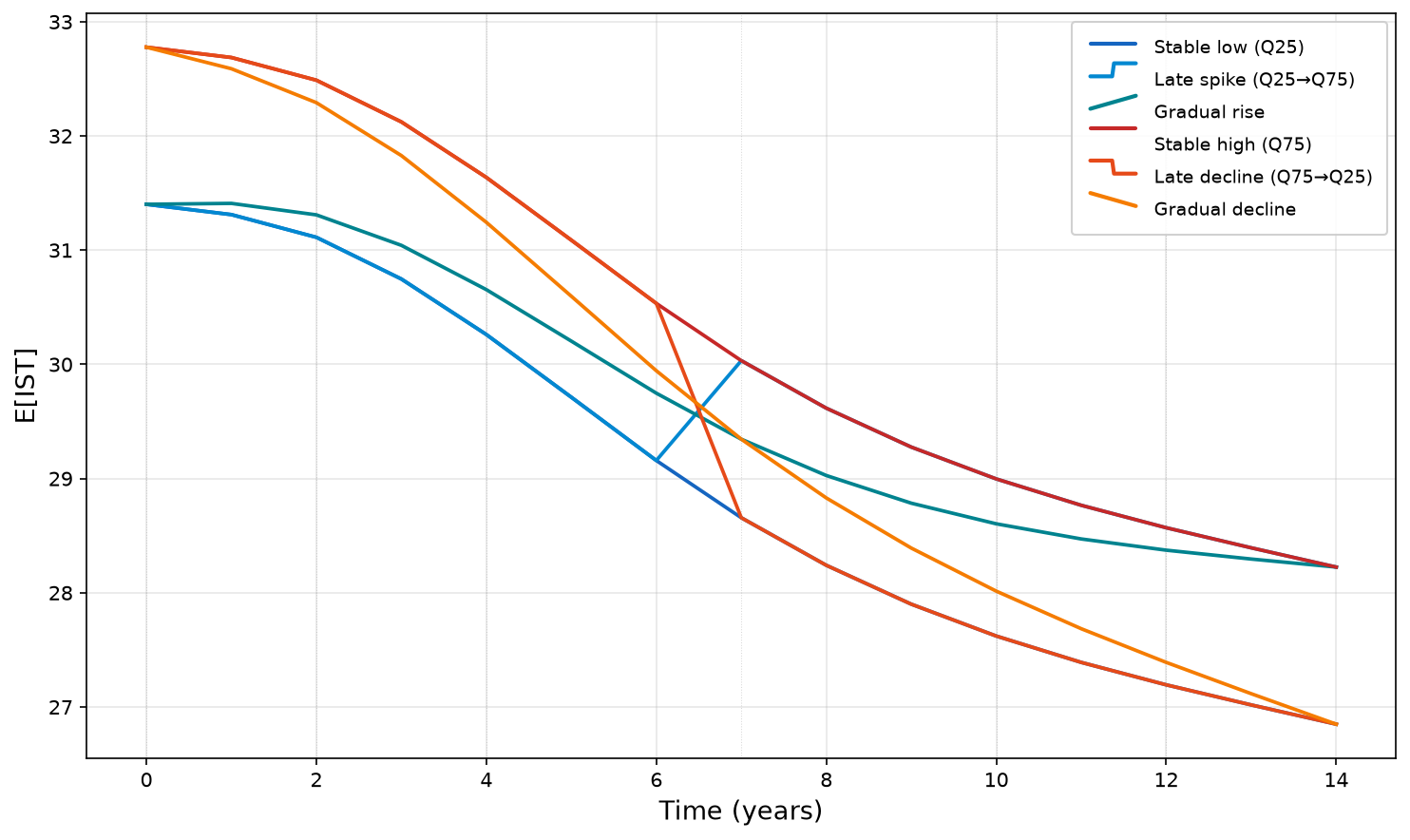}
\hfill
\includegraphics[width=0.48\textwidth]{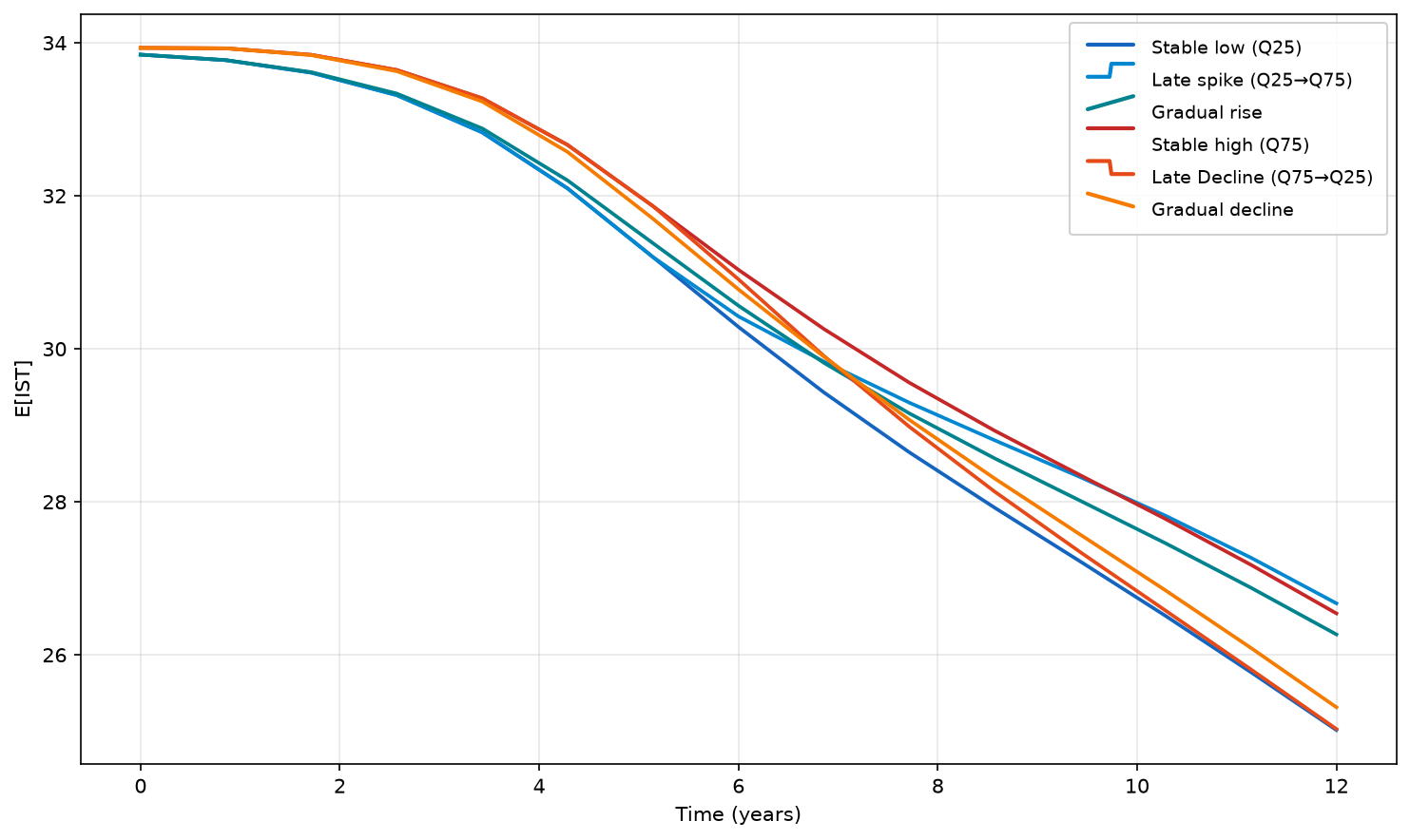}
\caption{trajectory-profile counterfactual predictions for HDL cholesterol in 3C cohort application.
Left: LMM. Right: Neural ODE-LMM.}
\label{fig:supp_hdl_pdp}
\end{figure}

Figure~\ref{fig:supp_hdl_pdp} reports the trajectory-profile counterfactual predictions for HDL cholesterol, complementing the BMI and glucose analyses in the main text (Figure~8).

For HDL cholesterol, the trajectory-profile analysis reveals a
qualitatively different pattern from BMI and glucose.
Profiles ending at high HDL at the final visit---stable high, late spike, and gradual rise---cluster at
the top of the cognitive trajectory, while profiles ending at low
HDL---stable low, early burden, and gradual decline---cluster at the
bottom. In particular, late spike (Q25~$\to$~Q75) outperforms
early burden (Q75~$\to$~Q25) at late visits, the opposite of the
pattern observed for BMI and glucose. This ordering is consistent with
a predominantly \emph{current-value} effect of HDL in late life: the
prognostic signal is carried by recent measurements rather than by
accumulated exposure over the preceding years.

Within the interquartile range examined here (Q25~$\to$~Q75), the
recovered effect is monotonic: higher current HDL is associated with
better cognition, in agreement with the cross-sectional findings of
\cite{crichton2014hdl} (2014) in a community cohort of adults aged~60 and
older, where higher current HDL-cholesterol was positively associated
with global cognition, Working Memory, and the MMSE after adjustment
for cardiovascular risk factors. The Q25--Q75 contrast, however,
cannot speak to behaviour at the extremes: the large Kaiser Permanente
study of \cite{ferguson2023hdl} ($n$~=~184{,}367, mean age~70) reported a
U-shaped association in which both the highest and lowest HDL quintiles
showed elevated dementia risk relative to the middle, with the
upper-tail effect appearing mainly above $\sim$2.1~mmo/L. Our contrast
lies well within the central range of that U and therefore recovers
its left arm only; testing for a turnaround at high HDL would require
wider counterfactual contrasts, which is a natural extension of the
analysis.

\begin{figure}[!t]
\centering
\includegraphics[width=0.48\textwidth]{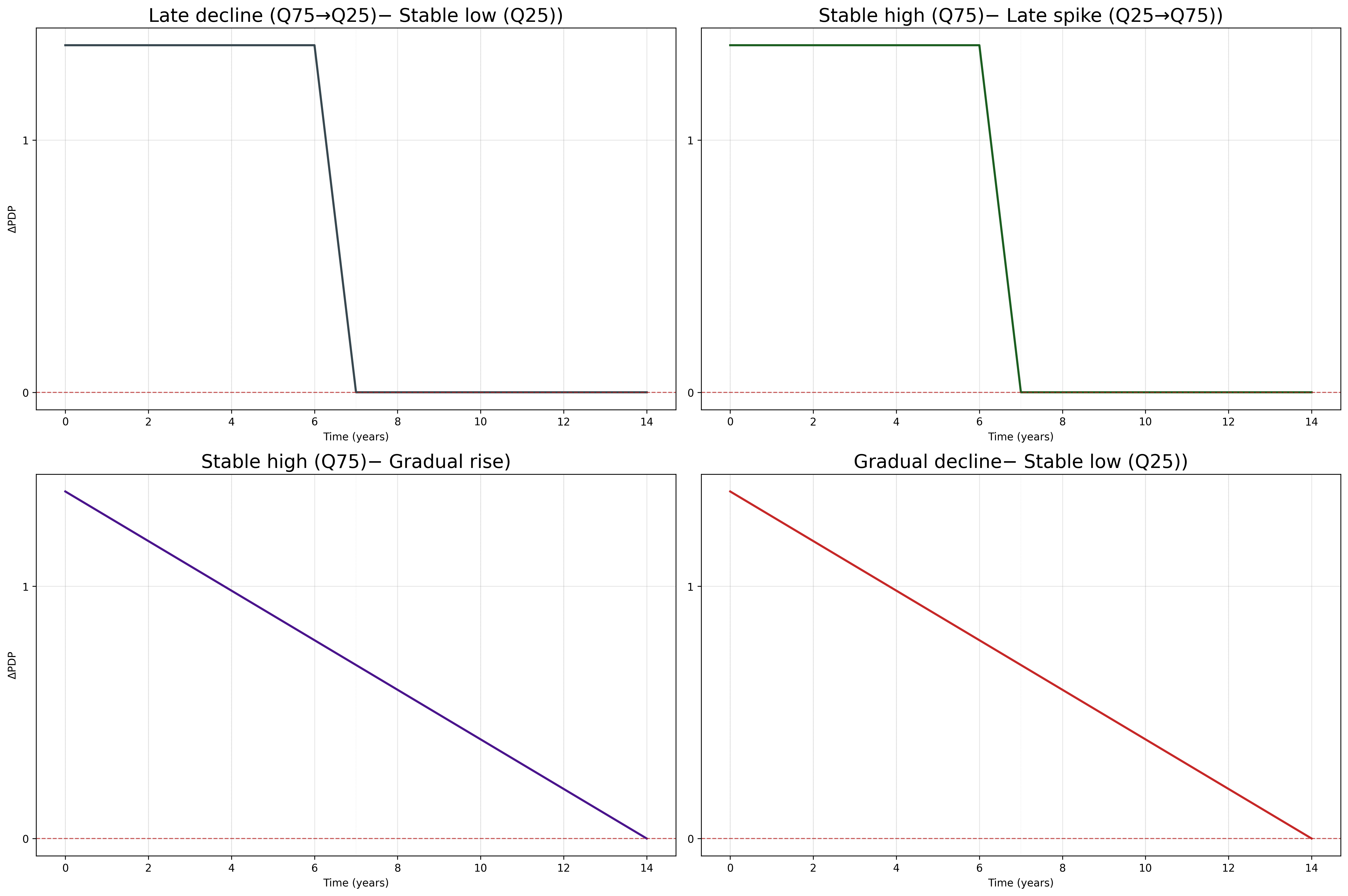}
\hfill
\includegraphics[width=0.48\textwidth]{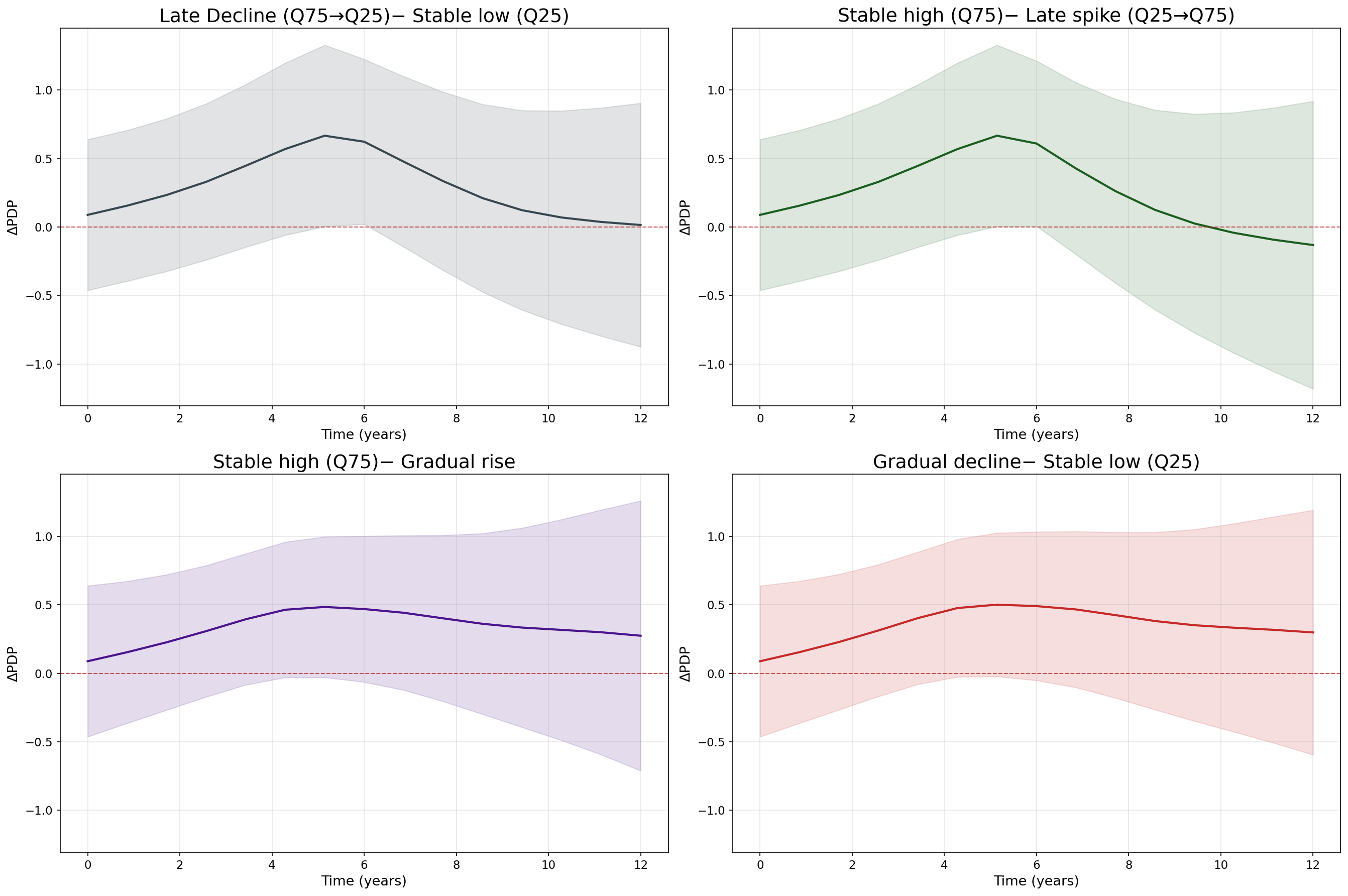}
\caption{Pairwise $\dPDP$ contrasts for HDL cholesterol in 3C cohort application.
Left: Classical LMM. Right: Neural ODE-LMM with 95\% delta-method CIs
(shaded bands).
The positive point estimates indicate that higher-HDL profiles are
associated with better cognition, but confidence intervals are wide
and generally include zero, consistent with a weaker effect than
glucose or BMI.}
\label{fig:supp_hdl_delta}
\end{figure}

Figure~\ref{fig:supp_hdl_delta} displays the pairwise $\dPDP$ contrasts.
Under the LMM, the pattern mirrors glucose but with reversed sign:
pairs with higher current HDL show positive $\dPDP$, and the contrast
drops to zero as soon as the profiles converge.
Under the Neural ODE-LMM, the $\dPDP$ point estimates are positive
for all four pairs (higher HDL $\to$ better cognition), but the
95\% delta-method confidence intervals are wide and include zero
for most of the follow-up window.
This is consistent with a weaker, predominantly current-value effect
of HDL on cognition within the interquartile range, in contrast to
the cumulative signal observed for glucose and BMI.

% ══════════════════════════════════════════════════════════════
\section{Zero-skip diagnostic: ODE-only trajectory-profile counterfactual predictions}
\label{sec:supp_zeroskip}
% ══════════════════════════════════════════════════════════════

To verify that the Neural ODE-LMM has learned cumulative covariate
effects through the ODE pathway rather than relying exclusively on the
instantaneous skip connections, we zero out all gated skip weights at
evaluation time and recompute the trajectory-profile counterfactual predictions using only
the ODE-decoded fixed effects.
Formally, we set
$\hat\mu_{ij}^{\text{ODE-only}} = \rho_\psi(\Lambda_i(t_{ij}))$
and suppress the skip contribution
$W_g^\top \bm{x}_{ij}^{(\text{dyn})}$.
If the ODE pathway encodes no covariate history, the resulting counterfactual predictions
will collapse to a single curve regardless of intervention profile;
conversely, persistent separation under different trajectory shapes
confirms that the ODE has internalised cumulative exposure information.

\begin{figure}[!t]
\centering
\includegraphics[width=0.65\textwidth]{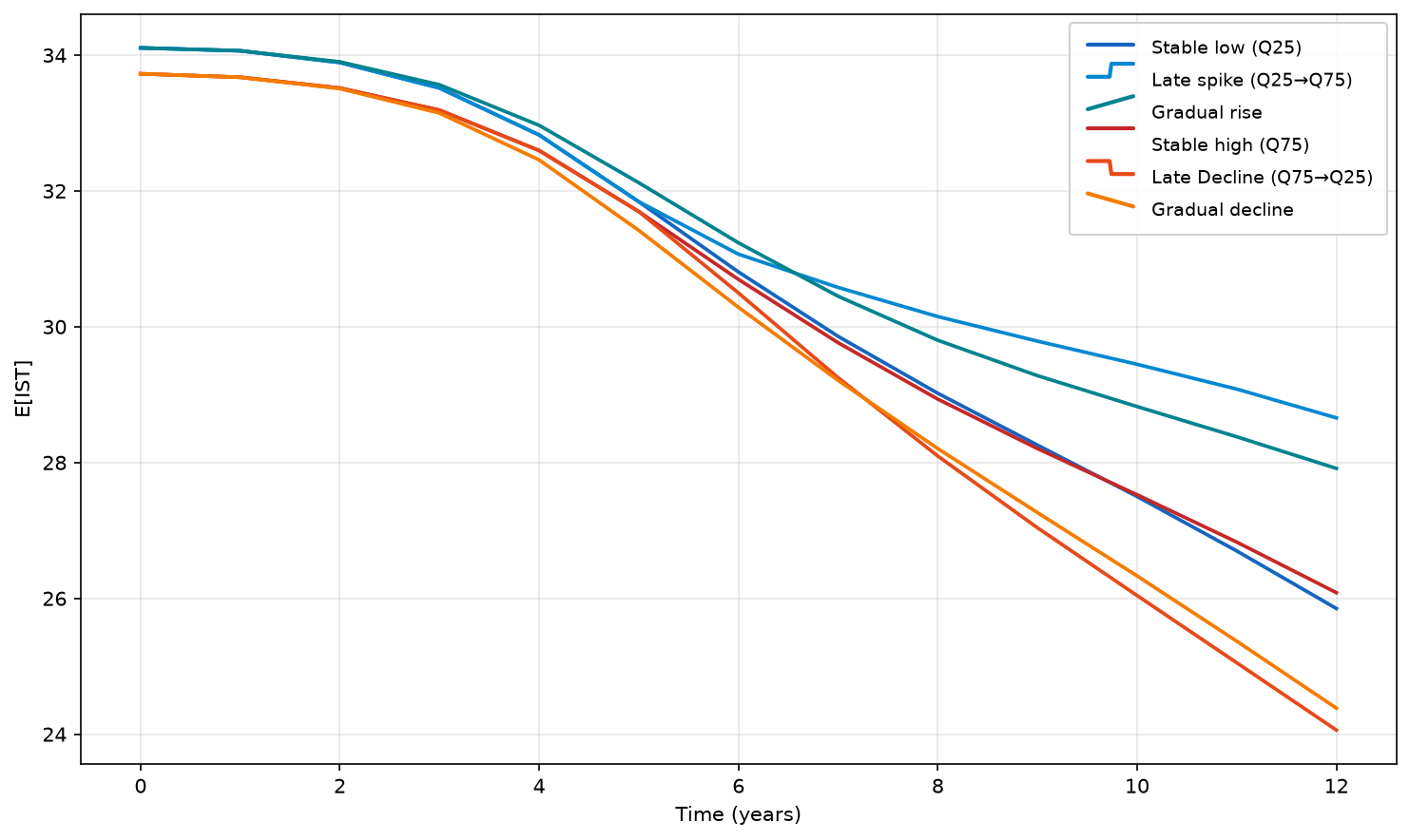}
\caption{ODE-only trajectory-profile counterfactual predictions for BMI in 3C cohort application (skip connections
zeroed). The persistent spread across profiles with identical endpoints
but different trajectories confirms that the ODE encodes cumulative BMI
exposure.}
\label{fig:supp_zeroskip_bmi}
\end{figure}

\begin{figure}[!t]
\centering
\includegraphics[width=0.65\textwidth]{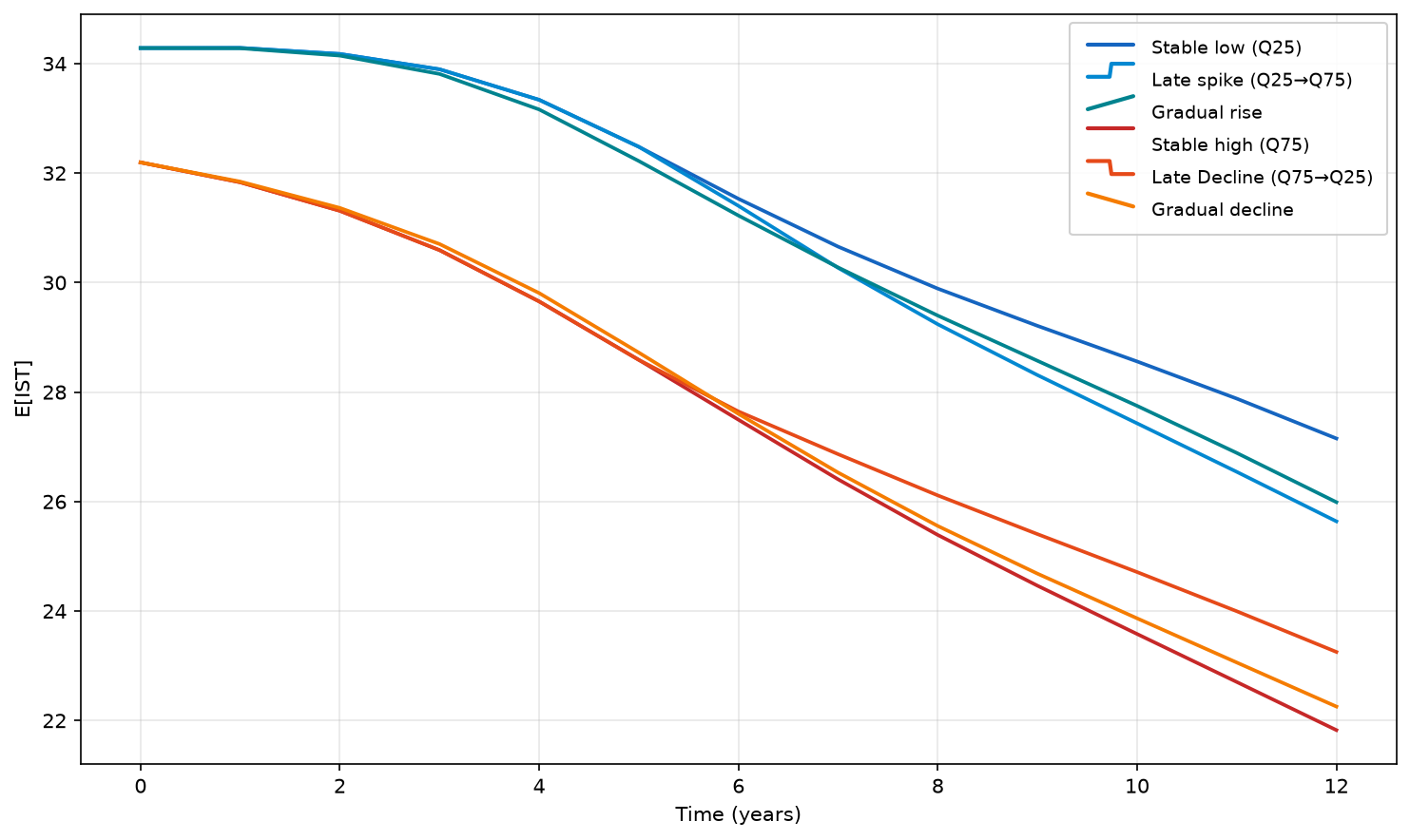}
\caption{ODE-only trajectory-profile counterfactual predictions for fasting glucose in 3C cohort application (skip
connections zeroed).}
\label{fig:supp_zeroskip_gluc}
\end{figure}

\begin{figure}[!t]
\centering
\includegraphics[width=0.65\textwidth]{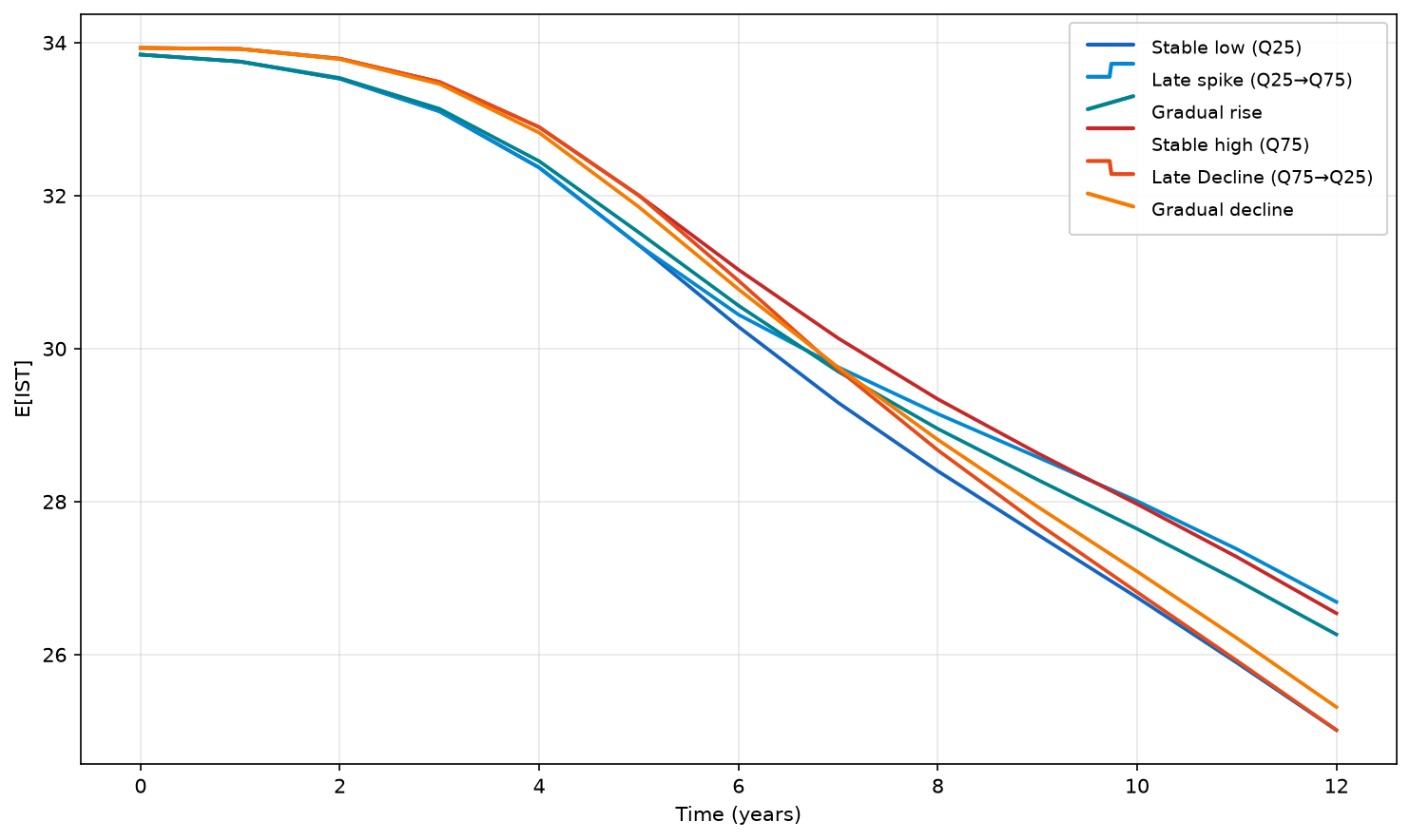}
\caption{ODE-only trajectory-profile counterfactual predictions for HDL cholesterol in 3C cohort application (skip
connections zeroed). The ordering, profiles ending at high HDL
outperform those ending at low HDL, is consistent with the full-model
counterfactual prediction, albeit with reduced spread.}
\label{fig:supp_zeroskip_hdl}
\end{figure}

Figures~\ref{fig:supp_zeroskip_bmi}--\ref{fig:supp_zeroskip_hdl}
display the ODE-only counterfactual predictions for BMI, fasting glucose, and HDL
cholesterol, respectively.
For all three covariates the trajectory ordering is preserved under
zero-skip evaluation, confirming that the ODE latent state~$\Lambda(t)$
encodes information about past covariate trajectories that persists
beyond the current value.
The spread is somewhat attenuated relative to the full-model counterfactual predictions
(Figures~8 and~\ref{fig:supp_gluc_pdp} in the main text and above),
consistent with the skip path carrying a portion of the contemporaneous
effect, but the cumulative signal is clearly present in the ODE
pathway alone.

% ══════════════════════════════════════════════════════════════
\section{Cumulative versus baseline-and-current diagnostic}
\label{sec:supp_cumdiag}
% ══════════════════════════════════════════════════════════════

A key question is whether the Neural ODE-LMM uses the full covariate
trajectory or relies only on the baseline and current values.
We construct a diagnostic that compares pairs of counterfactual profiles
that share the same starting and ending values but follow different
paths.
If the predicted cognitive trajectories under such pairs diverge, the
model has learned to use the intermediate trajectory; if
$\Delta = \E[\mathrm{IST} \mid \text{profile}_a] -
\E[\mathrm{IST} \mid \text{profile}_b] \approx 0$ at the endpoint,
the model effectively reduces to a baseline-plus-current-value
predictor.

Three profile designs are used for each covariate:
\begin{enumerate}
\item \emph{Constant low vs.\ temporary spike} (same start and end at
Q25): one profile remains at Q25 throughout; the other rises to Q75 at
the midpoint and returns.
\item \emph{Constant high vs.\ temporary dip} (same start and end at
Q75): one profile remains at Q75; the other drops to Q25 at the
midpoint and returns.
\item \emph{Same start and end, different curvature} (start at Q25,
end at Q75): linear, concave (late rise), and convex (early rise)
paths.
\end{enumerate}

\begin{figure}[!t]
\centering
\includegraphics[width=\textwidth]{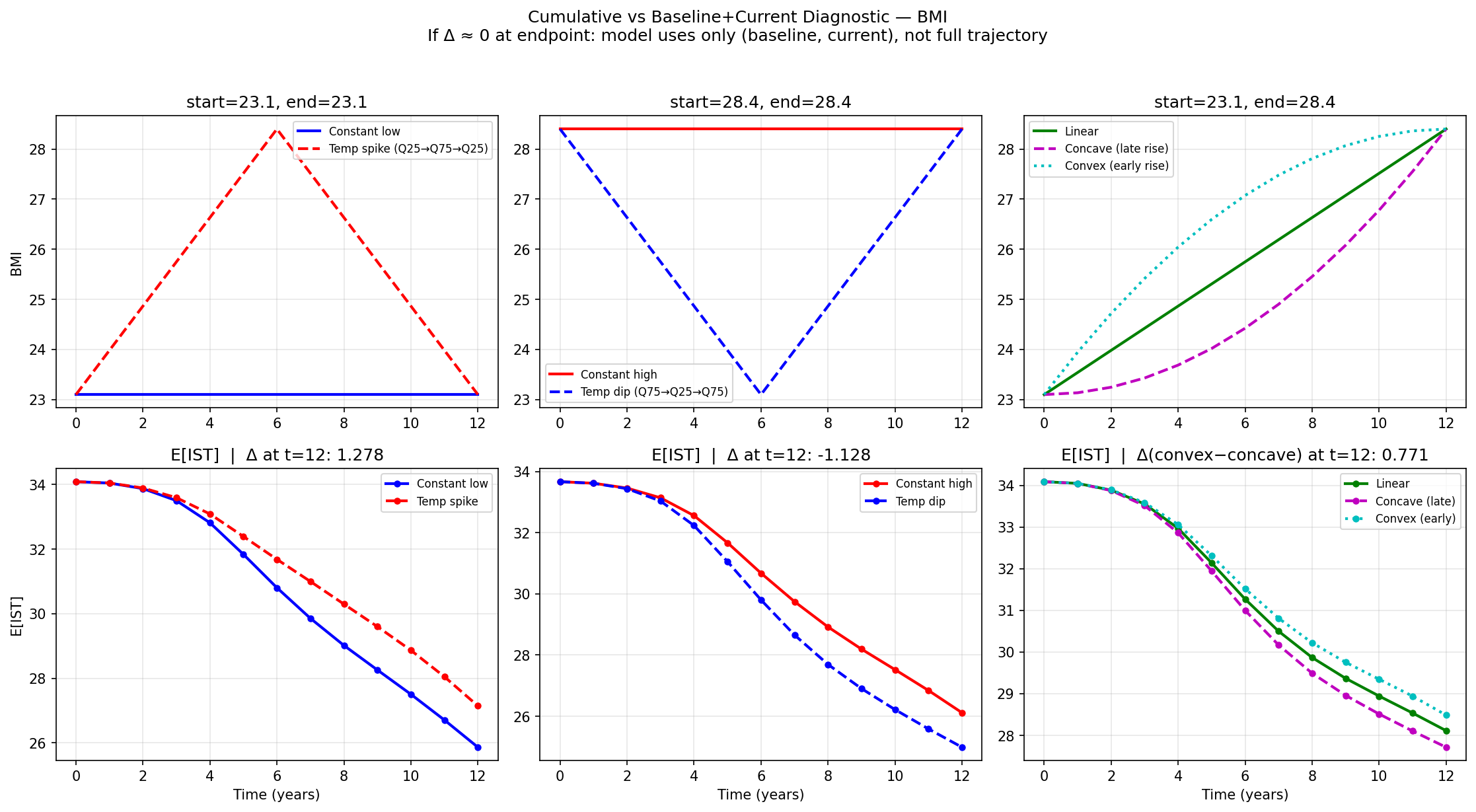}
\caption{Cumulative versus baseline-and-current diagnostic for BMI in 3C cohort application.
Upper panels: covariate profiles; lower panels: predicted E[IST].
Left: constant low vs.\ temporary spike ($\Delta_{t=12} = 1.28$).
Centre: constant high vs.\ temporary dip ($\Delta_{t=12} = -1.13$).
Right: linear vs.\ concave vs.\ convex paths
($\Delta_{\text{convex--concave}} = 0.77$).
Non-zero $\Delta$ at endpoint confirms the model uses the full
trajectory, not just baseline and current values.}
\label{fig:supp_cumdiag_bmi}
\end{figure}

\begin{figure}[!t]
\centering
\includegraphics[width=\textwidth]{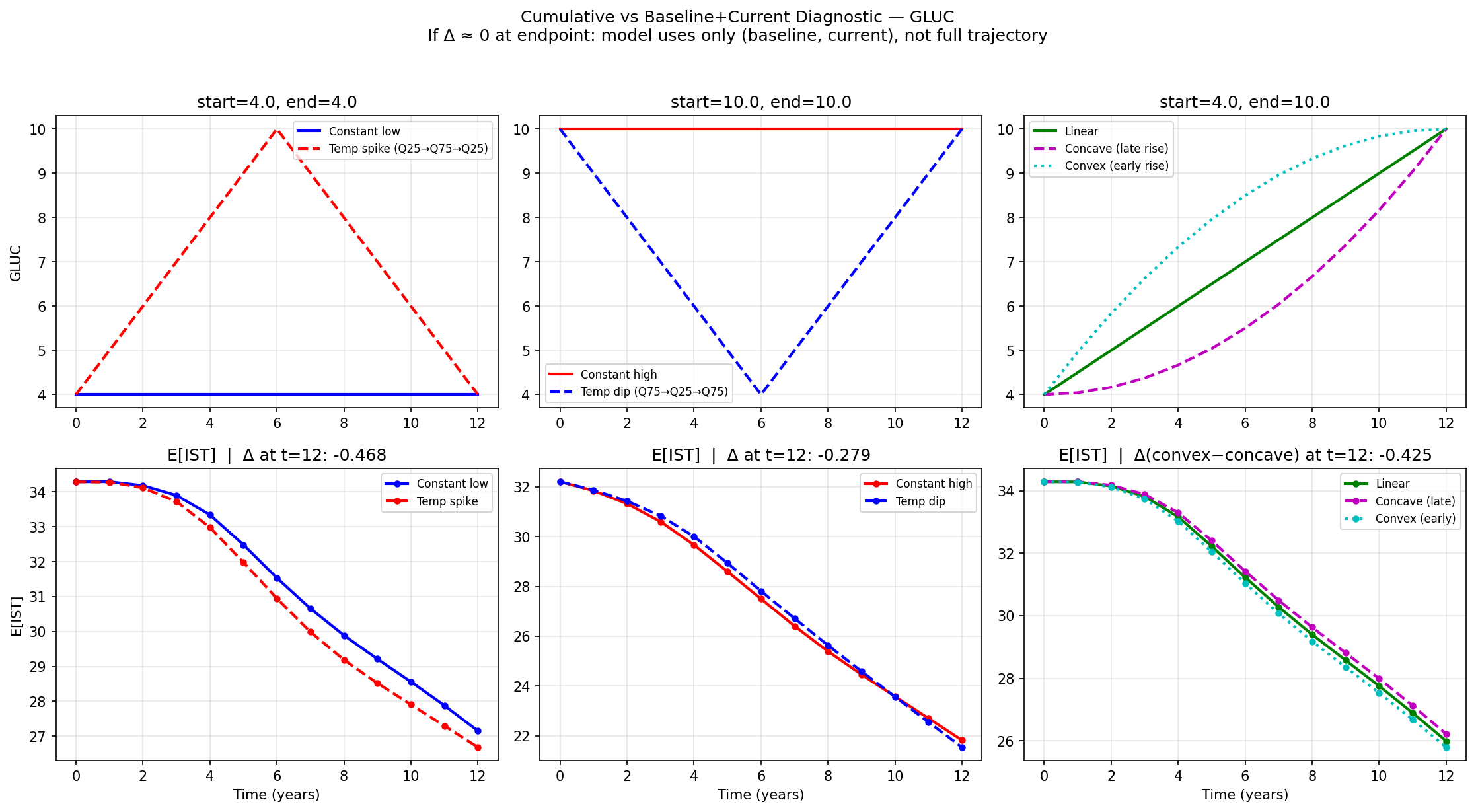}
\caption{Cumulative versus baseline-and-current diagnostic for fasting
glucose in 3C cohort application.
Left: constant low vs.\ temporary spike ($\Delta_{t=12} = -0.47$).
Centre: constant high vs.\ temporary dip ($\Delta_{t=12} = -0.28$).
Right: convex vs.\ concave ($\Delta = -0.43$).
All panels show non-zero endpoint differences, confirming cumulative
sensitivity.}
\label{fig:supp_cumdiag_gluc}
\end{figure}

Figures~\ref{fig:supp_cumdiag_bmi} and~\ref{fig:supp_cumdiag_gluc}
display this diagnostic for BMI and glucose.

For BMI, the diagnostic reveals substantial cumulative sensitivity:
a temporary spike from Q25 to Q75 and back yields
$\Delta_{t=12} = 1.28$ IST points relative to constant Q25 (left
panel), indicating that the transient high-BMI exposure leaves a lasting
imprint on predicted cognition despite both profiles ending at Q25.
Symmetrically, a temporary dip from Q75 to Q25 and back produces
$\Delta_{t=12} = -1.13$ relative to constant Q75 (centre panel).
The curvature comparison (right panel) shows that early BMI gains
(convex path) are more detrimental than late gains (concave path),
with $\Delta_{\text{convex--concave}} = 0.77$ at $t=12$.

For glucose, the pattern is qualitatively similar but smaller in
magnitude: $\Delta_{t=12} = -0.47$ for the spike comparison and
$-0.28$ for the dip comparison.
The sign reversal relative to BMI (negative $\Delta$ for the spike)
reflects the fact that for glucose, all profile groups share a
declining E[IST], and the spike profile declines slightly faster.
These non-zero endpoint differences confirm that the Neural ODE-LMM
has learned to encode cumulative covariate exposure for both BMI and
glucose, consistent with the ODE-only zero-skip analysis of
Section~\ref{sec:supp_zeroskip}.

% ══════════════════════════════════════════════════════════════

% ══════════════════════════════════════════════════════════════
\section{Skip-connection weight norms for real data application}
\label{sec:supp_skip_norms}
% ══════════════════════════════════════════════════════════════

Table~\ref{tab:supp_skip_norms} reports the Frobenius norm
$\|W_g\|_F$ of the first-layer decoder weights corresponding to each
covariate skip group, for the Neural ODE-LMM fitted on the 3C
cohort with group lasso penalty $\lambda_{\mathrm{GL}} = 0.1$.
Concretely, the first layer of~$\rho_\psi$ has weight matrix
$W \in \R^{h \times (d + K_{\mathrm{skip}})}$, where the first~$d$
columns multiply the latent state~$\Lambda(t)$ and the remaining
columns multiply the skip inputs (covariate values, masks, static
covariates).
For each covariate group~$g$, $W_g$ denotes the sub-matrix of columns
assigned to that group.
A norm near zero indicates that the group lasso has effectively
eliminated the instantaneous pathway for that covariate, routing its
effect exclusively through the ODE latent state.
The norm of the latent-state columns is
$\|W_{[:,:d]}\|_F = 6.27$.

All five time-varying covariates have skip norms below $10^{-3}$,
four to five orders of magnitude smaller than the latent-state
pathway, confirming that the group lasso
has driven the skip connections to near-zero for every covariate.
This means the fitted model routes essentially all time-varying covariate
information through the ODE, and the trajectory-profile counterfactual predictions
reported in the main text and above reflect the ODE pathway
almost entirely, consistent with the zero-skip diagnostic of
Section~\ref{sec:supp_zeroskip}, which showed minimal attenuation
when skip connections were explicitly zeroed.

\begin{table}[!t]
\centering
\caption{Skip-connection weight norms of the Neural ODE-LMM model fitted
on the 3C cohort ($\lambda_{\mathrm{GL}} = 0.1$).
$\|W_g\|_F$ is the Frobenius norm of the first-layer decoder columns
assigned to each skip group.
For reference, the norm of the columns multiplying the latent
state~$\Lambda(t)$ is $\|W_{[:,:d]}\|_F = 6.27$.}
\label{tab:supp_skip_norms}
\begin{tabular}{@{}llr@{}}
\hline\\[-9pt]
Covariate group & Type & $\|W_g\|_F$ \\
\hline\\[-9.75pt]
BMI               & Dynamic & 0.0008 \\
Systolic BP       & Dynamic & 0.0010 \\
Diastolic BP      & Dynamic & 0.0009 \\
Fasting glucose   & Dynamic & 0.0005 \\
HDL cholesterol   & Dynamic & 0.0005 \\
\hline
\end{tabular}
\end{table}

% ══════════════════════════════════════════════════════════════
\section{Latent state visualisation}
\label{sec:supp_latent}
% ══════════════════════════════════════════════════════════════

\begin{figure}[!t]
\centering
\includegraphics[width=0.65\textwidth]{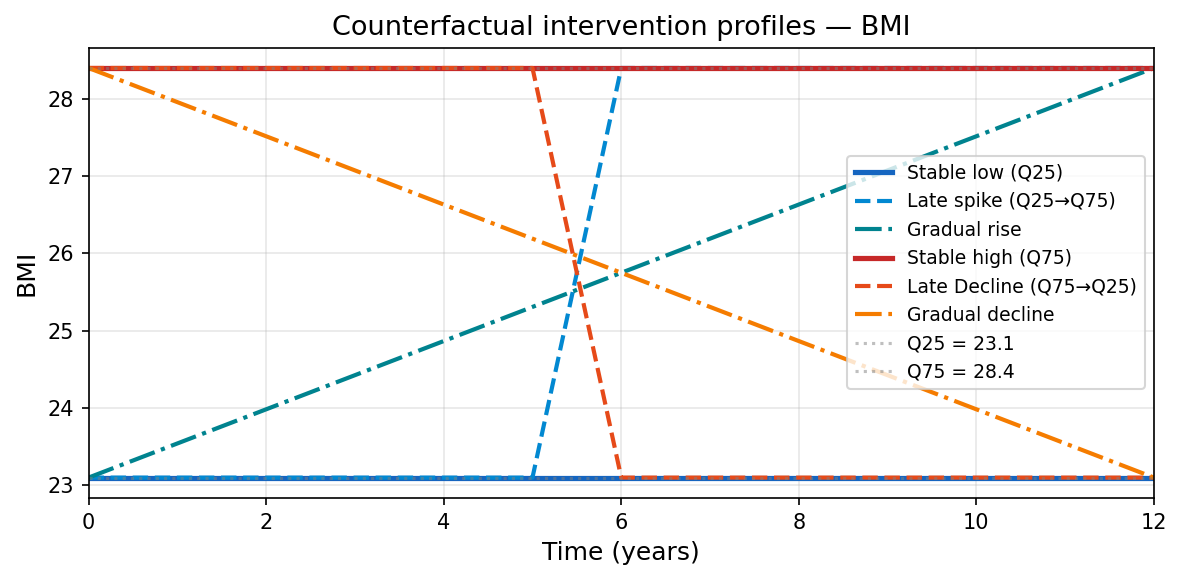}
\caption{Counterfactual intervention profiles for BMI, spanning the
interquartile range Q25\,=\,23.1 to Q75\,=\,28.4.
Analogous profiles are constructed for glucose and HDL cholesterol.}
\label{fig:supp_profiles_bmi}
\end{figure}

To examine how the ODE encodes covariate information, we conduct two
analyses on the fitted 3C model.
First, we define a set of counterfactual profiles for each covariate
(Figure~\ref{fig:supp_profiles_bmi}) spanning the interquartile range
(Q25--Q75) with distinct trajectory shapes: stable low, stable high,
late spike (Q25$\to$Q75), late decline (Q75$\to$Q25), gradual rise,
and gradual decline.
For each profile, we replace the target covariate with the prescribed
trajectory for all subjects, re-integrate the ODE, and compute the
population-averaged latent trajectory per dimension:
\[
  \bar{\Lambda}_k(v, t) = \frac{1}{N}\sum_{i=1}^{N}
  \Lambda_{i,k}^{(\mathrm{cov} \leftarrow v)}(t),
  \qquad k = 1,\ldots,d.
\]

\begin{figure}[!t]
\centering
\includegraphics[width=0.65\textwidth]{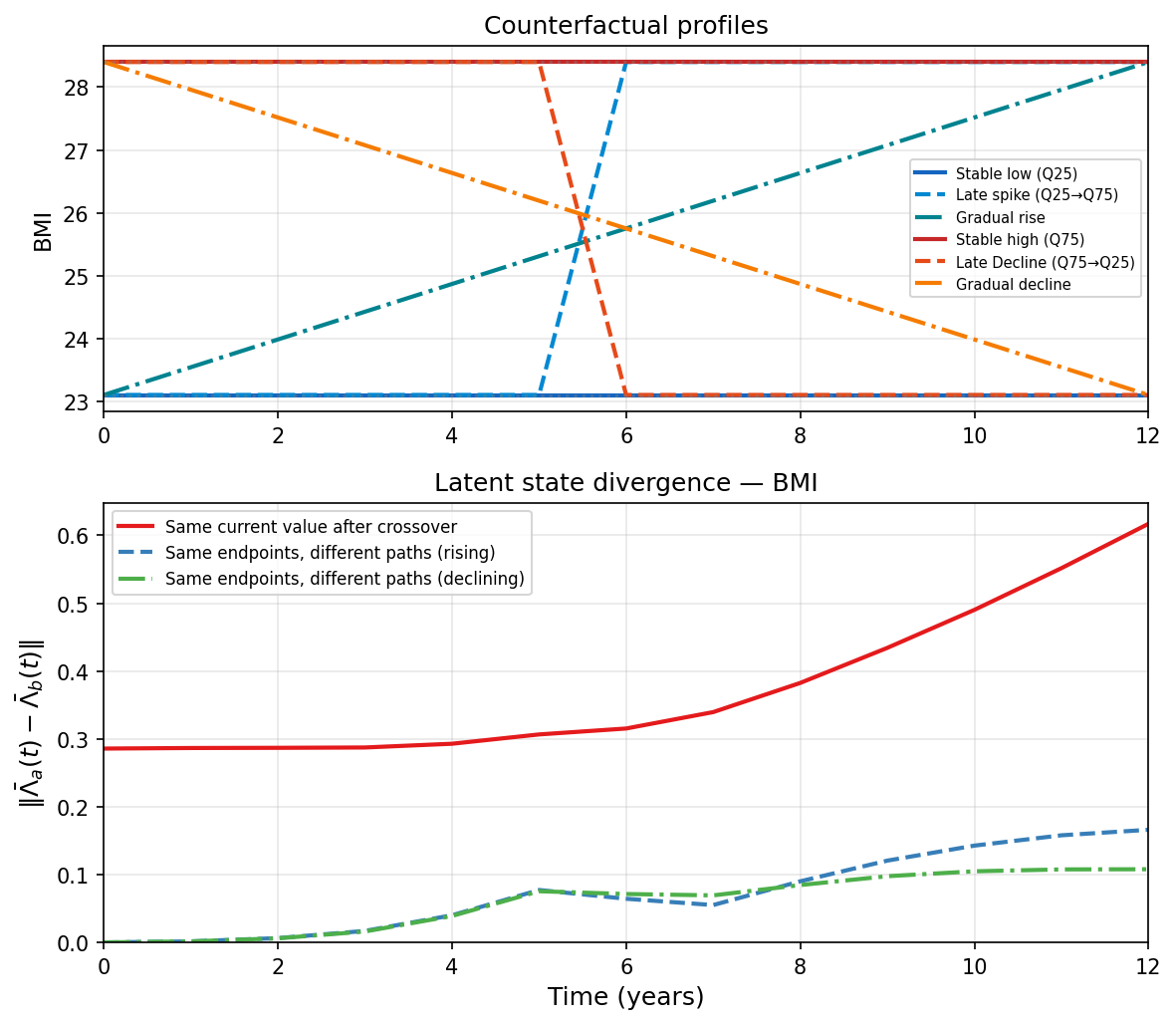}
\caption{Latent state divergence
$\|\bar\Lambda_a(t) - \bar\Lambda_b(t)\|$ for three pairs of BMI
counterfactual profiles.
The ``same current value after crossover'' pair (red) shows large
divergence, confirming that the ODE encodes trajectory history.
The ``same endpoints, different paths'' pairs (blue and green) show
smaller but non-zero divergence, confirming sensitivity to the path
shape.}
\label{fig:supp_latent_dist_bmi}
\end{figure}

\begin{figure}[!t]
\centering
\includegraphics[width=0.65\textwidth]{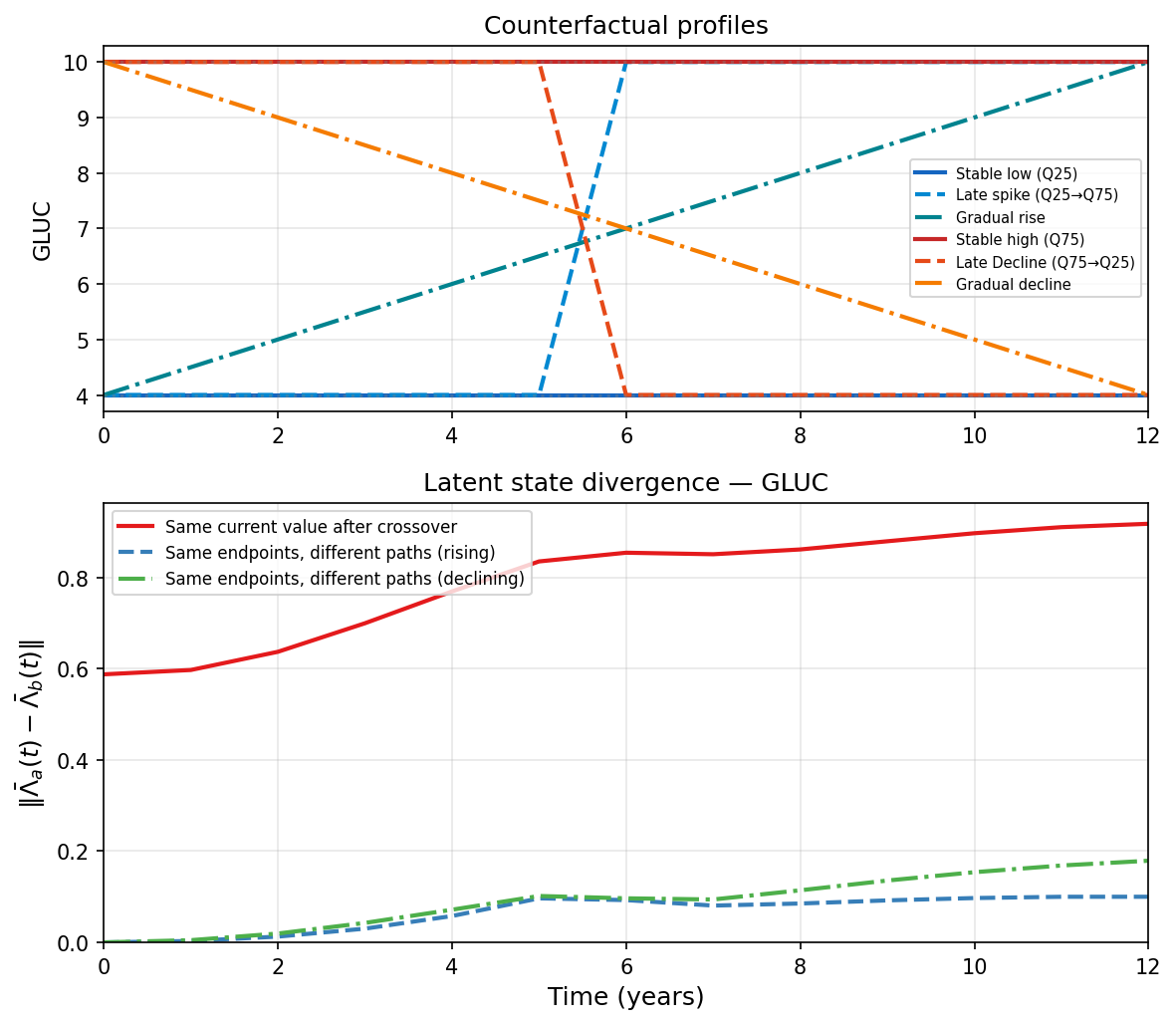}
\caption{Latent state divergence for three pairs of glucose
counterfactual profiles. Same structure as
Figure~\ref{fig:supp_latent_dist_bmi}.}
\label{fig:supp_latent_dist_gluc}
\end{figure}

\paragraph{Latent state divergence.}
To quantify whether the ODE distinguishes different covariate
trajectories, we compute the $L^2$ norm between the
population-averaged latent states under pairs of profiles,
$\|\bar\Lambda_a(t) - \bar\Lambda_b(t)\|$.
Figure~\ref{fig:supp_latent_dist_bmi} shows this divergence for three
BMI profile pairs and Figure~\ref{fig:supp_latent_dist_gluc} for
glucose.
Profiles that share the same endpoints but follow different paths
(``same endpoints, different paths'') produce non-zero latent
divergence, confirming that the ODE state encodes trajectory history
rather than merely the current covariate value.
The ``same current value after crossover'' pair (stable low vs.\ late
decline, which share the same covariate value after the crossover
point but have different histories) exhibits the largest divergence,
consistent with the ODE accumulating distinct historical information.

\begin{figure}[!t]
\centering
\includegraphics[width=\textwidth]{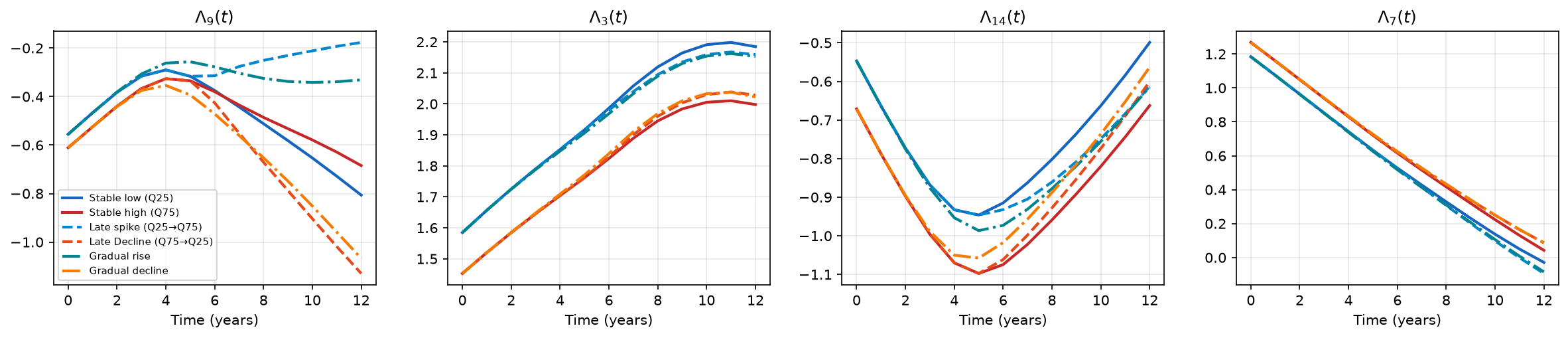}
\caption{Latent counterfactual predictions for the four most responsive dimensions to BMI
intervention profiles. Each panel shows one dimension $\Lambda_k(t)$
under the six counterfactual BMI profiles of
Figure~\ref{fig:supp_profiles_bmi}.}
\label{fig:supp_latent_pdp_bmi}
\end{figure}

\begin{figure}[!t]
\centering
\includegraphics[width=\textwidth]{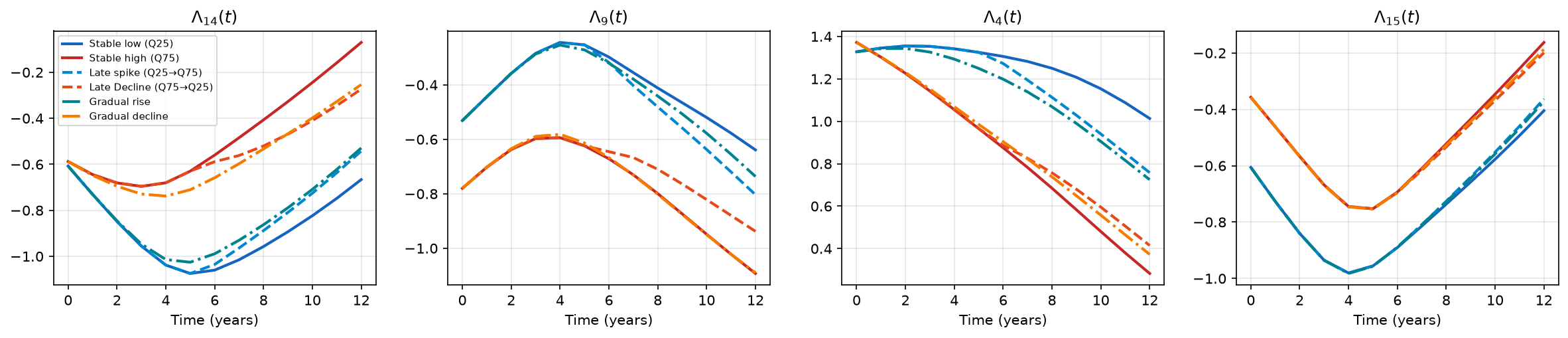}
\caption{Latent counterfactual predictions for the four most responsive dimensions to glucose
intervention profiles.}
\label{fig:supp_latent_pdp_gluc}
\end{figure}

\paragraph{Dimension-level latent counterfactual prediction}
Not all latent dimensions respond equally to covariate perturbations.
Figures~\ref{fig:supp_latent_pdp_bmi}
and~\ref{fig:supp_latent_pdp_gluc} display the population-averaged
latent trajectory $\bar{\Lambda}_k(v, t)$ for the four most responsive
dimensions (ranked by maximal inter-profile spread at $t = 12$).
For BMI, dimensions $\Lambda_9$ and $\Lambda_3$ show clear separation
by trajectory shape, while $\Lambda_{14}$ and $\Lambda_7$ respond
with more moderate differentiation.
For glucose, $\Lambda_{14}$ and $\Lambda_9$ are the most responsive,
with the stable-high profile producing the largest departures.
The shared responsiveness of $\Lambda_9$ and $\Lambda_{14}$ to both BMI
and glucose suggests that certain latent dimensions encode overlapping
cardiometabolic exposure information.

\end{document}